%% file: main.tex
\documentclass[10pt,journal,compsoc,review]{IEEEtran}
\ifCLASSOPTIONcompsoc
  \usepackage[nocompress]{cite}
\else
  \usepackage{cite}
\fi

\input{preamble}
\ifCLASSINFOpdf
\else
\fi

\usepackage[switch]{lineno}

\begin{document}
%\linenumbers
\input{sec/0_metadata}

\markboth{IEEE Transactions on Visualization and Computer Graphics}%
{Shell \MakeLowercase{\textit{et al.}}: Bare Advanced Demo of IEEEtran.cls for IEEE Computer Society Journals}



\IEEEtitleabstractindextext{%
\input{sec/0_abstract}
}

\maketitle

% is not selected <OR> if conference mode is selected - because compsoc
% conference papers position the abstract like regular (non-compsoc)
\IEEEdisplaynontitleabstractindextext

% \fi
\IEEEpeerreviewmaketitle

\begin{figure*}
  \includegraphics[width=1\textwidth]{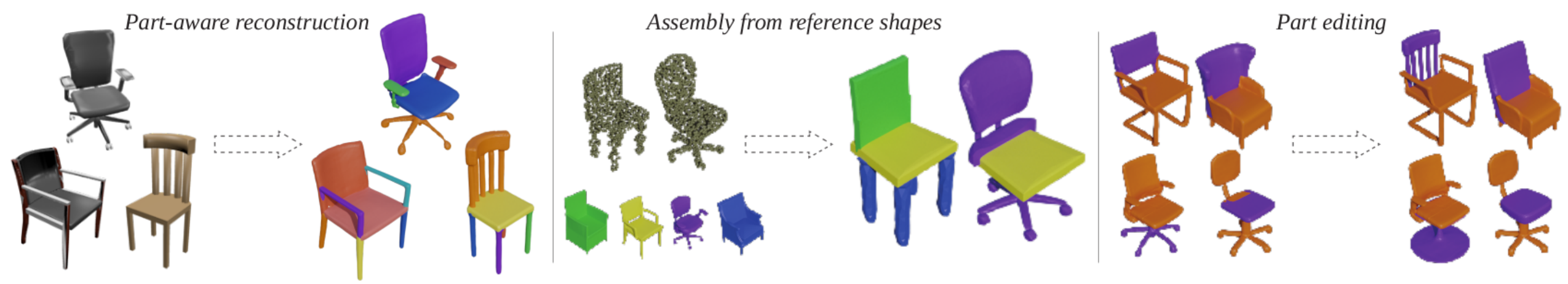}
%  \vspace{-20pt}
   \caption{
    We introduce a shape representation for 3D reconstruction through a set assembly of neural implicits (\textbf{left}). Our~approach results in state-of-the-art reconstruction quality. It can also perform reconstruction constrained on a database of reference parts or shapes (\textbf{middle}) -- the model here uses only shapes provided by the user. Finally, it allows part-level editing of outputs (\textbf{right}).
     \vspace{6pt}}
\label{fig:teaser}
\end{figure*}

\input{sec/1_introduction}

\input{sec/2_related}

\input{sec/3_method}

\input{sec/4_experiments}

\input{sec/5_results}

\input{sec/6_conclusions}

\ifCLASSOPTIONcaptionsoff
  \newpage
\fi



%\bibliographystyle{IEEEtran}

%\bibliographystyle{ACM-Reference-Format}
%\bibliography{sample-base}
%\bibliography{macros,main}

%\begin{thebibliography}
\bibliographystyle{IEEEtran}
\bibliography{macros,main}

\begin{IEEEbiography}[{\includegraphics[width=1in,height=1.25in,clip,keepaspectratio]{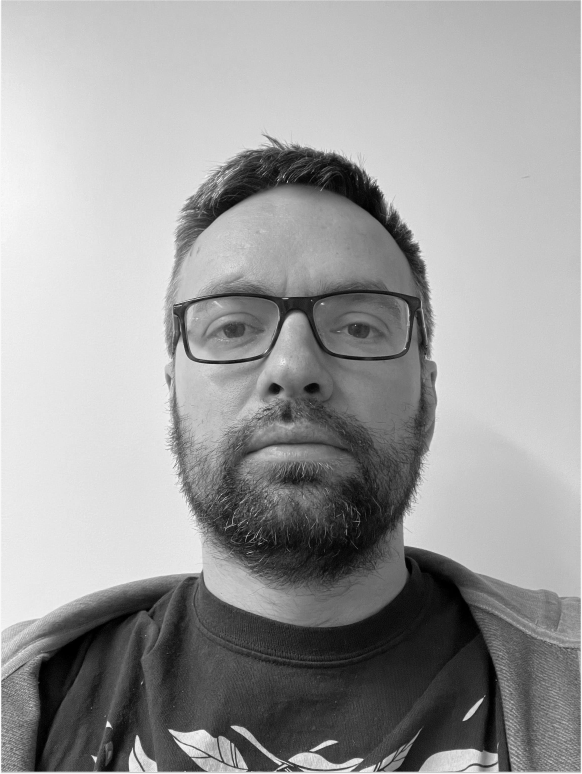}}]{Dmitrii Petrov}
 is currently a MS/PhD. student in University of Massachusetts Amherst working under supervision of Prof. Evangelos Kalogerakis. His research interests lie on intersection of computer vision and computer graphics. His work is focused on high-quality image and 3D shape generation with additional emphasis on fine-grained control and editability of resulting images and shapes. 
\end{IEEEbiography}

\begin{IEEEbiography}[{\includegraphics[width=1in,height=1.25in,clip,keepaspectratio]{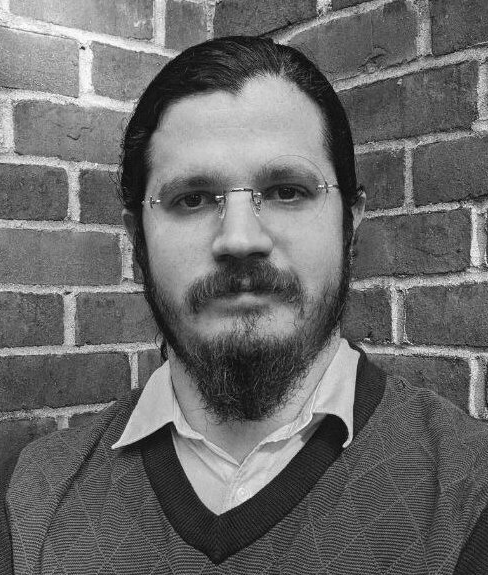}}]{Matheus Gadelha}
is currently a Research Scientist at Adobe Research. He received his PhD from University of Massachusetts Amherst while being supervised by Prof. Rui Wang and Prof. Subhransu Maji. He is interested in Computer Graphics, Vision and their intersections with Machine Learning. His work is focused on models and representations of tridimensional data for both discriminative and generative models with a particular interest in learning them directly from images.
\end{IEEEbiography}

%\newpage

\begin{IEEEbiography}[{\includegraphics[width=1in,height=1.25in,clip,keepaspectratio]{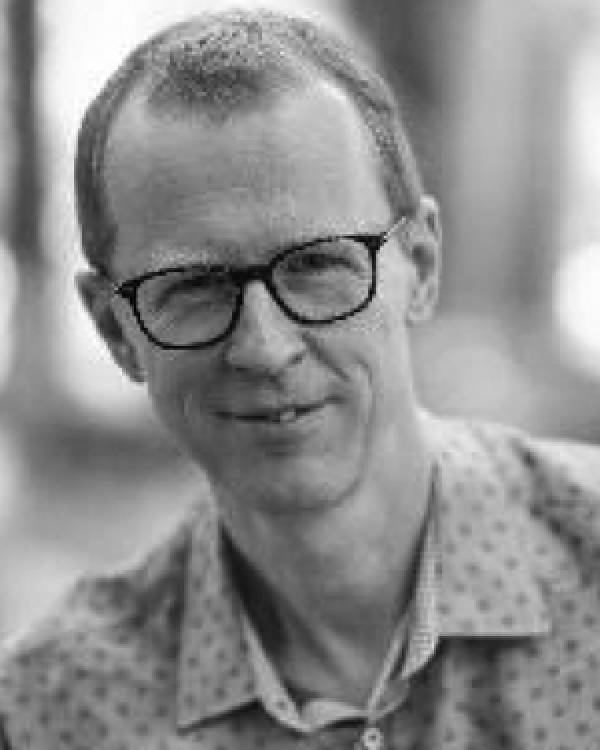}}]{Radom\'{i}r~M\v{e}ch}
heads the Procedural Imaging Group at Adobe Research. He leads a mix of research teams, senior individual contributors, and one invaluable technical artist. The researchers span areas of 2D and 3D design, image processing, modeling, natural media simulation, generative models (e.g., GANs), and HCI. His areas of research are procedural modeling, with a particular focus on interaction with procedural models and casual modeling, rendering, 3D printing, and image evaluation.
\end{IEEEbiography}

\begin{IEEEbiography}[{\includegraphics[width=1in,height=1.25in,clip,keepaspectratio]{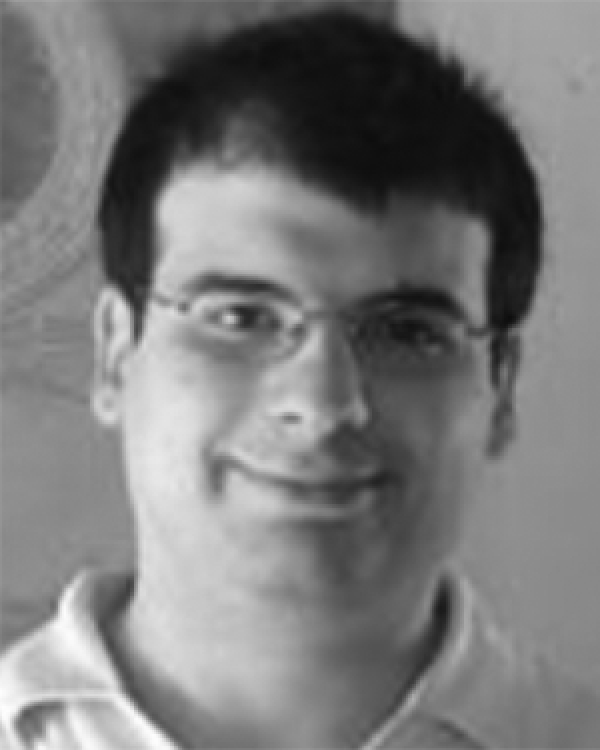}}]{Evangelos Kalogerakis}
is an Associate Professor at the University of Massachusetts Amherst, where he leads a group of students working on graphics and vision. His research deals with the development of AI/ML techniques that help people to easily create and process representations of the 3D visual world. He is particularly interested in algorithms that generate 3D models of objects, scenes, animations, and intelligently process 3D scans, geometric data, collections of shapes, images, and video. 
\end{IEEEbiography}
%H.~Kopka and P.~W. Daly, \emph{A Guide to {\LaTeX}}, 3rd~ed.\hskip 1em plus

%\end{thebibliography}


%\end{IEEEbiography}


%\newpage



%\vfill


\end{document}

\input{sec/X_supplementary}

\bibliographystyle{ACM-Reference-Format}
%\bibliography{sample-base}
\bibliography{macros,main}

%% file: preamble.tex
\usepackage{amsmath}
\usepackage{booktabs}
\usepackage{amssymb}
\usepackage{graphicx}
\usepackage{xcolor}
\usepackage{overpic}
\usepackage{enumitem} %< control spacing in itemize/enumerate/...
\usepackage{overpic} %< add raw math symbols to figures
\usepackage{color}
\usepackage{colortbl}
\usepackage{hyperref}
\definecolor{turquoise}{cmyk}{0.65,0,0.1,0.3}
\definecolor{purple}{rgb}{0.65,0,0.65}
\definecolor{dark_green}{rgb}{0, 0.5, 0}
\definecolor{orange}{rgb}{0.8, 0.6, 0.2}
\definecolor{red}{rgb}{0.8, 0.2, 0.2}
\definecolor{darkred}{rgb}{0.6, 0.1, 0.05}
\definecolor{blueish}{rgb}{0.0, 0.3, .6}
\definecolor{light_gray}{rgb}{0.7, 0.7, .7}
\definecolor{pink}{rgb}{1, 0, 1}
\definecolor{greyblue}{rgb}{0.25, 0.25, 1}
\usepackage{graphicx}
\usepackage[export]{adjustbox}

\DeclareMathOperator{\argmin}{arg\,min}

\newcommand{\eg}{\emph{e.g., }}

\newcommand{\etal}{\emph{et al}. }
\renewcommand{\paragraph}[1]{\vspace{1em}\noindent\textbf{#1}.}

\newcommand{\bc}{\mathbf{c}}

\newcommand{\bq}{\mathbf{q}}
\newcommand{\br}{\mathbf{r}}
\newcommand{\bs}{\mathbf{s}}
\newcommand{\bt}{\mathbf{t}}

\newcommand{\bx}{\mathbf{x}}

\newcommand{\bO}{\mathbf{O}}

\newcommand{\bQ}{\mathbf{Q}}

\newcommand{\bGamma}{\mbox{\boldmath$\Gamma$}}

\newcommand{\btheta}{\mbox{\boldmath$\theta$}}

\def\L{{\cal L}}

\def\R{{\cal R}}

\usepackage{blindtext}

%% file: sec/0_metadata.tex
\title{ANISE: Assembly-based Neural Implicit Surface rEconstruction}

\author{Dmitry~Petrov$^1$*,
        Matheus~Gadelha$^2$,
        Radom\'{i}r~M\v{e}ch$^2$,
        Evangelos~Kalogerakis$^1$,% <-this % stops a space
\IEEEcompsocitemizethanks{\IEEEcompsocthanksitem $^1$ University of Massachusetts, Amherst, MA, USA.\protect\\
$^2$ Adobe Research, San Jose, CA, USA\protect\\
*Corresponding author: to.dmitry.petrov@gmail.com\protect\\}% <-this % stops a space
}

%% file: sec/0_abstract.tex
\begin{abstract}
We present ANISE, a method that reconstructs a 3D~shape from partial observations (images or sparse point clouds) using a part-aware neural implicit shape representation. The shape is formulated as an assembly of neural implicit functions, each representing a different part instance. In contrast to previous approaches, the prediction of this representation proceeds in a coarse-to-fine manner. Our model first reconstructs a structural arrangement of the shape in the form of geometric transformations of its part instances. Conditioned on them, the model predicts part latent codes  encoding their surface geometry. Reconstructions can be obtained in two ways:  (i) by directly decoding the part latent codes to part implicit functions, then combining them into the final shape; or (ii) by using part latents to retrieve similar part instances in a part database and assembling them in a single shape.
We demonstrate that, when performing reconstruction by decoding part representations into implicit functions, our method achieves state-of-the-art part-aware reconstruction results from both images and sparse point clouds.
When reconstructing shapes by assembling parts retrieved from a dataset, our approach significantly outperforms traditional shape retrieval methods even when significantly restricting the database size.
We present our results in well-known sparse point cloud reconstruction and single-view reconstruction benchmarks.
\end{abstract}

%% file: sec/1_introduction.tex
\section{Introduction}
\label{sec:intro}

Neural implicit surface representations have gained increasing popularity for 3D reconstruction due to their ability to model  shapes as continuous surfaces  without topology restrictions. Despite their advantages, they also have  shortcomings. Most existing approaches are agnostic to  part structure. This contrasts the observation that most man-made objects are compositional i.e., they are made out of parts. As a result, they often end up producing surfaces with implausible part arrangements. In addition, their outputs are far from being controllable. For example, if users want to edit parts in the output shape they need to perform  post-processing segmentation steps, which can further amplify errors during editing.  

We propose \emph{ANISE}, a new \emph{part-aware} shape reconstruction method based on neural implicits. 
Specifically, given a partial shape observation (image or point cloud), it reconstructs shapes as a combination of parts, each with its own geometric representations. 
We propose shape reconstruction in two different manners: as a union of part implicit functions, or by retrieving parts in a reference database and assembling them into a final shape.
The former generates well-connected shapes without gaps or self-intersections whereas the later perfectly preserves the
quality of the parts in the reference set, providing more control over the reconstruction process.
Moreover, regardless of the way the parts can be assembled, our method can be trained and tested on shapes with varying number of parts i.e., we do not assume a fixed number of parts, and impose only an upper bound on their number.

Another key feature of our approach is that it explicitly \emph{disentangles geometry from structure}. Instead of forcing the network to learn how to predict the whole surface at once, we decompose this task into these two simpler learning problems representing a coarse-to-fine shape reconstruction. More specifically, our model first reconstructs a structural arrangement of the shape in the form of geometric transformations of its parts. Conditioned on them, the model predicts a latent code per part encoding its geometry. Our experiments demonstrate more accurate surface reconstructions compared to previous neural implicit methods that model the  surface as a whole. 

Another important property of ANISE is that it does not impose any particular order over the shape parts, which tends to be ambiguous in the case of 3D shapes. This contrasts the recent part-aware approach of PQ-Net~\cite{pqnet}, which assembles shapes using neural implicits for parts, but relying on a recurrent network trained on a sequence of parts in a particular order. 
When compared to PQ-Net in a single-view reconstruction setting, our method increases the F-Score by $16.8\%$. 

Finally, our part retrieval and assembly approach establishes a new state-of-the-art for retrieval-based single-view reconstruction. Compared to previous approaches, our method requires shape databases \emph{an order of magnitude smaller} to achieve the same reconstruction quality. This is due to the fact that previous state-of-the-art retrieval-based approaches first retrieve full shapes and then deforming them, thus cannot replace parts or adjust the structure of the retrieved shapes. ANISE performs retrieval on the part level, resulting in assemblies that capture the underlying structure and parts of a shape much more effectively.

We summarize our contributions as the following:
\begin{itemize}
  \item A coarse-to-fine method to assemble a set of neural implicits. Unlike other part-based neural assembly methods, our method first predicts part transformations (cuboid approximations to parts), then predicts part geometry. This leads to superior part reconstruction quality compared to other part-aware reconstruction methods.
  \item A network architecture for assembling neural implicits, which can be fine-tuned end-to-end using supervision from full shapes instead of individual parts. This network significantly outperforms state-of-the-art methods for part-aware reconstruction (Tables ~\ref{tab:rgb_reconstruction} and ~\ref{tab:sota}).
  \item A simple yet surprisingly effective method for retrieving parts and assembling them to create new shapes matching a partial shape observation (image or point cloud). This part retrieval and assembly approach (PR\&A) yields significant improvements over state-of-the-art shape retrieval and deformation counterparts (Table~\ref{tab:rgb_reconstruction}). %
\end{itemize}

%% file: sec/2_related.tex
\input{fig/overview} %< bloody latex and its heuristics for figure placement

\section{Related work}
\label{sec:related}

3D shape reconstruction either from point clouds \cite{kazhdan2006poisson,hanocka2020point2mesh,erler2020points2surf,peng2021shape,williams2019deep,jiang2020shapeflow} or single RGB images \cite{prasad2006single,tatarchenko2019single,groueix2018papier, tatarchenko2017octree,richter2018matryoshka,tulsiani2017multi,wu2018learning} is a well studied area in computer vision. Here we discuss the most related works for shape reconstruction either with implicit functions or with part-aware approaches.  

\paragraph{Implicit function learning for 3D reconstruction}  
In recent years, several approaches have proposed employing implicit functions (or coordinate-based architectures) to represent 3D shapes~\cite{occnet,imnet,deepsdf}.
These methods represent the surface of an object as the level-set of a function parametrized by a neural network. This function can be either seen as a binary classifier for occupied (or empty) regions of space~\cite{occnet,imnet} or a signed distance field (SDF)~\cite{deepsdf,disn}. Other techniques tried to improve the quality of the shapes represented by those functions through localized 
representations~\cite{convoccnet,deeplocalshapes,ldif,localgrid,disn,neuralparts,li2021d2im}.
Similarly, other methods generate parameters of a fixed, previously defined implicit functions, either by predicting planes to generate convex polyhedra~\cite{bspnet,cvxnet}, directly estimating the weights of a smaller MLP~\cite{deepmetafunctionals} or a set
of radial basis functions~\cite{sif}. Despite their quality, none of these models is designed to reason about meaningful shape parts; \eg a particular component will not always represent the full leg of a chair, or the top of a table, or an engine of an airplane, and so on. Our work aims to describe shapes as an assembly of implicit functions parameterized by neural networks, where every function is responsible for generating a  meaningful part. This allows users to perform part-aware shape operations, like editing specific elements and reconstructing shapes by assembling objects from pre-determined parts.

\paragraph{Assembly-based shape modeling}
The seminal work by Funkhouser \etal~\cite{modelingbyexample} employed user input for interactive segmentation
and hand-crafted features for part exploration and shape composition.
This synthesis-by-examples approach was later followed by techniques based on probabilistic models~\cite{componentsynthesis,probreasoning,Chaudhuri:2013:ACC}, evolutionary techniques~\cite{fitdiverse}, 3D shape programs \cite{tian2018learning}, among others.

Lately, several datasets with part annotations~\cite{shapenet,partnet} have allowed the development of less constrained and more automatic approaches
powered by deep neural architectures.
A number of them are focused on generating structural information~\cite{3dprnn,genshapehandles,structurenet} then rely on  simplified part geometry models like \cite{3dprnn, genshapehandles}. 

Several deep learning methods have been proposed for assembly-based modeling over the last years. Huang \etal introduced a deep learning method to generate shapes as an assembly of template parts modeled as sparse point sets \cite{Huang15BSM}. GRASS introduced  a recursive neural network that represents shapes as binary part hierarchies whose leaves are decoded into  low-resolution ($32^3$) volumetric part representations~\cite{grass}. StructureNet proposes a similar hierarchical model encoding structural part relationships, yet parts are represented as sparse point clouds \cite{structurenet}. MRGAN \cite{gal2021mrgan} introduced a generative adversarial network for disentangled generation of semantic parts through self-supervised  point-cloud generators. 
The coarse shape representations of the above approaches simplify their tree-structured representation encoding and decoding, yet severely limits the quality of reconstructed parts. Niu \etal~\cite{niu2018im2struct} introduced single view reconstruction model through recovery of cuboid hierarchies. 
Dubrovina \etal~\cite{Dubrovina_2019_ICCV} proposed assembly-based neural shape modeling by first predicting voxelized parts, then part transformations through a spatial transformer.  Li \etal~\cite{li2020learning} proposed another voxel-based method for generating parts, then assembling them through regressed transformations. Bokhovkin \etal~\cite{bokhovkin2021towards} proposed part-based approach for semantic part completion from RGB-D scene scans. Similarly to GRASS, all these approaches rely on low-resolution voxel grids. Their high memory consumption further limits their ability to reconstruct high-quality shapes.
Incorporating more fine-grained part representations, such as implicit functions, in structure-aware models is not straightforward. Challenges include mapping the input partial scans or single views into compact part-based representations that effectively encode the whole implicit assembly, then decoding and combining these representations into a coherent shape with fine-grained geometry. Another limitation applicable to Dubrovina \etal~\cite{Dubrovina_2019_ICCV} and Li \etal~\cite{li2020learning} is that they group part instances of the same \emph{semantic} label (e.g., all chair legs) under the same transformation. ANISE, on the other hand, builds on top of implicit function representations (leading to more detailed shapes) and does not require semantic labels, producing one transformation for each part instance; e.g., each chair leg instead of a group of legs.

\paragraph{Neural implicit assemblies} Instead of relying on voxel representations or points clouds, a number of recent approaches were introduced to model shapes as a collection of impicits. Genova \etal \cite{sif} introduced a shape representation through a large collection of localized axis-aligned anisotropic 3D gaussians, called Structured Implicit Functions -- SIFs. In a follow-up work, Genova \etal \cite{ldif} introduced local deep implicit functions (LDIF) that were arranged and blended into shapes similarly to SIF. Deng \etal \cite{deng2020cvxnet} introduced CvxNet: shape approximation through a set approximately convex implicit functions. Paschalidou \etal \cite{neuralparts} introduced Neural Parts that models shapes as a set of template parts obtained through homeomorphic deformations of spheres.
Hertz \etal \cite{hertz2022spaghetti} proposed SPAGHETTI that represents shapes as Gaussian mixture of part codes processed through a transformer model and assembled with predicted part transformations.  All these methods offer much higher-quality shape generation compared to previous voxel- and point-based variants. They also have the advantage of being trained in an unsupervised or self-supervised manner. However, their main limitation is that their implicit assemblies are not directly interpretable i.e., it is hard for users to replace or modidy SIFs, LDIFs, or convexes since they do not always correspond to well-defined parts (e.g., semantic parts).

More closely related to our work is the assembly-based neural implicit method by Wu \etal~(PQ-Net). Like our method, PQ-Net generates well-defined parts through neural implicits. In contrast to ANISE, their part generation  is performed sequentially.  Our approach treats parts as sets instead of sequences, which simplifies the generation procedure –- generating shapes with several parts accounts a single forward pass in our model. 
More importantly, we are able to obtain supervision from fully assembled shapes to fine-tune our
model, leading to further qualitative and quantitative improvements. In contrast to PQ-Net, we also employ a coarse-to-fine generative procedure, which we demonstrate to be highly beneficial (Section~\ref{res:ablation}).
When combined, all these characteristics lead to a significant improvement over PQ-Net in shape reconstruction from point clouds and single RGB images (Tables~\ref{tab:rgb_reconstruction}~and~\ref{tab:sota}).

\paragraph{Shape retrieval} Another line of work related to ours consists of reconstructing shapes by retrieving similar ones from a database.
The similarity measure is usually performed in some embedding space that encodes data of multiple modalities (\eg  point clouds, images, meshes),
allowing the retrieval to be performed conditioned on various sources of data~\cite{imagepurification,productnet}.
When compared to reconstruction approaches, retrieval counterparts offer the advantage of yielding high-quality shapes -- stock models
from shape repositories.
A characteristic example of such approaches is ComplementMe \cite{sung_complementme_2017}, a part suggestion approach for incomplete point shapes based on probabilistic part retrieval operating on point cloud embeddings. Its limitation is  that the retrieved result might not be a good match for the data queried by the user.
Previous works address this issue by exhaustively deforming the retrieved shape until it matches the target data~\cite{searchclassify} or by
retrieving parametric models that can be better adjusted to the user query~\cite{retrievalparametric}.
More recently, Uy \etal proposed a joint retrieval and deformation model that is capable of deforming specific shape parts to enhance
its similarity to the target data~\cite{jldr}.
However, despite being able to deform parts independently, the model is still constrained to the retrieved shape.
We derive a simple yet surprisingly effective part retrieval and assembly (PR\&A) approach -- use part latents and part transformations predicted by our model to retrieve similar parts in a database and correctly position assemble them.
In other words, given a database of shapes and a partial shape observation (image or point cloud) our model is able to create a new shape by \emph{combining parts from different shapes} in the database which closely match the partial observation.
We show that our method significantly outperforms state-of-the-art retrieval and deformation approaches and is even able
to do so by using shape databases with a fraction of the original size (Figure~\ref{fig:cond_retrieval_vs_size}).

%% file: fig/overview.tex
\begin{figure*}
\begin{center}
\includegraphics[width=.99\textwidth]{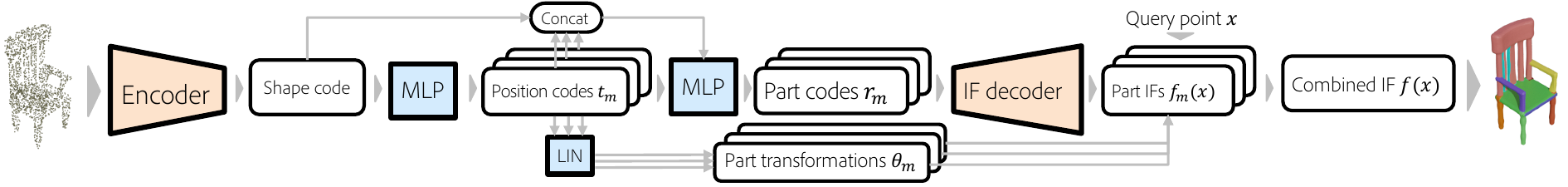}
\end{center}
%\vspace{-3mm}
\caption{
\textbf{Network architecture.} ANISE takes a point cloud or RGB image as an input and outputs a neural implicit function evaluated at query point $\bx$ in 3D space. Our architecture starts with an input encoder which produces a global shape code. Based on this code, the network produces part transformation codes $\bt_m$. From them and the shape code, the network predicts a set of part codes $\br_m$. These part codes are passed to an implicit function decoder along with the query points $\bx$, which provides implicit functions for each part. Using the predicted transformations $\btheta_m$, these functions are transformed and combined into a complete shape implicit function.}
%The Output of ANISE can be edited on part level and constrained to retrieval-based reconstruction: model can reconstruct input shape using parts from reference database of arbitrary size. }
\label{fig:architecture}
\end{figure*}

%% file: sec/3_method.tex
\section{Method}

An overview of our approach is depicted in Figure \ref{fig:architecture}. It consists of three modules. The first module  predicts a coarse part arrangement in the form of part transformations (``part transformation'' module, Section \ref{sec:structure}). 
Then, conditioned on these part transformations, the next module predicts an implicit function representing the actual geometry
(``part geometry module'', Section \ref{sec:geometry}). 
The last module transforms each part implicit functions according to the predicted transformations and combines them into a single output implicit function $f(\bx)$ (``shape assembly module'',  Section \ref{sec:implicit}). 
This process can be thought as a coarse-to-fine surface prediction procedure, where shape is first predicted coarsely in the form of part transformations (\eg placement, size, deformations, etc) to be later refined into fine geometry. 

The network is trained with objectives that encourage both correct structure prediction and part geometry prediction based on a repository of segmented shapes
(Section \ref{sec:training}). Part labels are not required i.e., the segmentation does not need to be ``semantic'' but should be consistent across shapes. The whole architecture is trained to produce output implicit functions close to the ground-truth. 

\subsection{Part transformation module}
\label{sec:structure}

This module takes as input observation $\bO$, in the form of a single RGB image or sparse point cloud, and encodes it into a 256-dimensional shape code $\bs$. 
This shape representation is then decoded by a MLP that produces $M$ transformation codes. More specifically, the MLP implements a function \mbox{$g(\bs)=\{\bt_m\}_{m=1}^M$}, where $\bt_m$ is a $128$-dimensional feature representation of the $m^{th}$ part transformation.  
Each representation is then decoded into actual transformation parameters $\btheta_m$.  
In our experiments, we show that very simple transformations (just positioning and scale) are enough to produce shapes competitive
to current state of the art approaches. From this aspect, these transformations can be thought of as predicting a cuboid -- a coarse representation of each part.

It is important to note that the number $M$ represents an upper bound of parts for the output shape ($M=10$ in our experiments for all shape categories). 
During training, only a subset of the predicted tranformations is matched  with ground-truth ones, as discussed in Section \ref{sec:training}. 
We note that we tried predicting an extra binary
output representing the existence of a part, but it did not improve results. 
In practice, the network learns not to make use of extra parts i.e., they are decoded to implicits without zero crossings, or in other words, they are not filled with surface geometry, which counter-effects the need for such output parameter. 
We also note that the order of the predicted transformations does not matter since the subsequent modules of our architecture are invariant to part order. 
During training, the matching between predicted and training transformations is also order-insensitive.

\subsection{Part geometry module}
\label{sec:geometry}

This module first takes as input the shape representation $\bs$ and each transformation representation $\bt_m$ 
and predicts a per-part geometric feature representation $\br_m$. 
This is done through $M$ shared MLPs, each implementing the function \mbox{$\br_m = h( \bt_m, \bs)$}. 
Then, an implicit surface decoder decodes each part's feature representation $\br_m$  to its implicit  function $f_m(\bx, \br_m)$. 
The zero level set of the function represents the surface of the part $m$.
The module is trained to produce parts that are centered at the origin and scaled to have unit scale across axis with largest scale. This helps the decoder to focus on the geometry of parts, since their location and size are already captured in the part transformation module. 

As discussed in Section \ref{sec:training}, we found useful to decompose the prediction of the part implicit function into two stages (part feature representation prediction, then part implicit prediction) because this enables the ability to pre-train the implicit surface decoder that results in better performance. We also note that we tried other variants, such as predicting the part implicit functions without conditioning on part transformations, or skipping the part feature prediction part, yet these resulted in worse performance, as discussed in our experiments (see Table~\ref{tab:ablations}). 
\input{fig/supplementary/preprocessing_parts}

\subsection{Shape assembly module}
\label{sec:implicit}

This module combines the part implicit functions into a single shape.
We present two different approaches for accomplishing this task: part retrieval and assmebly (PR\&A), and implicit function assembly.
The former retrieves parts from a database and assembles them using the predicted transformations, prefectly
preserving the quality of the assets on the database.
The later allows the creation of more plausible connections between parts -- since the process is fully differentiable we are able to fine-tune the assembly on complete shapes and improve part connections with the drawback of losing the exact
correspondence to parts on a database.

\paragraph{Part retrieval \& assembly (PR\&A)}
For this approach, instead of using the implicit functions to generate geometry directly, we use the part latents ($\br \in \bGamma$) to query similar parts in a part database (also referred to as \emph{reference set}).
Let $\bQ=\{\bq_{k}\}_{k=1}^K$ be our part database, where $K$ is their total number parts and $\bq_k$ is the geometric
representation of the $k$-th part.
Let $\phi(\bq_k)$ be a function that projects the geometric part representation into $\bGamma$.
Then, we can use the predicted part presentations $\br_m$ to query a part $\hat{\bq}_m$ such that $\hat{\bq}_m = \argmin_{\bq_i \in \bQ} ||\phi(\bq_i)-\br_m||_2$.
The final shape can be assembled by transforming each part $\hat{\bq}_m$ according to its predicted transformation $\btheta_m$.
This approach has a couple of advantages.
First, it is agnostic to the geometric representation -- it can be applied as long as there is a way of computing a latent representation from its geometric one. This means that the final shape will keep the details and quality of the retrieved parts.
Second, this can be used to provide the user of the system more control over the assembled shape by simply curating which parts will be present in the reference set (see Figure~\ref{fig:cond_rec2}).
A very important drawback of this approach comes from its first advantage.
Since parts exactly correspond to the ones in the database, the connection between them might not be plausible.
We propose to tradeoff the quality of the individual parts with the plausibility of the whole shape by also developing and alternative \emph{implicit function assembly}, described in the next paragraph.

\paragraph{Implicit function assembly}
We can also assemble complete shapes by combining the implicit function representation of the individual parts in a differentiable manner.
For SDF, the geometric union of implicit functions can be expressed using a minimum operation \cite{Museth02}, after transforming the part according to the decoded transformation parameters $\btheta_m$. 
%re-positioning and re-scaling them based on the predicted box positions $\{\bc_m\}_{m=1}^M$ and scales $\{d_m\}_{m=1}^M$ 
\begin{equation}
f(\bx) = \min_m  T(f_m;\btheta_m)(\bx) \label{eq:combine_implicits}
\end{equation}
where $T$ is a functional that transforms the implicit function $f_m$ according to $\btheta_m$. The operation can be thought of as min pooling over $M$ part implicit functions modified with predicted transformations. Backpropagation through this operation enables learning part implicits such that their combination produces a coherent shape as a whole.
For example, in case of a transformation $\btheta_m$ consisting of part translation $\bc_m$ and uniform scaling $\alpha_m$, implemented in our experiments, this formula can be written as \cite{Museth02,Reiner11}:
\begin{equation}
f(\bx) = \min_m \alpha_m f_m\Big(\frac{\bx - \bc_m}{\alpha_m}\Big) 
\label{eq:combine_implicits_simple}
\end{equation}
where query point $\bx$ is transformed from shape centered space to a part centered and scaled space, thus allowing the evaluation of the part implicit function.
In practice, we show that even this simple part transformation model is enough to
achieve high reconstruction quality, surpassing state-of-the-art part-aware reconstruction
approaches in various metrics.
The scaling factor $\alpha_m$ in front of the implicit function is important to ensure that the scale of  implicit functions hold after the union (e.g., in the case of scaling signed distance functions, distances should also be scaled \cite{Museth02,Reiner11}).

\section{Training}
\label{sec:training}

We now describe the training procedure for ANISE. One possibility to train our architecture is to just provide a single supervisory signal in the form of target implicit function values obtained from ground-truth shapes. More specifically, given  point samples $\{\bx_i^{(p)}\}$ from $\R^3$, where $i$ is a point sample index and $p$ is a training shape index, a common procedure in  neural implicit fitting approaches \cite{disn,deepsdf,deeplocalshapes} is to measure the signed  distance $s_i^{(p)}$ of each point to the training shape's surface, then minimize a loss between predicted implicit function values and signed distances:
$L_{shape} = \sum_{i,p} \L \big( f(\bx_i^{(p)}), s_i^{(p)} \big)$, where $\L$ is set to L1 loss or clamped L1 loss \cite{disn}. 
Our architecture is fully differentiable, thus, this approach is entirely possible in our method.
Unfortunately, it fails to deliver successful results: we observed that a single part is produced enclosing the whole training shape, while the rest collapse to negligible surfaces.
As a result, no structure is captured from the network, and the  performance of the method ends up being  similar to structure-agnostic neural implicit methods (see supplementary material).   

We instead found that the best performance is achieved when we make use of shapes segmented into parts and then provide supervision in the form of part implicit functions, part transformations, and part feature representations. Thus, we follow a multi-stage training procedure described in the following paragraphs. First, we pre-train the part implicit decoder used in the part geometry module (Sec. \ref{sec:geometry}). Second, we train the part arrangement and part geometry module together using supervision from part transformations and its feature representations. Finally, we fine-tune the whole network end-to-end using supervision from the implicit functions obtained from the complete shapes. The input to our training procedure is a set of shapes segmented into parts. We assume that the input shapes are consistently oriented. In our experiments, we make use of PartNet~\cite{partnet}, a popular repository of  3D\ meshes with part segmentations. Even though we make use of the shape segmentations, our training \emph{does not} use their \emph{semantic} labels.

\input{fig/comparison_svr_disn.tex}

\input{fig/comparison_svr_parts}

\subsection{Pre-training stage}
\label{sec:pretraining}
Our first stage involves pre-training the part implicit surface decoder $f_m$. The decoder is conditioned on the part geometric feature representation, thus, we need to generate those first for our training shapes. To achieve this, we train an encoder-decoder network that takes as input individual parts from the segmented training shapes, encodes them to a $256$-dimensional descriptor in the bottleneck, then decodes them to part implicit functions. More specifically, given each training part, we center it to the origin and scale it uniformly such that the largest dimension of bounding box is one. In addition, since the mesh of the part  often has holes at the part boundaries after excluding the rest of the shape, we reconstruct its mesh to a watertight version so that valid signed distance functions can be computed (more details in the supplementary material). Then we  sample $10^5$ points inside of bounding box of each part.

The encoder has the form of a PointNet \cite{qi2017pointnet} with skip connections that processes each part's points. The output of the encoder is the feature representation of each part, denoted as $\hat \br_n^{(p)}$, where $n$ is a training part index for a shape $p$. The decoder has the same architecture with our implicit surface decoder $f_m$ in the part geometry module (Section \ref{sec:geometry}). During training we add noise noise to part codes to make sure that representation space is smooth, so latent codes can be smoothly interpolated and the distance in this space corresponds to geometric similarity of parts.

\allowbreak
The network is trained to minimize the L1 loss 
$$
L_{part} = \sum_{i,p} ||~\big(f_m(\bx_i^{(p)})~- s_i^{(p)}||_1,
$$
where $s_i^{(p)}$ is the target signed distance function value of the sample point $\bx_i^{(p)} \in \R^3$ to the mesh part surface (details about sampling are provided in the supplementary material). The result of this stage is both a trained part implicit surface decoder $f_m$ and geometry-encoding part features $\hat \br_n^{(p)}$ used in the subsequent stages.

\subsection{Main training stage} 
We proceed with training our network using supervision from training part transformations and part feature representations. 
Specifically, for each training shape and part, we first extract an axis-aligned transformation with center position and scale parameters $\hat \btheta_n^{(p)} = [\hat \bc_n^{(p)}, \hat d_n^{(p)}]$. We use isotropic scaling because it is compatible with SDF-based part assembly (see \ref{sec:implicit}).

We use two losses to train our network. The first loss penalizes discrepancies between predicted transformation parameters and ground-truth ones. Specifically, given each predicted transformation $m$ with parameters $\btheta_m^{(p)} = [\bc_m^{(p)}, d_m^{(p)}]$ for each training shape $p$, our loss is expressed as follows:
\begin{equation}
\mathcal{L}_{\text{1}}\!\!=\!\!\frac{1}{N_p} \sum_{n=1}^{N_p} \min_m || \hat \btheta_n^{(p)} - \btheta_m^{(p)} ||_2  \\ + \frac{1}{M}\sum_m^M \min_n || \hat \btheta_n^{(p)} - \btheta_m^{(p)} ||_2
\end{equation}
where $N_p$ is the number of training parts for shape $p$. The first term finds the most similar predicted transformation to each ground truth one and penalizes their differences. It can be thought of as maximizing recall in the matching and ensures that model covers all of ground truth transformations.
The second term builds the opposite association and ensures precision of predictions. This bidirectional distance offers the best performance. We note that the above matching is invariant to the order of predicted parts thus allowing for non-sequential set prediction.

\allowbreak
The second loss penalizes discrepancies between predicted part feature representations $\br_m^{(p)}$ and  training ones~$\hat \br_n^{(p)}$. In this manner the part geometry module is encouraged to predict part representations that are close to the ones produced during pre-training and that can be decoded to implicit functions according to our pretrained implicit surface decoder. Specifically, given each predicted transformation, we find its most similar training one in terms of their transformation parameters: $n' = \argmin_n ||\hat \btheta_n^{(p)}~-~\btheta_m^{(p)}~||_2$. Then we penalize differences in their part feature representations:
\begin{equation}
\mathcal{L}_{\text{2}}\! = \!\frac{1}{M} \sum_{m=1}^{M}  || \hat \br_{n'}^{(p)} - \br_m^{(p)} ||_2  
\end{equation}
In this manner, the network is encouraged to predict part feature representations that are consistent with the ones produced during pretraining. 
The network is trained using a weighted sum of the losses $\L_1 + \lambda \L_2$, where $\lambda$ is a hyperparameter that manages a trade-off between global part structure encoded in $\bt_m$ and part geometries encoded in $\br$. We set it to $0.02$ in all our experiments. 

In the final stage of our training (\textbf{fine-tuning}), the network is further fine-tuned to predict implicit functions that are as close as possible to the training ones for whole shapes. Here we minimize  $L_{shape} = \sum_{i,p} || \big( f_m(\bx_i^{(p)}) - s_i^{(p)} ||_1$, where $s_i^{(p)}$ is the target signed distance function values  of sample point $\bx_i^{(p)} \in \R^3$ to the  surface of the training shape~$p$.

%% file: fig/supplementary/preprocessing_parts.tex
\begin{figure*}[!h]
\begin{center}
% \begin{overpic} 
% [width=\linewidth]
% {example-image-a}
% \end{overpic}
\setlength\tabcolsep{-8pt}
\begin{tabular}{c@{\hskip 5pt}llllllll}
\rotatebox[origin=c]{90}{Watertight mesh} & \includegraphics[width=.15\linewidth,valign=m]{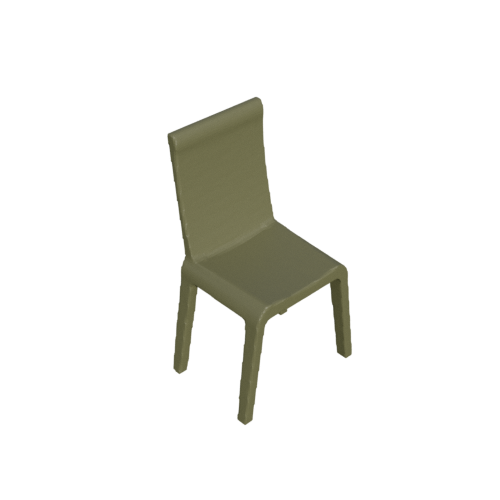} & \includegraphics[width=.15\linewidth,valign=m]{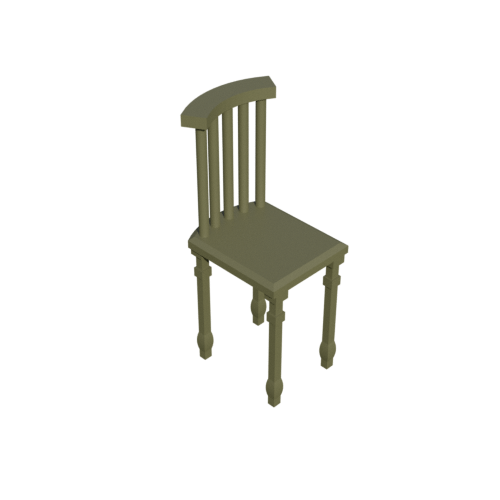} & \includegraphics[width=.15\linewidth,valign=m]{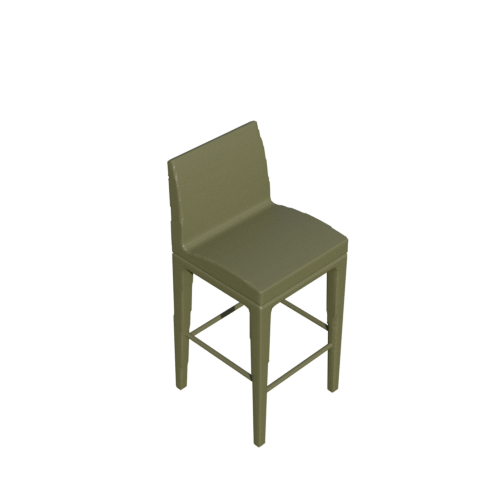} & 
\includegraphics[width=.15\linewidth,valign=m]{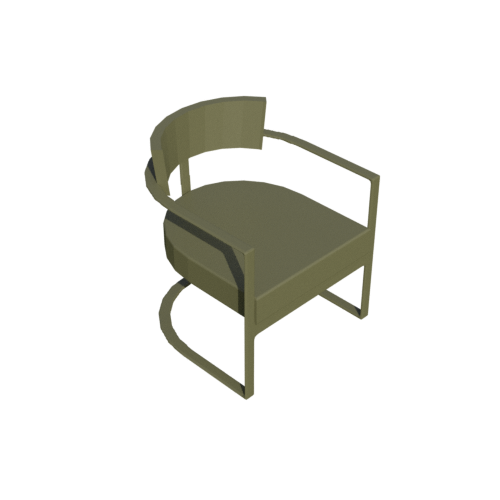} &
\includegraphics[width=.15\linewidth,valign=m]{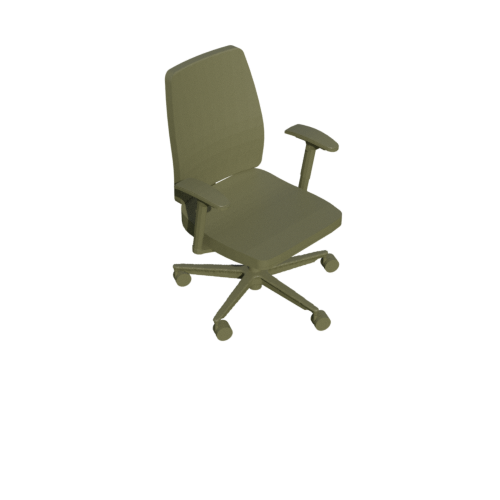} &
\includegraphics[width=.15\linewidth,valign=m]{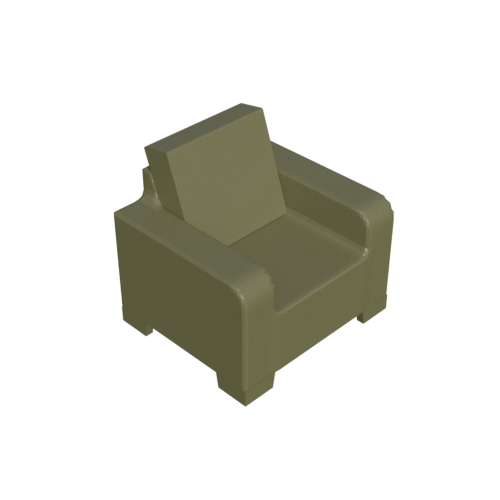} &
\includegraphics[width=.15\linewidth,valign=m]{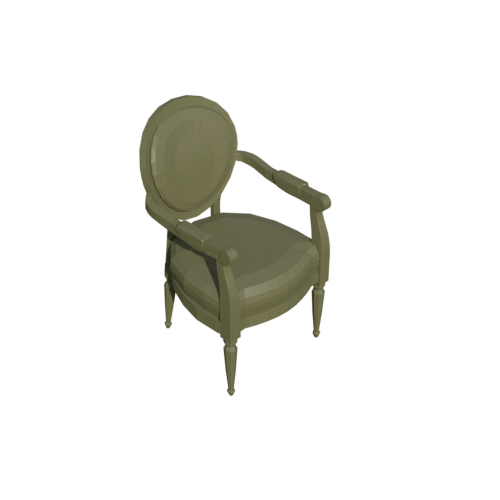} &
\includegraphics[width=.15\linewidth,valign=m]{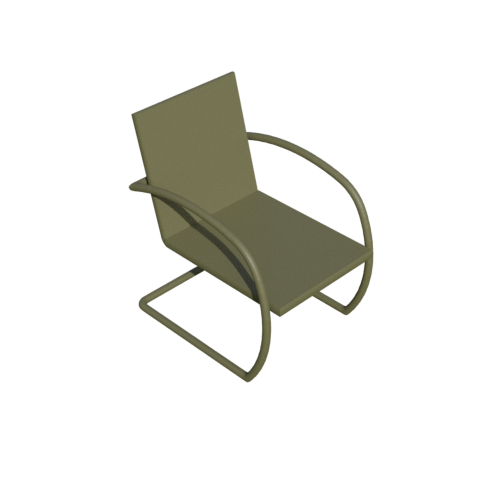}\\
\rotatebox[origin=c]{90}{Original parts} & \includegraphics[width=.15\linewidth,valign=m]{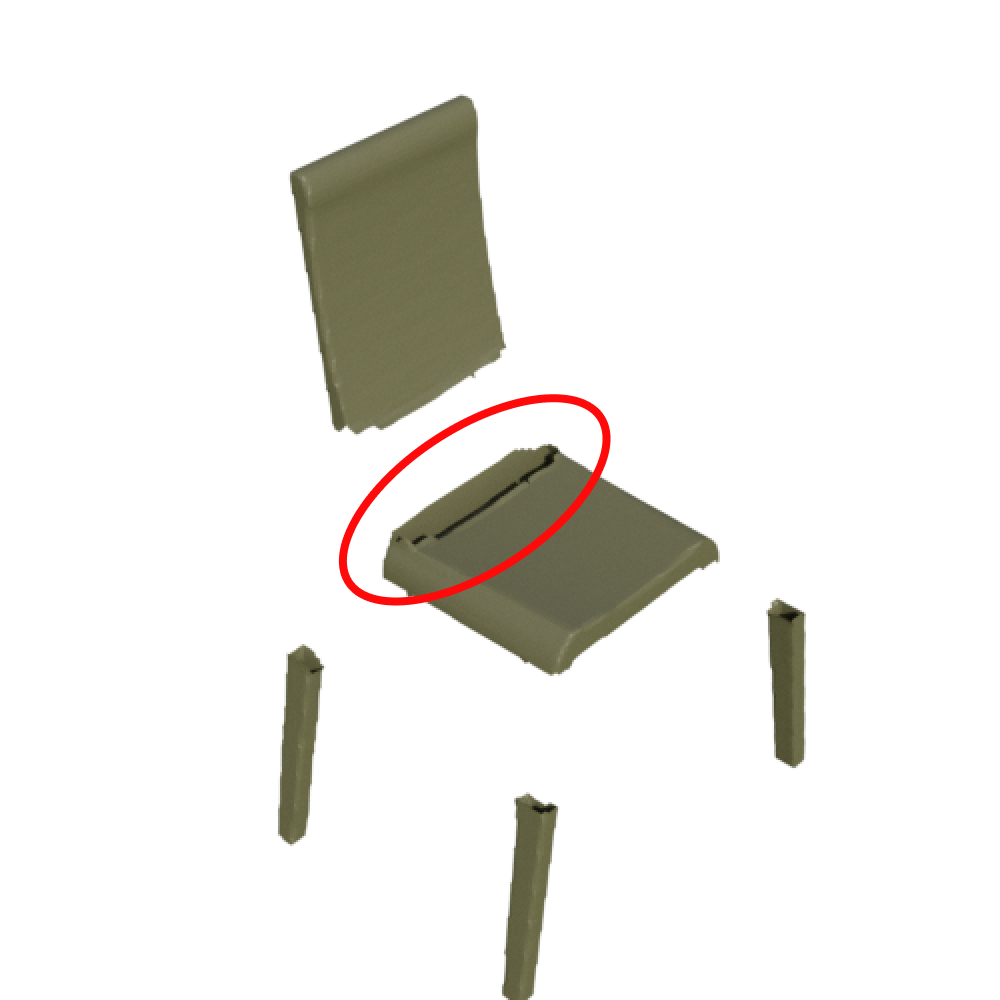} & \includegraphics[width=.15\linewidth,valign=m]{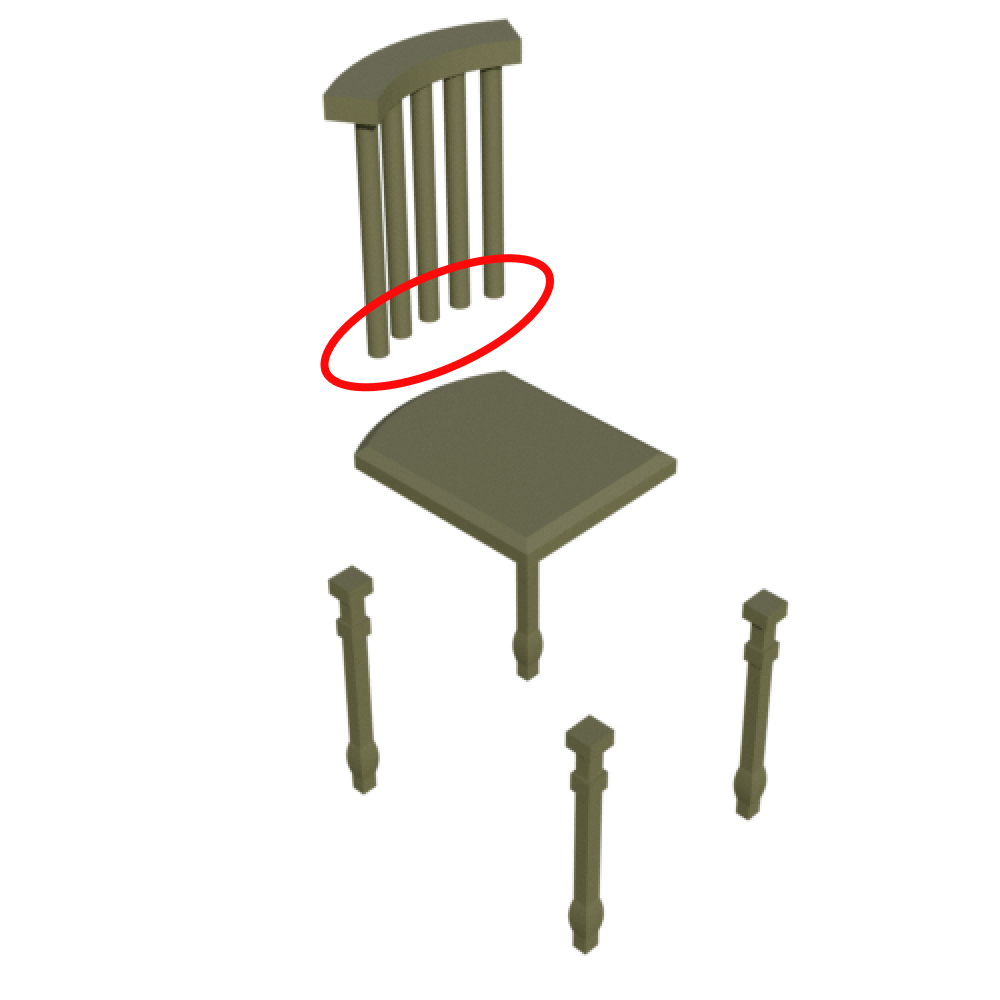} & \includegraphics[width=.15\linewidth,valign=m]{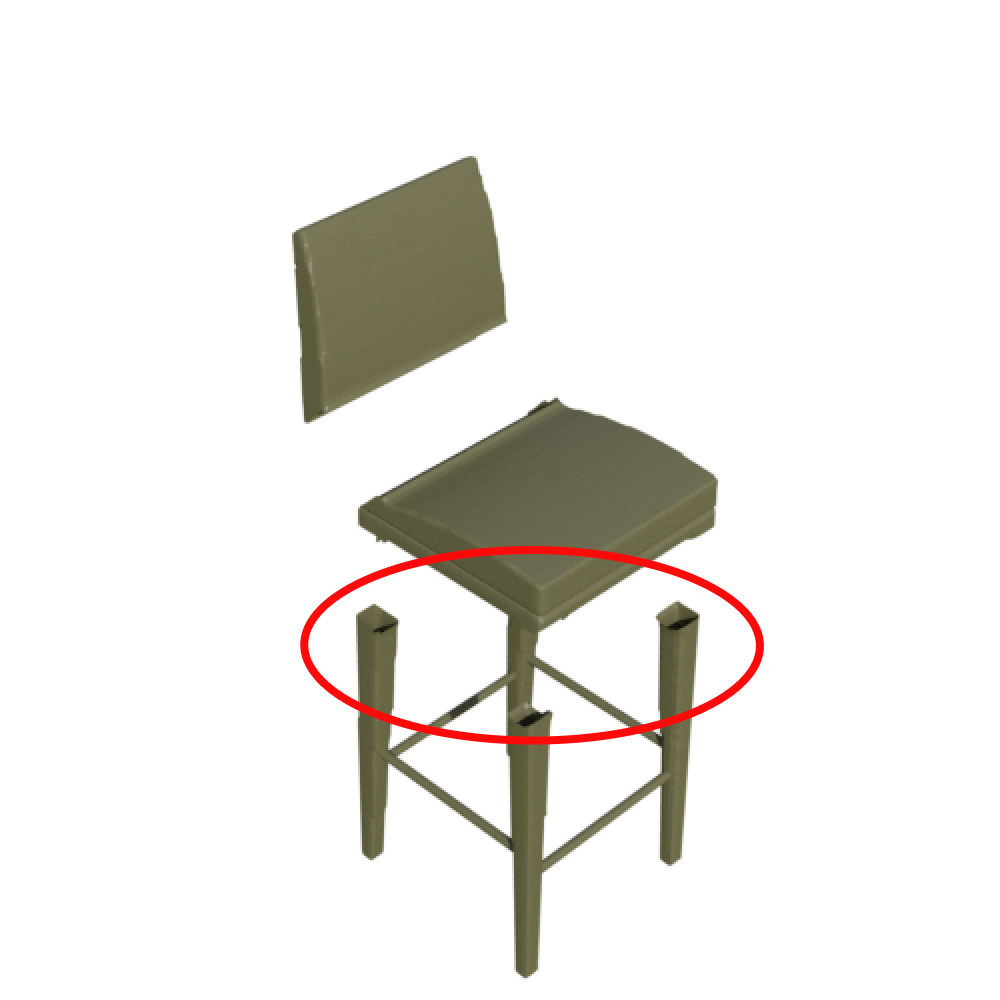} & 
\includegraphics[width=.15\linewidth,valign=m]{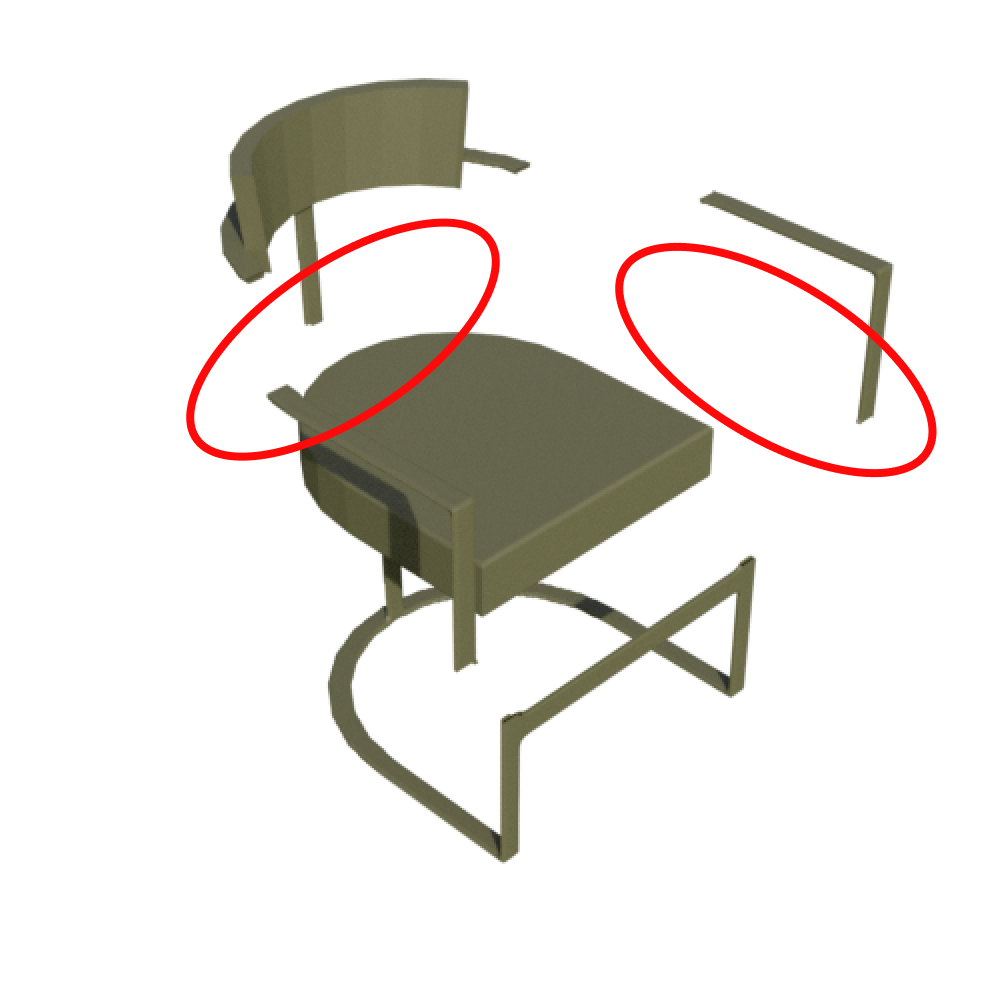} &
\includegraphics[width=.15\linewidth,valign=m]{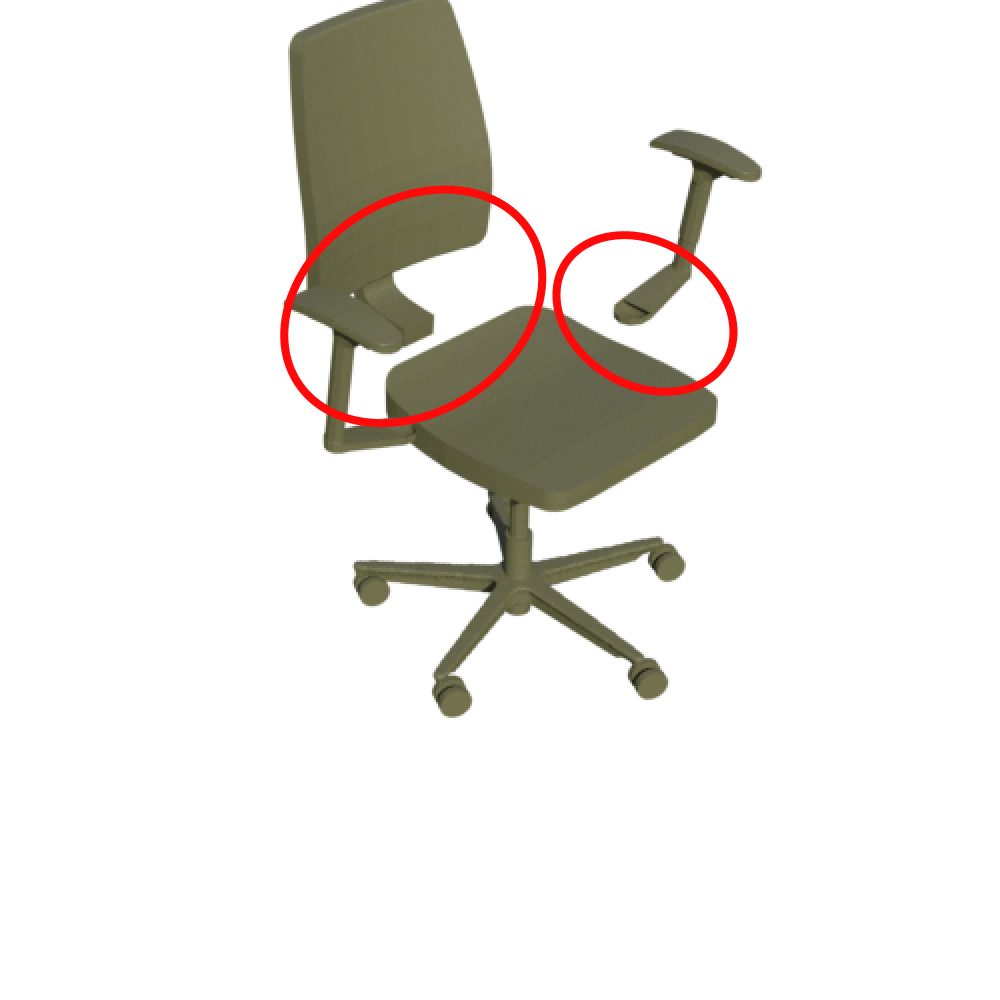} &
\includegraphics[width=.15\linewidth,valign=m]{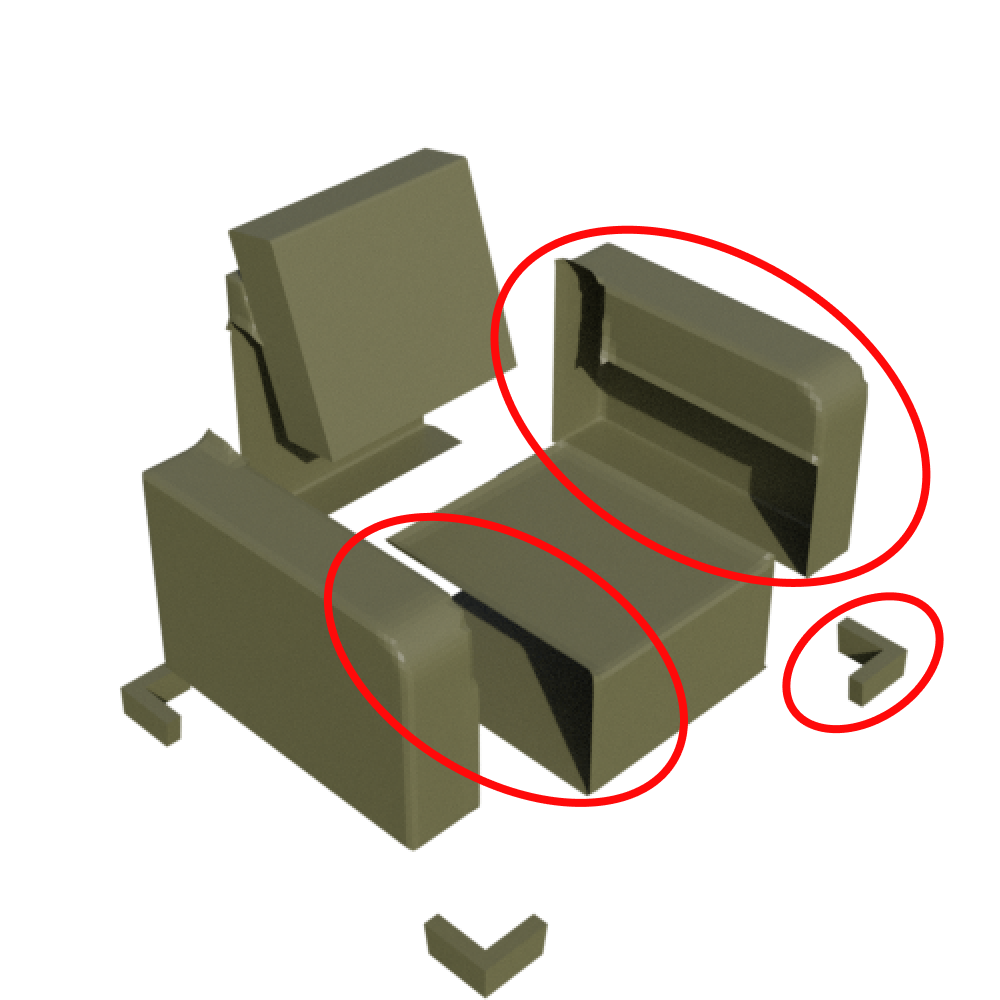} &
\includegraphics[width=.15\linewidth,valign=m]{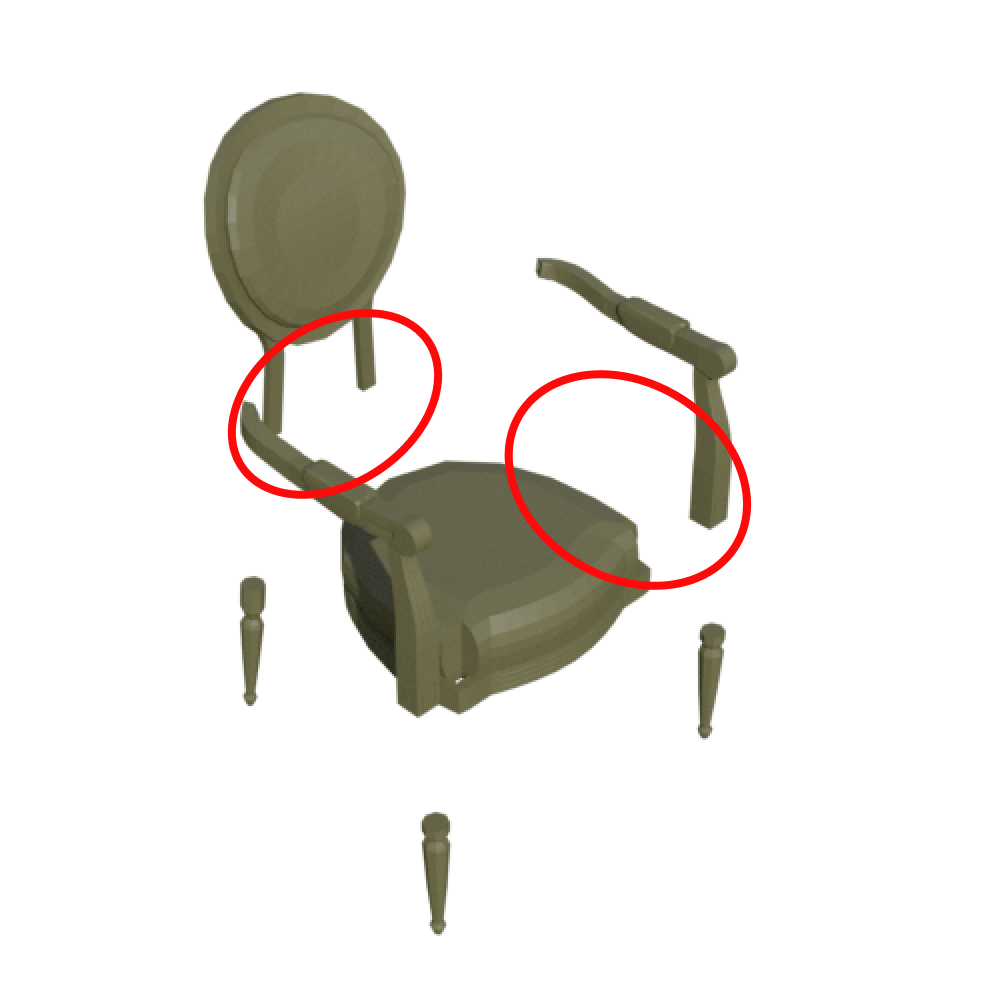} &
\includegraphics[width=.15\linewidth,valign=m]{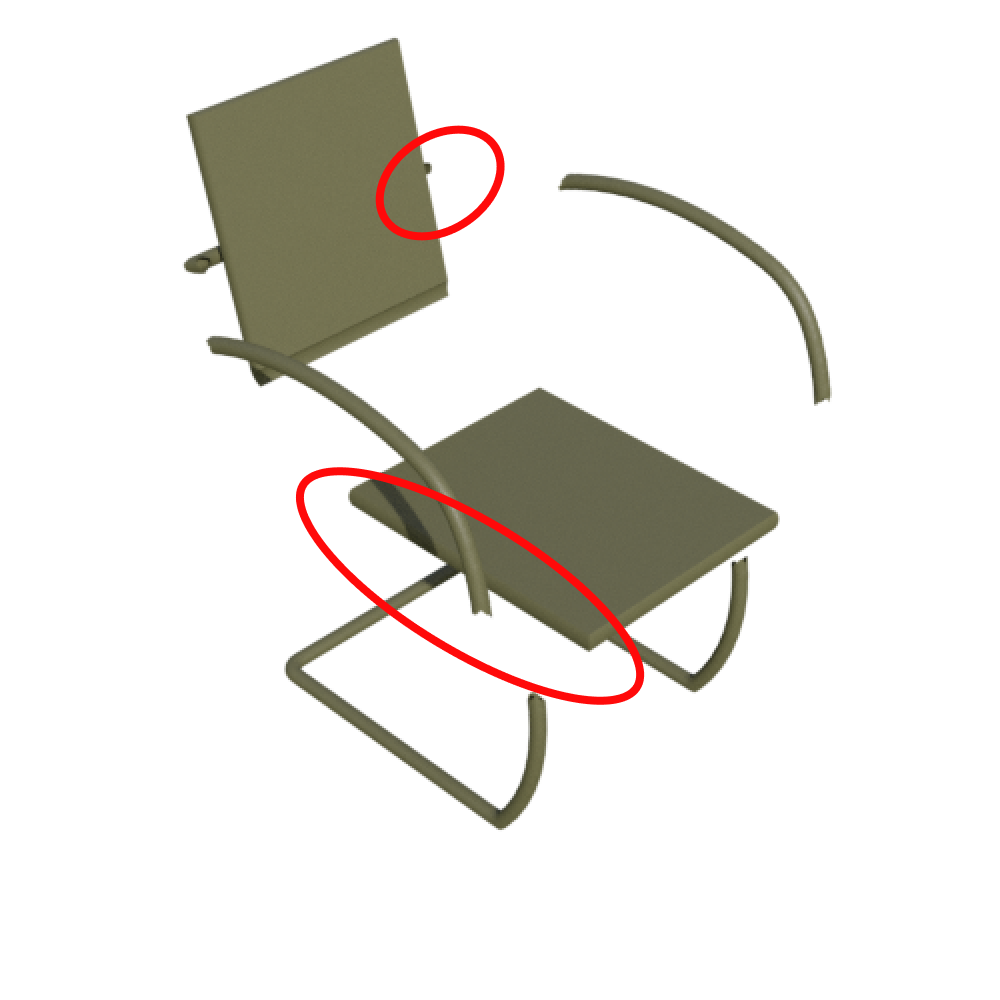} \\
\rotatebox[origin=c]{90}{Watertight parts} & 
\includegraphics[width=.15\linewidth,valign=m]{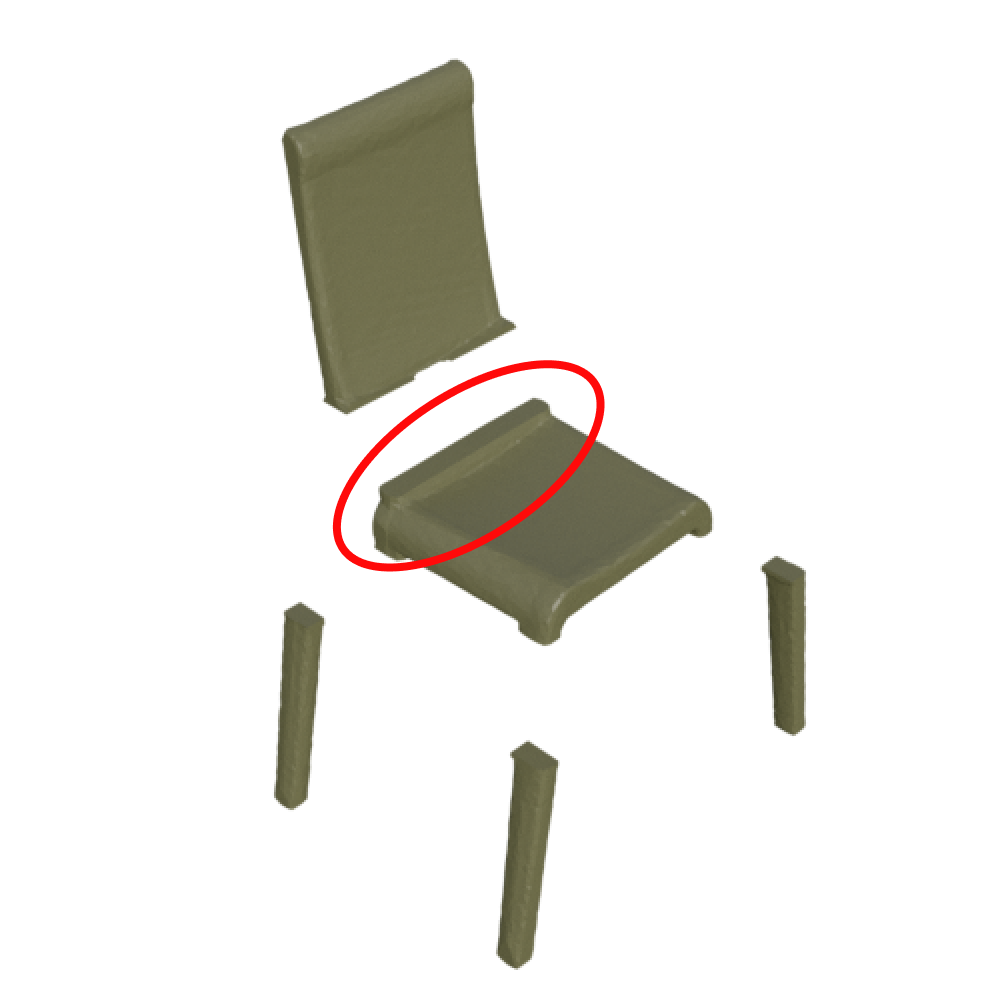} & 
\includegraphics[width=.15\linewidth,valign=m]{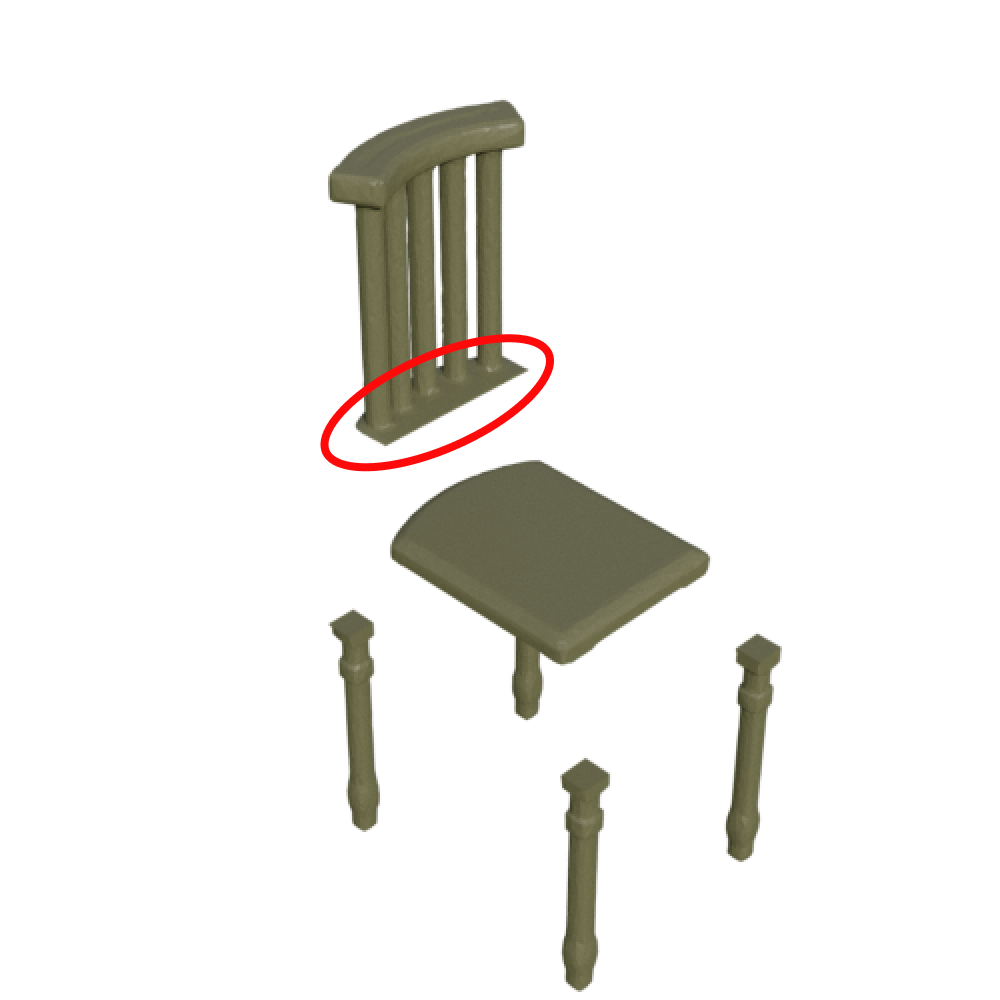} & \includegraphics[width=.15\linewidth,valign=m]{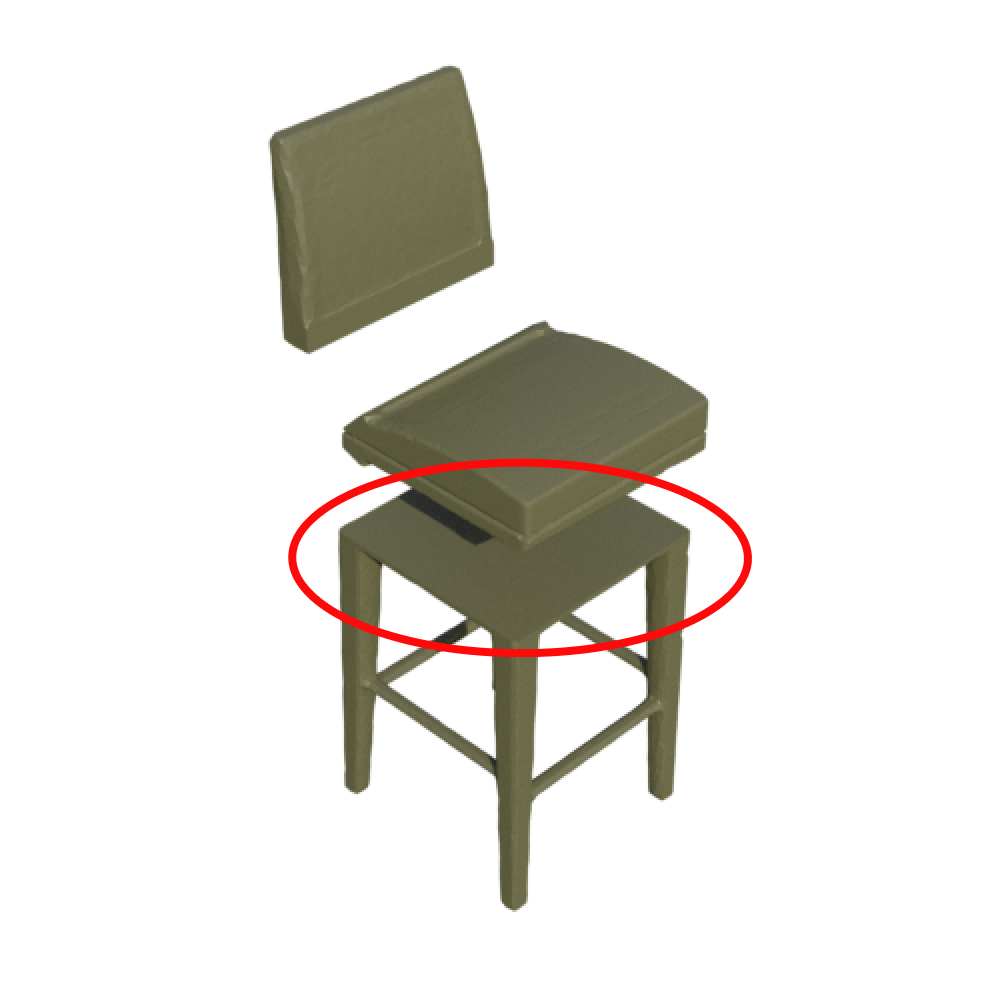} & 
\includegraphics[width=.15\linewidth,valign=m]{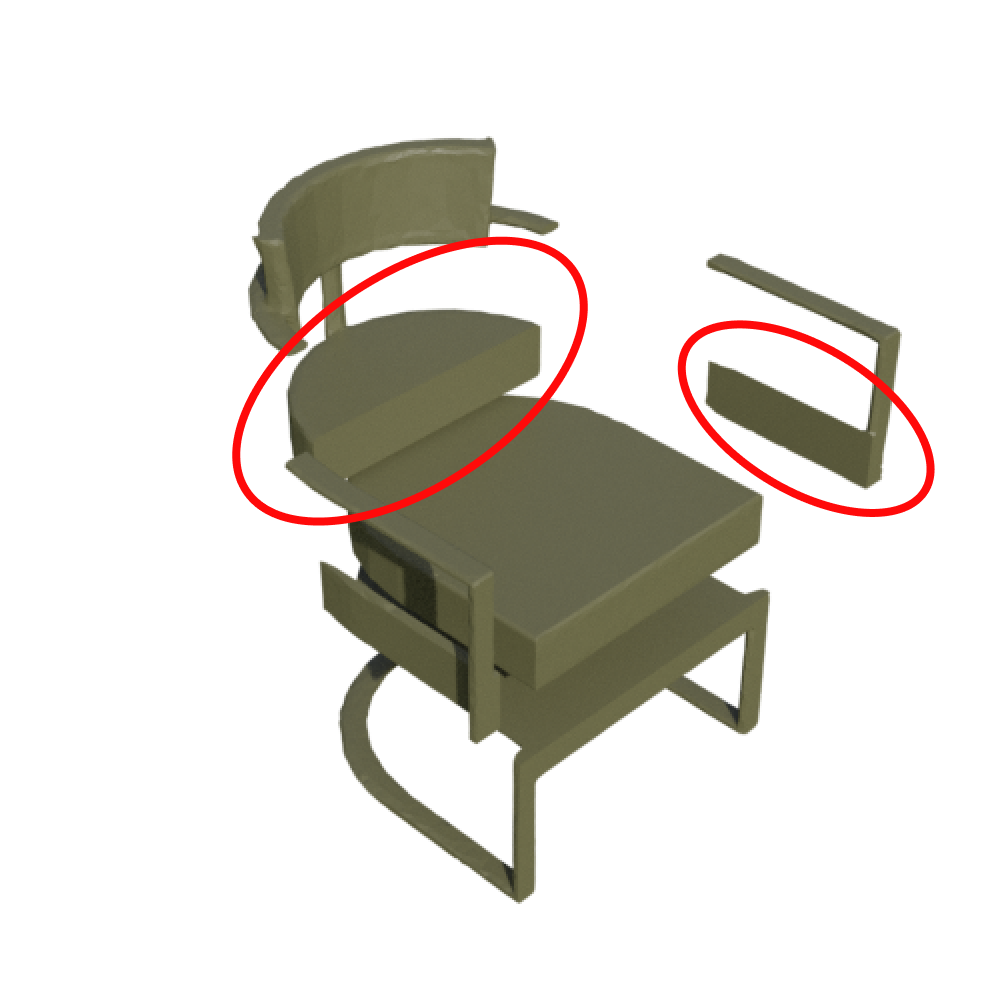} &
\includegraphics[width=.15\linewidth,valign=m]{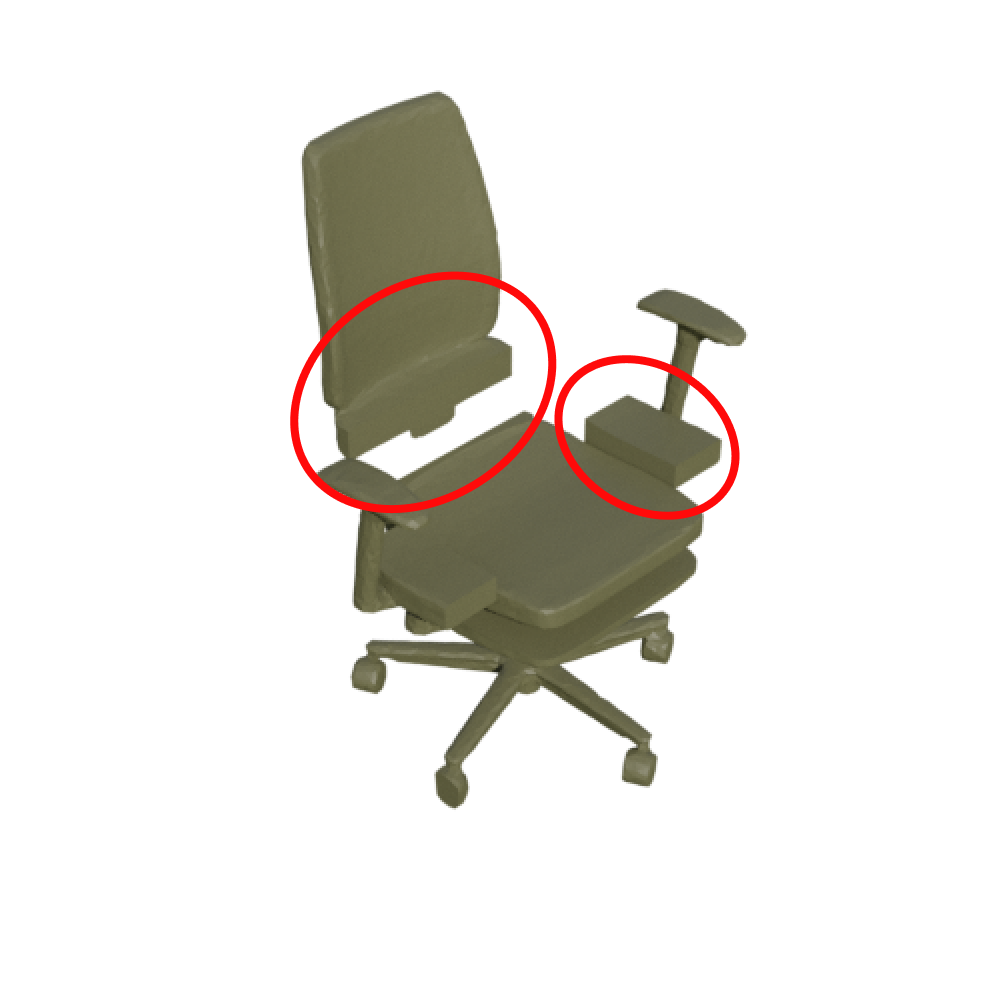} &
\includegraphics[width=.15\linewidth,valign=m]{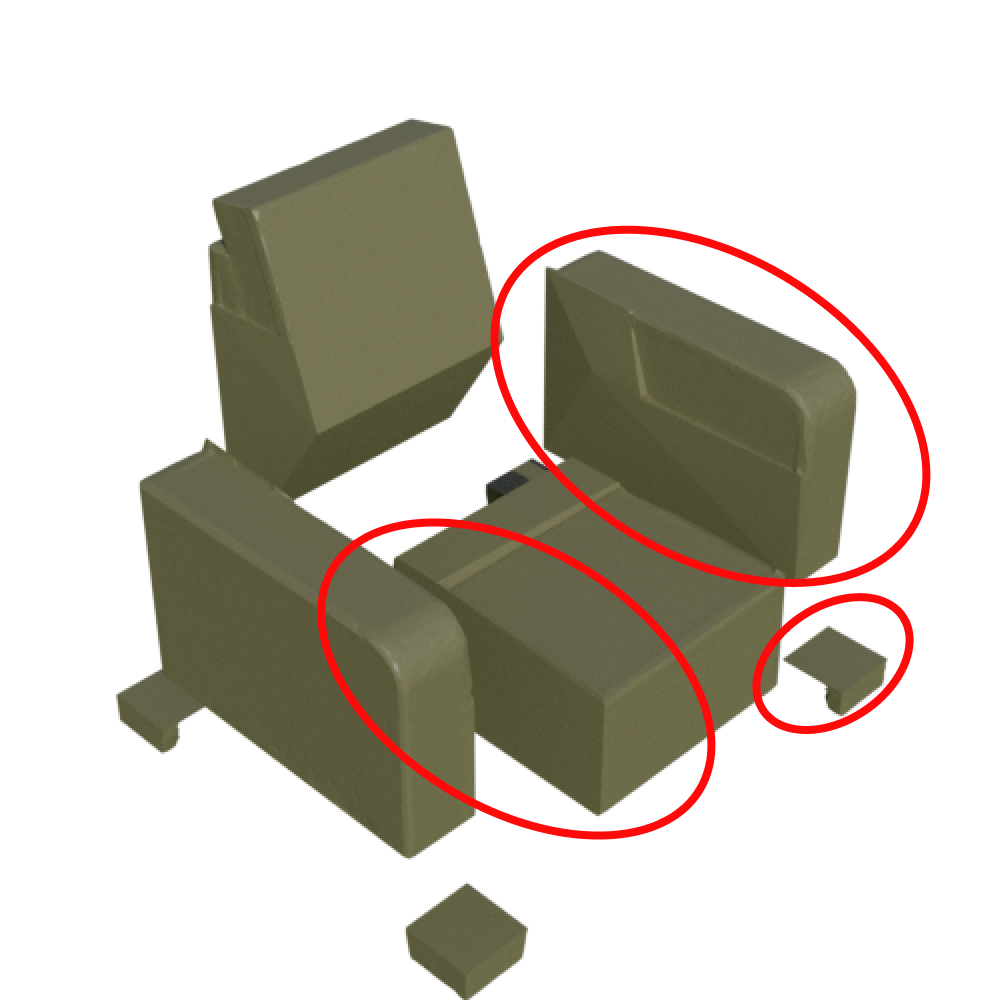} &
\includegraphics[width=.15\linewidth,valign=m]{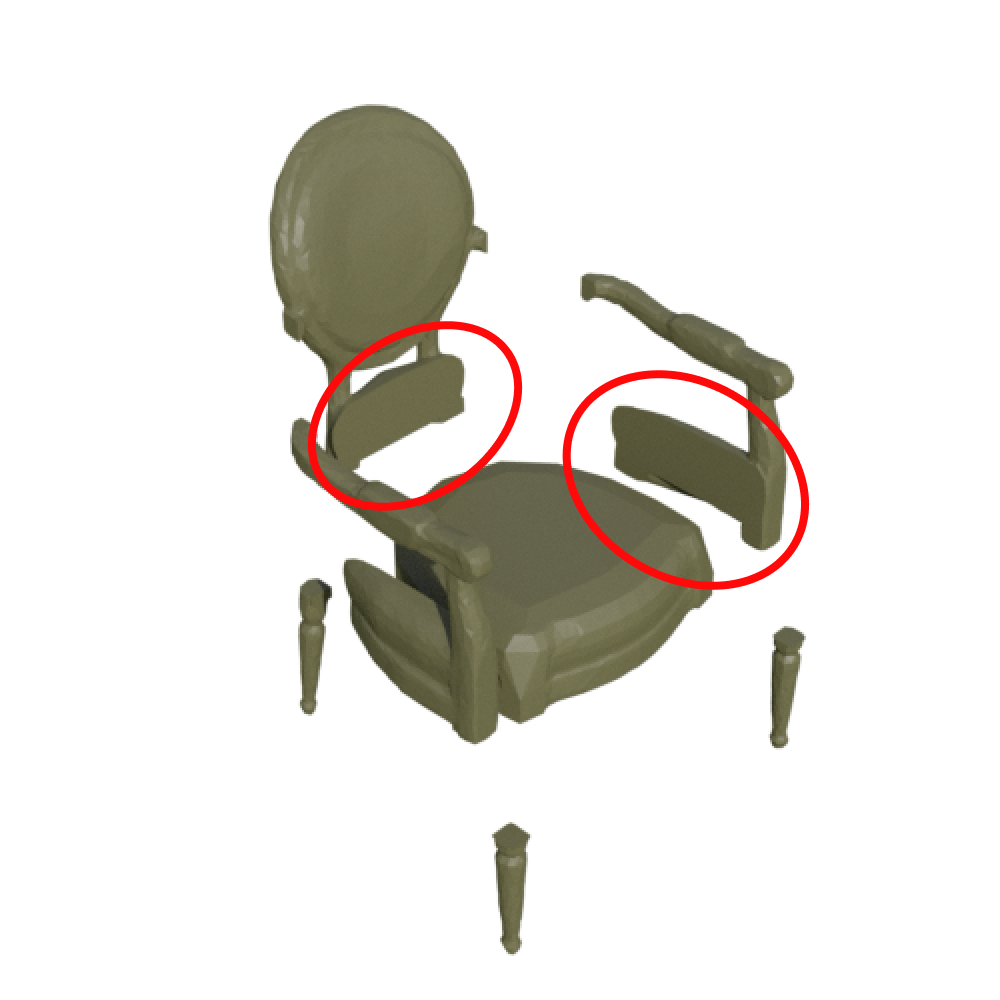} & 
\includegraphics[width=.15\linewidth,valign=m]{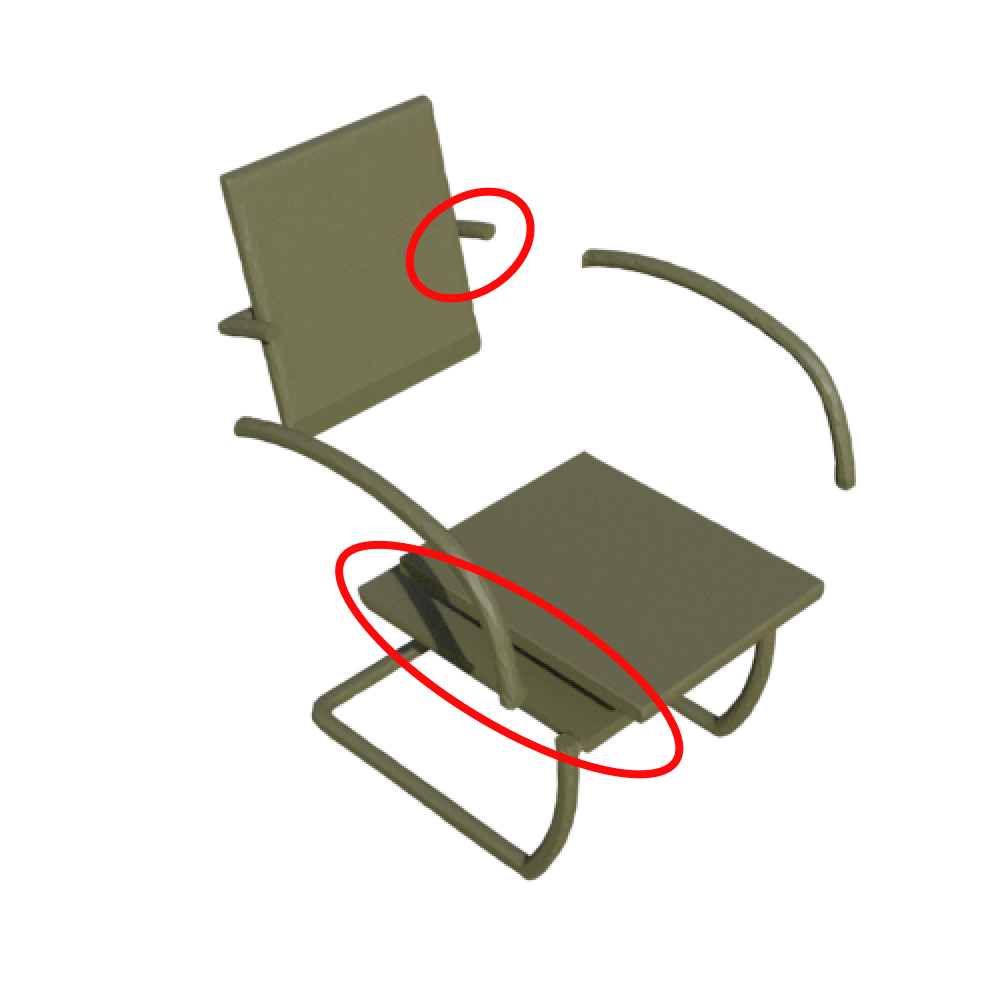}\\
\end{tabular}
\end{center}
\caption{
\textbf{Preprocessing examples.} Characteristic results of our preprocessing pipeline. Overall, our approach is often able to convert non-watertight, segmented  parts  into watertight ones. For example, it effectively closes holes at part connection regions, and turns shell-like parts into solid box-like parts. However, it sometimes introduces preprocessing artifacts like ghost surfaces in part connection areas; or additional volume for parts with complex geometry. Best viewed zoomed in on-screen.
%Still our model 
%does not seem to reproduce these artifacts
%is able to generalize over these artifacts (see Figure \ref{fig:chair_parts}). 
}
\label{fig:preprocessing}
\end{figure*}

%% file: fig/comparison_svr_disn.tex
\begin{figure*}[!ht]
\begin{center}
% \begin{overpic} 
% [width=\linewidth]
% {example-image-a}
% \end{overpic}
\setlength\tabcolsep{-4pt}
\begin{tabular}{c@{\hskip 5pt}llllllll}
\rotatebox[origin=c]{90}{Input}\vspace{-5pt} & \includegraphics[width=.14\linewidth,valign=m]{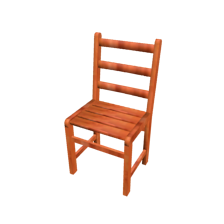} & 
\includegraphics[width=.14\linewidth,valign=m]{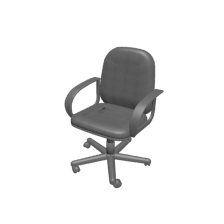} &
\includegraphics[width=.14\linewidth,valign=m]{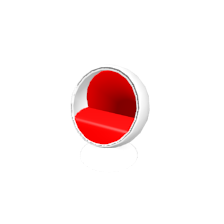} &
\includegraphics[width=.14\linewidth,valign=m]{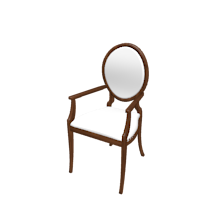} &
\includegraphics[width=.14\linewidth,valign=m]{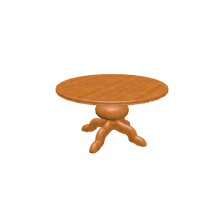} &
\includegraphics[width=.14\linewidth,valign=m]{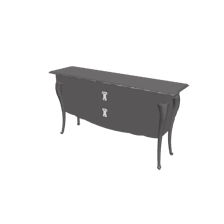} &
\includegraphics[width=.14\linewidth,valign=m]{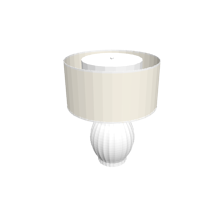} & 
\includegraphics[width=.14\linewidth,valign=m]{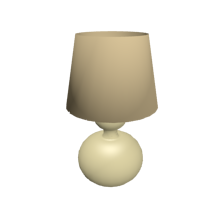} 
\hspace{-5pt}

 \\
\rotatebox[origin=c]{90}{DISN} \vspace{-5pt}& \includegraphics[width=.14\linewidth,valign=m]{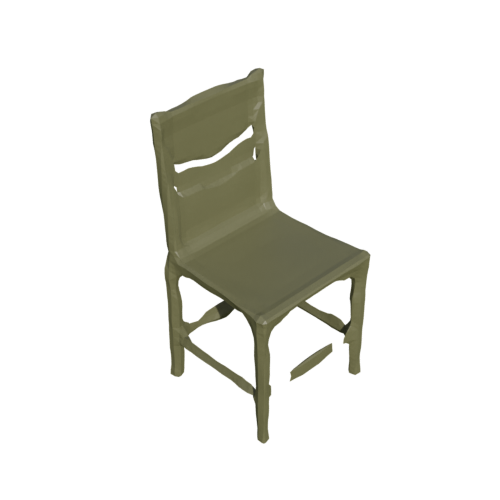} & 
\includegraphics[width=.14\linewidth,valign=m]{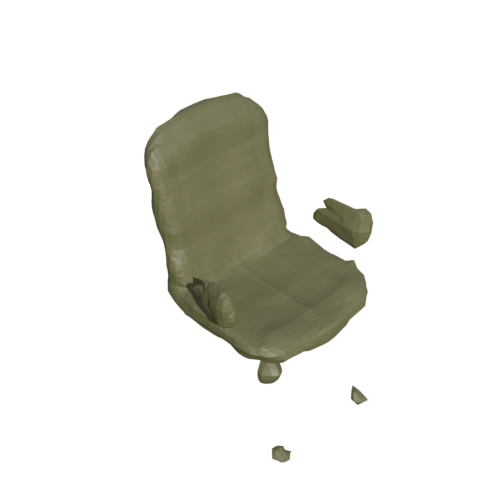} & 
\includegraphics[width=.14\linewidth,valign=m]{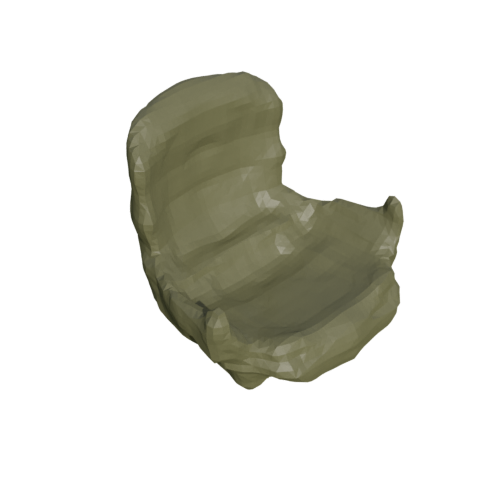} & 
\includegraphics[width=.14\linewidth,valign=m]{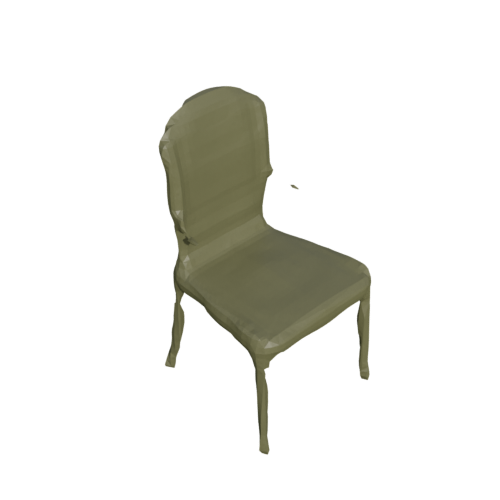} & 
\includegraphics[width=.14\linewidth,valign=m]{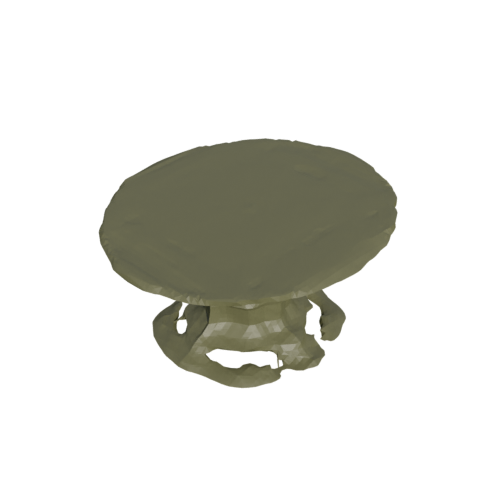} & 
\includegraphics[width=.14\linewidth,valign=m]{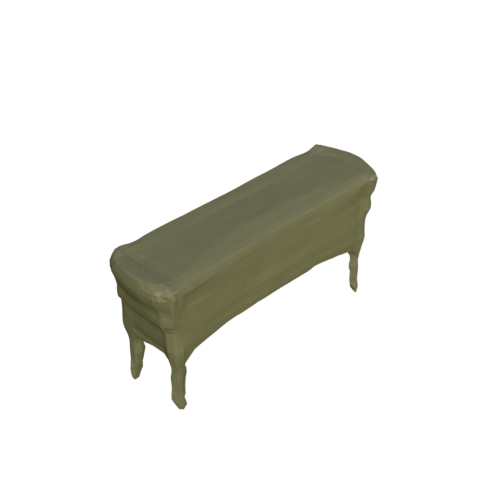} & 
\hspace{-5pt}
\includegraphics[width=.14\linewidth,valign=m]{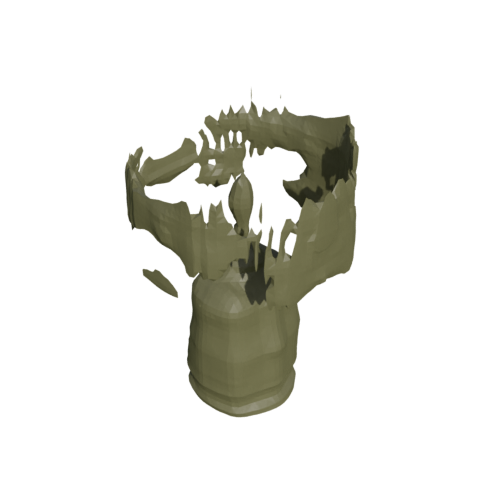}  &
\includegraphics[width=.14\linewidth,valign=m]{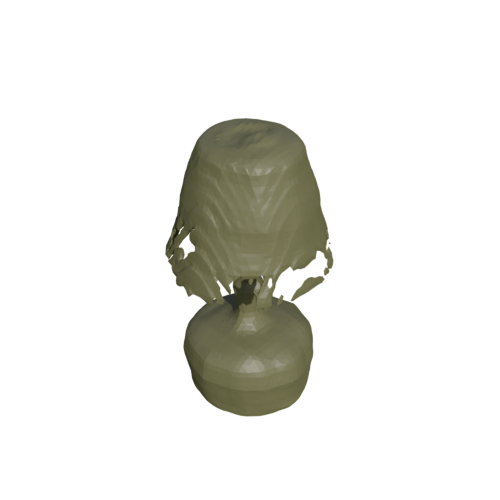}  
 \\
\rotatebox[origin=c]{90}{D2IM}\vspace{-5pt} & \includegraphics[width=.14\linewidth,valign=m]{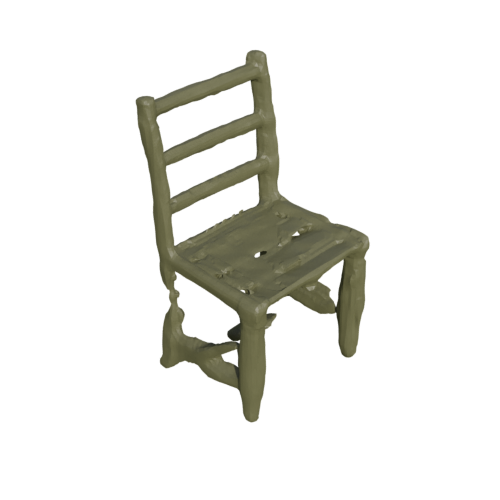}& 
\includegraphics[width=.14\linewidth,valign=m]{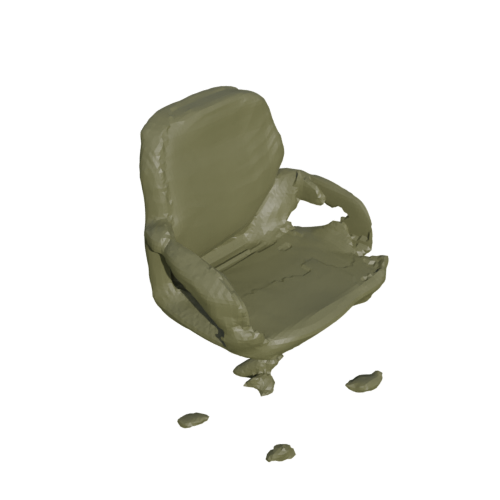}& 
\includegraphics[width=.14\linewidth,valign=m]{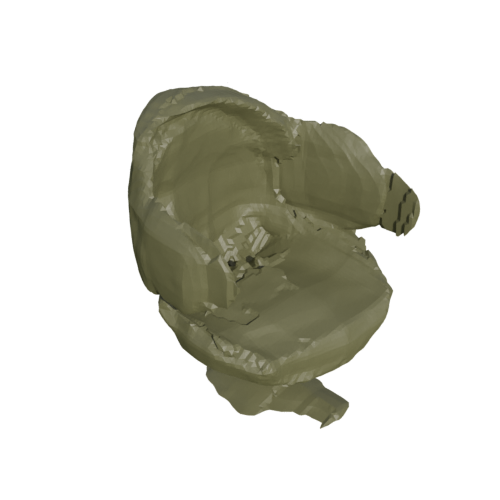}& 
\includegraphics[width=.14\linewidth,valign=m]{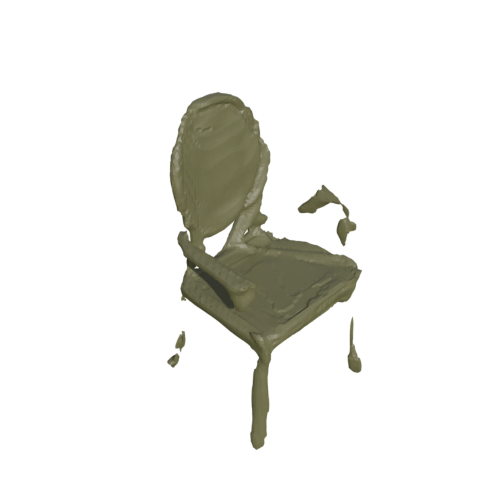}& 
\includegraphics[width=.14\linewidth,valign=m]{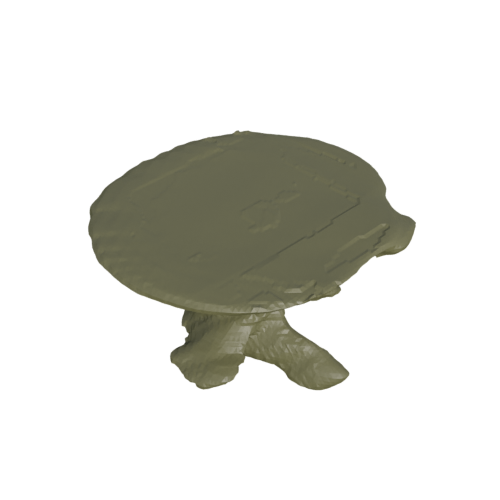}& 
\includegraphics[width=.14\linewidth,valign=m]{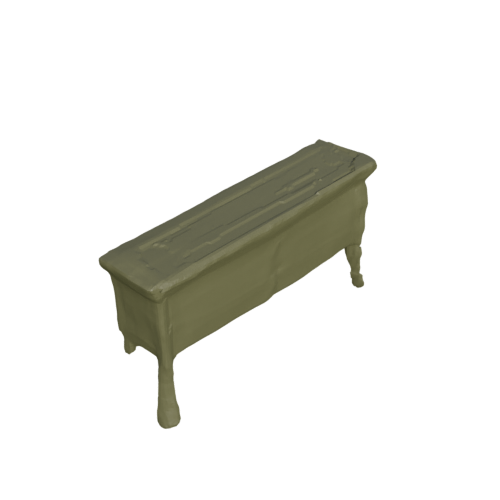}& 
\includegraphics[width=.14\linewidth,valign=m]{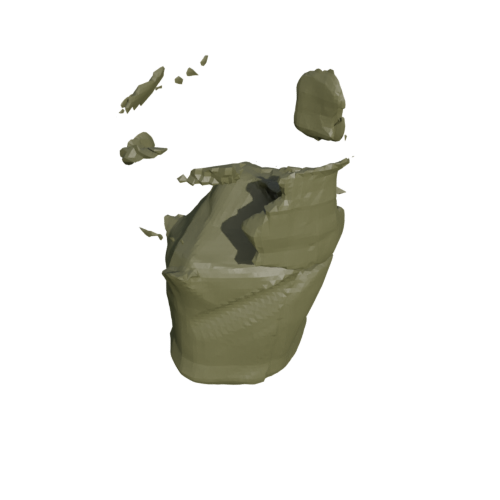} &
\includegraphics[width=.14\linewidth,valign=m]{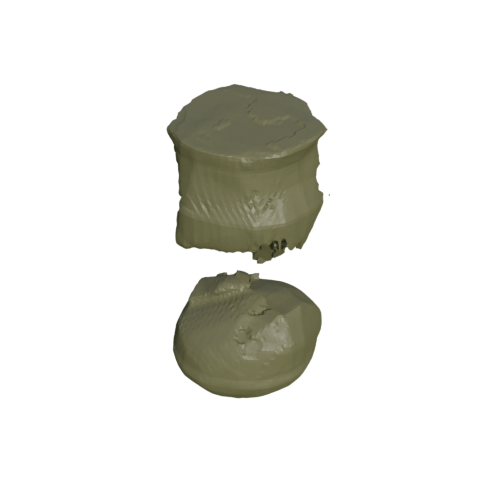} 
\hspace{-5pt}

 \\
\rotatebox[origin=c]{90}{Ours}\vspace{-5pt} & \includegraphics[width=.14\linewidth,valign=m]{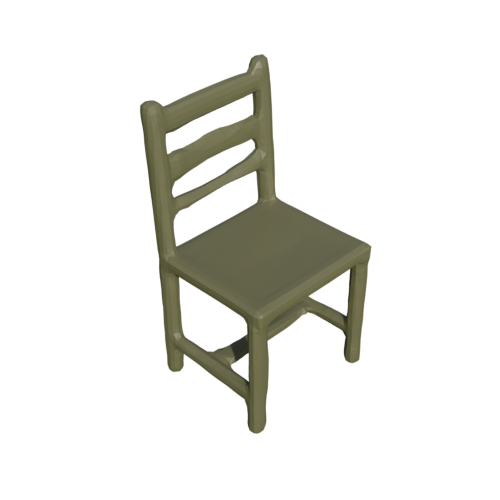} & 
\includegraphics[width=.14\linewidth,valign=m]{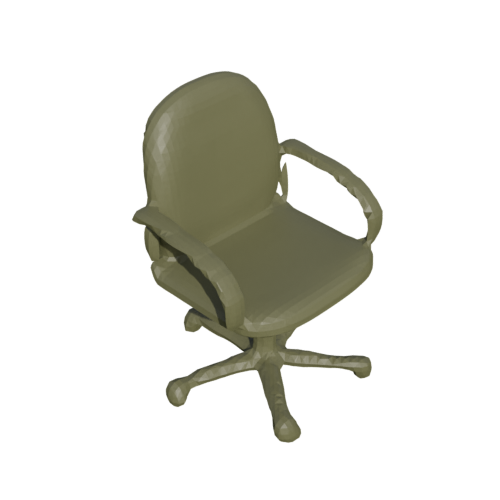} & 
\includegraphics[width=.14\linewidth,valign=m]{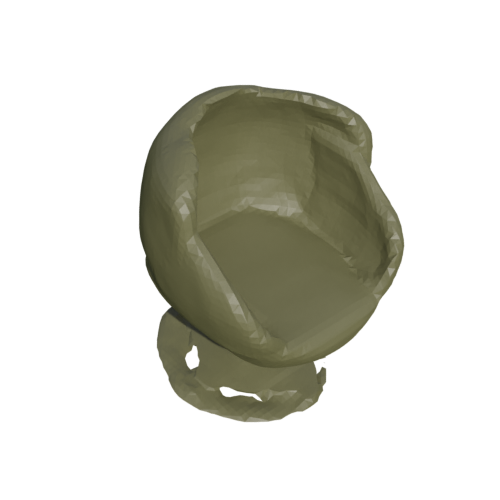} & 
\includegraphics[width=.14\linewidth,valign=m]{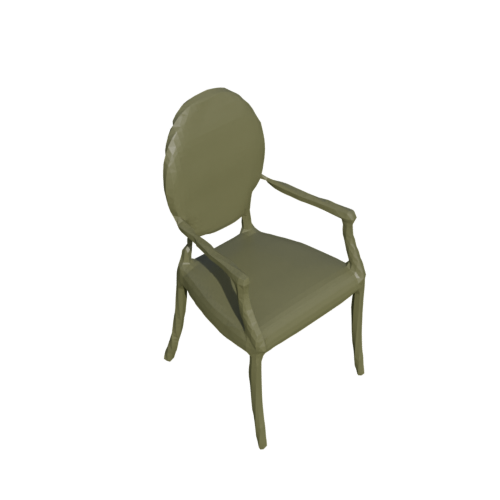} & 
\includegraphics[width=.14\linewidth,valign=m]{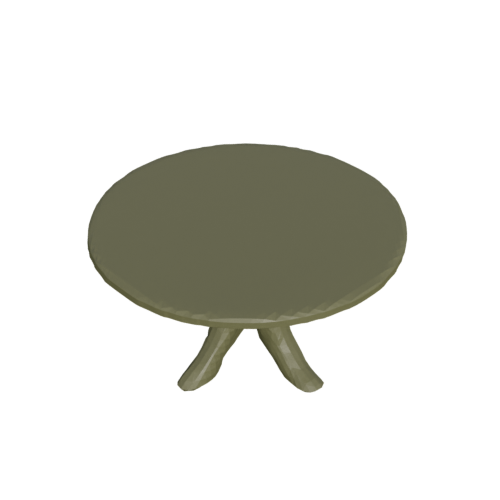} & 
\includegraphics[width=.14\linewidth,valign=m]{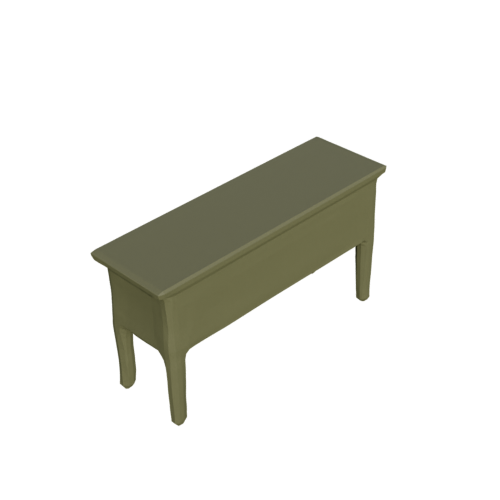} & 
\includegraphics[width=.14\linewidth,valign=m]{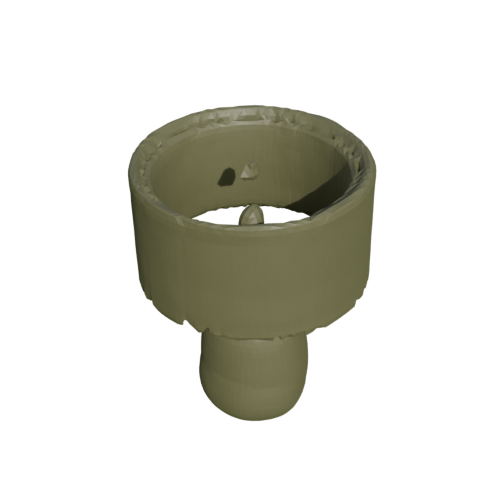} &
\includegraphics[width=.14\linewidth,valign=m]{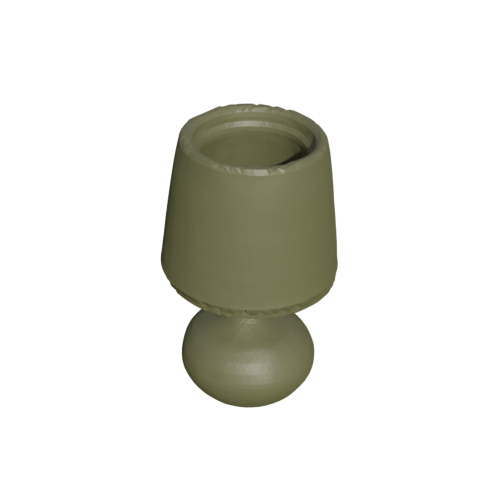} 

\hspace{-1pt}

 \\
\rotatebox[origin=c]{90}{GT}\vspace{-5pt} & \includegraphics[width=.14\linewidth,valign=m]{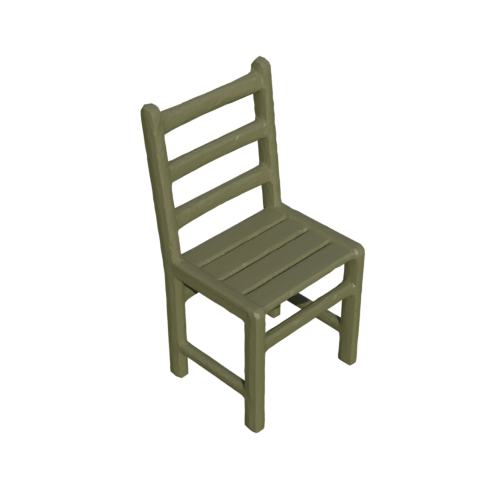} &
\includegraphics[width=.14\linewidth,valign=m]{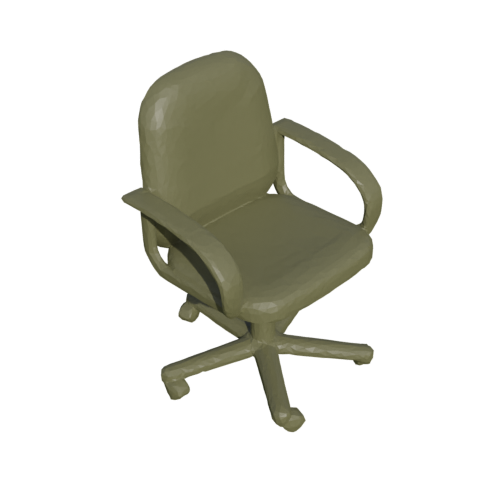} & 
\includegraphics[width=.14\linewidth,valign=m]{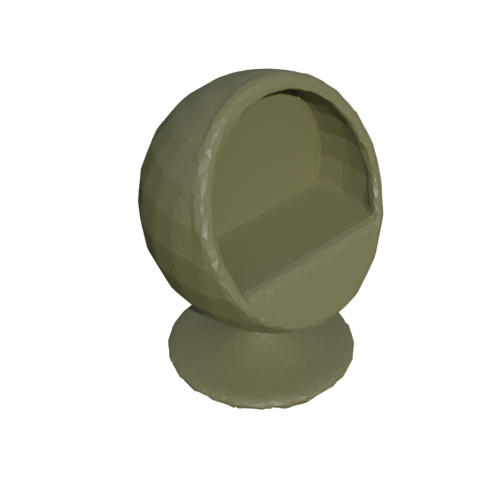} & 
\includegraphics[width=.14\linewidth,valign=m]{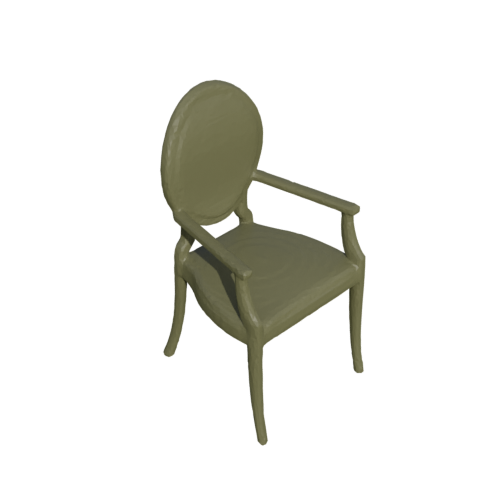} & 
\includegraphics[width=.14\linewidth,valign=m]{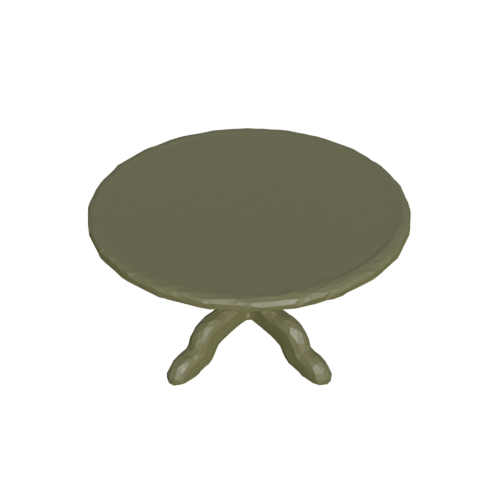} & 
\includegraphics[width=.14\linewidth,valign=m]{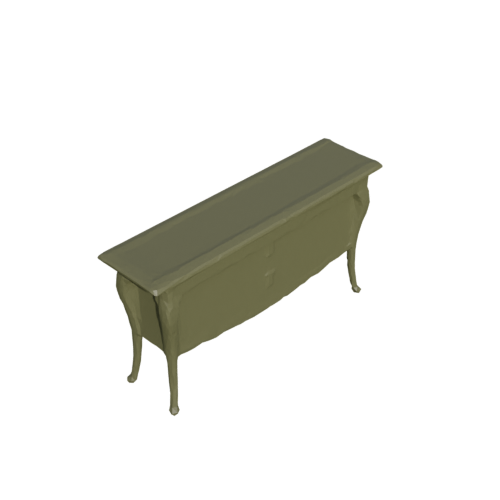} & 
\includegraphics[width=.14\linewidth,valign=m]{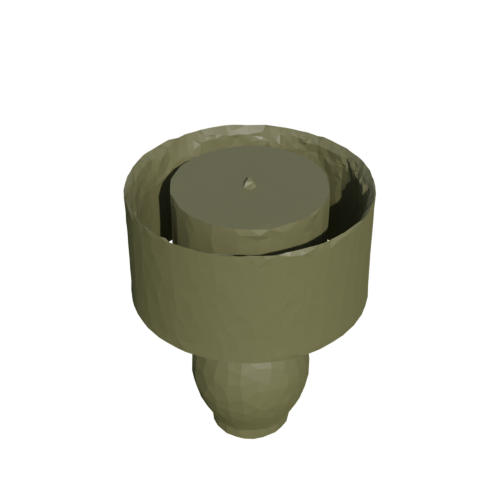} &
\includegraphics[width=.14\linewidth,valign=m]{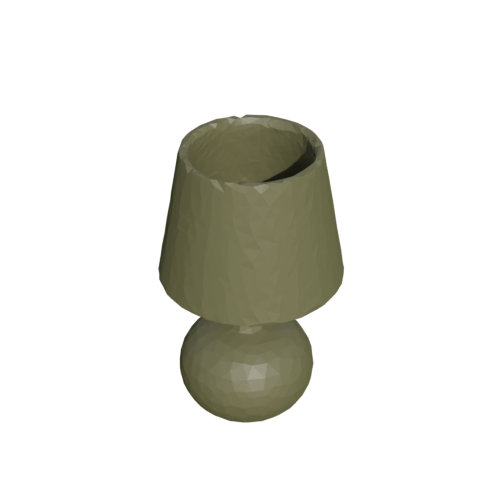} 
\vspace{4pt}
\\

\arrayrulecolor{black}
\end{tabular}

\end{center}
%\vspace{-16pt}
\caption{\textbf{ANISE vs part-unaware single view reconstruction methods.} We compare with with D2IM and DISN  on ShapeNet. Our model is able to reconstruct fine details of the input like chair armrests or office chair swivel legs.}
\label{fig:comparison_svr_combined}
\end{figure*}

%% file: fig/comparison_svr_parts.tex
\begin{figure*}[!ht]
\begin{center}
% \begin{overpic} 
% [width=\linewidth]
% {example-image-a}
% \end{overpic}
\setlength\tabcolsep{-4pt}
\begin{tabular}{c@{\hskip 5pt}llllllll}

\rotatebox[origin=c]{90}{Input}\vspace{-5pt} & \includegraphics[width=.14\linewidth,valign=m]{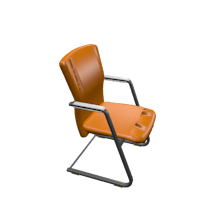} & 
\includegraphics[width=.14\linewidth,valign=m]{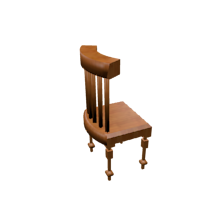} &
\includegraphics[width=.14\linewidth,valign=m]{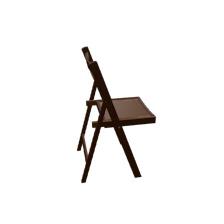} &
\includegraphics[width=.14\linewidth,valign=m]{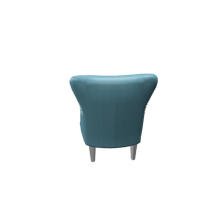} &
\includegraphics[width=.14\linewidth,valign=m]{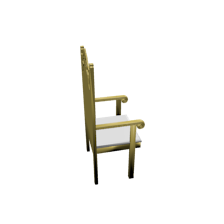} &
\hspace{-5pt}
\includegraphics[width=.14\linewidth,valign=m]{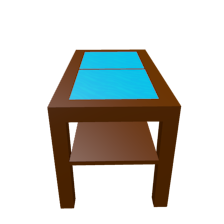} &
\includegraphics[width=.14\linewidth,valign=m]{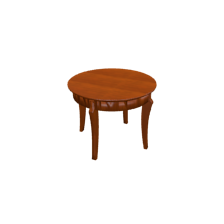} &
\includegraphics[width=.14\linewidth,valign=m]{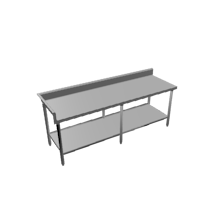} 

 \\
\rotatebox[origin=c]{90}{JLRD} \vspace{-5pt}& \includegraphics[width=.14\linewidth,valign=m]{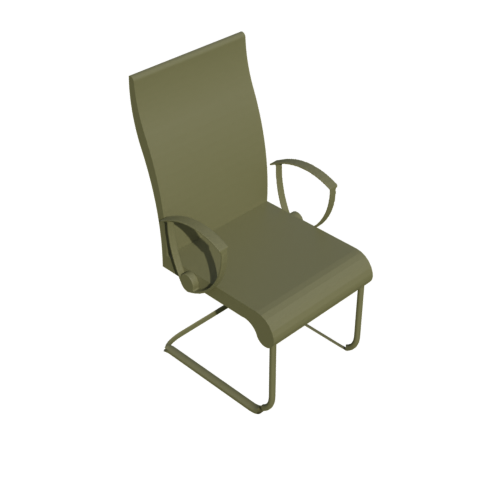} & 
\includegraphics[width=.14\linewidth,valign=m]{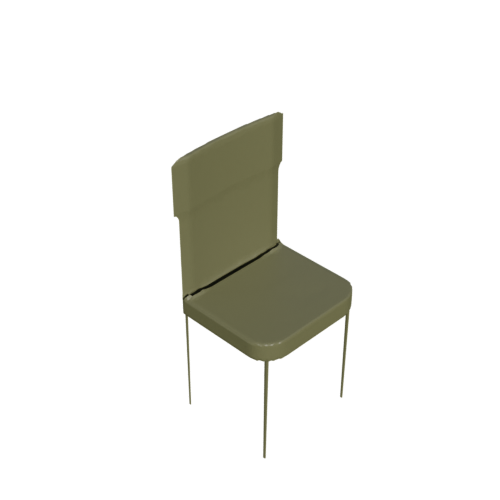} & 
\includegraphics[width=.14\linewidth,valign=m]{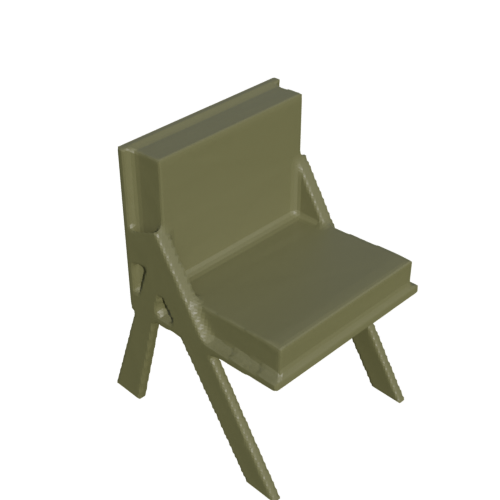} & 
\includegraphics[width=.14\linewidth,valign=m]{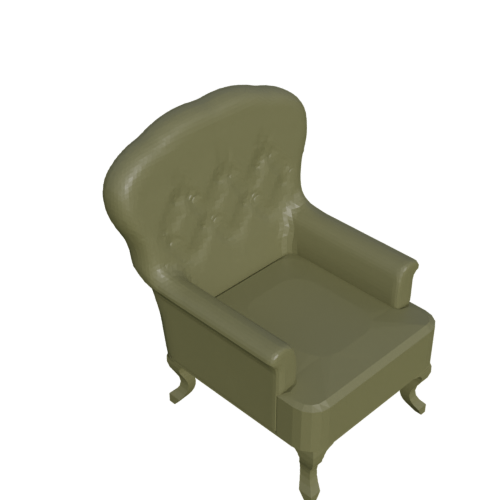} & 
\includegraphics[width=.14\linewidth,valign=m]{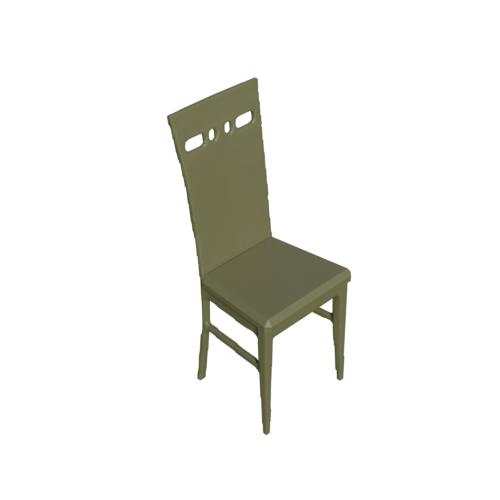} & 
\hspace{-5pt}
\includegraphics[width=.14\linewidth,valign=m]{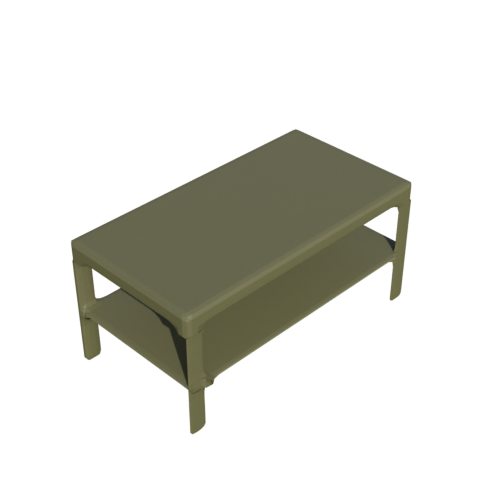}  &
\includegraphics[width=.14\linewidth,valign=m]{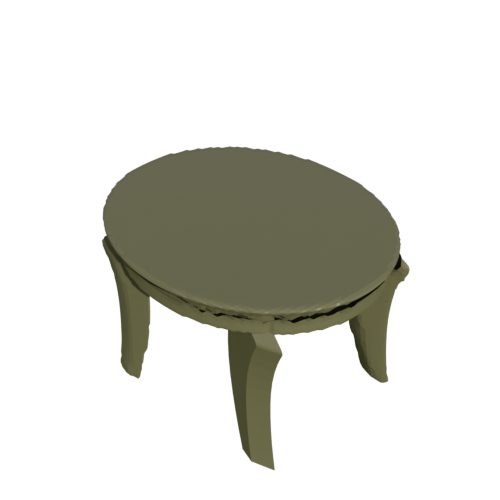}  &
\includegraphics[width=.14\linewidth,valign=m]{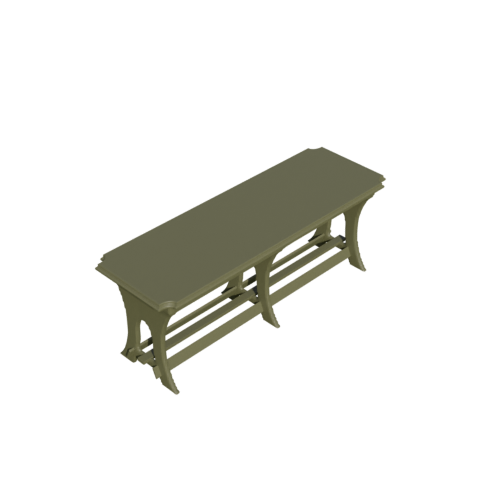}  
 \\
\rotatebox[origin=c]{90}{PQ-Net}\vspace{-5pt} & \includegraphics[width=.14\linewidth,valign=m]{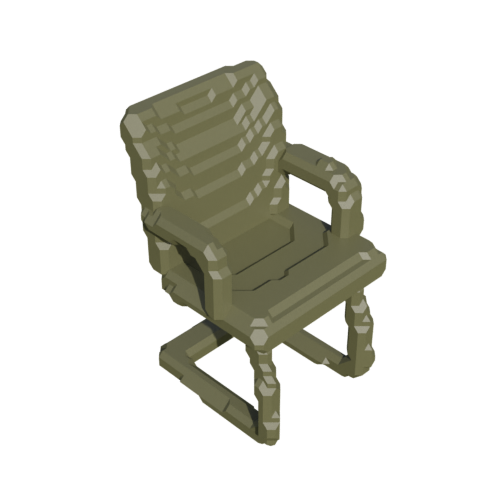}& 
\includegraphics[width=.14\linewidth,valign=m]{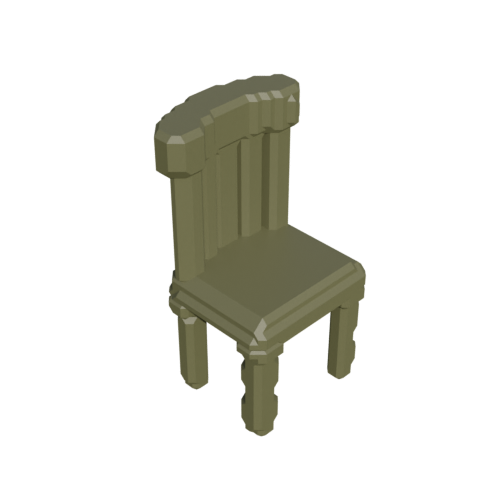}& 
\includegraphics[width=.14\linewidth,valign=m]{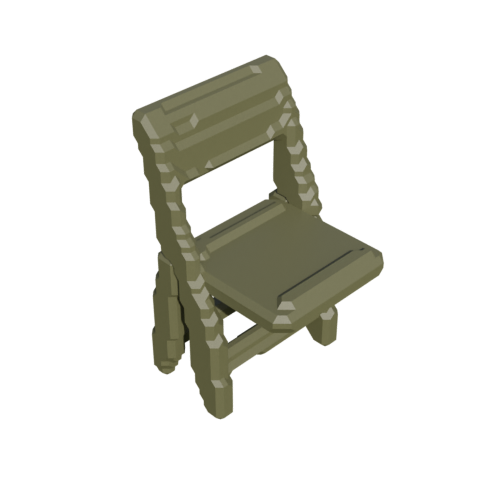}& 
\includegraphics[width=.14\linewidth,valign=m]{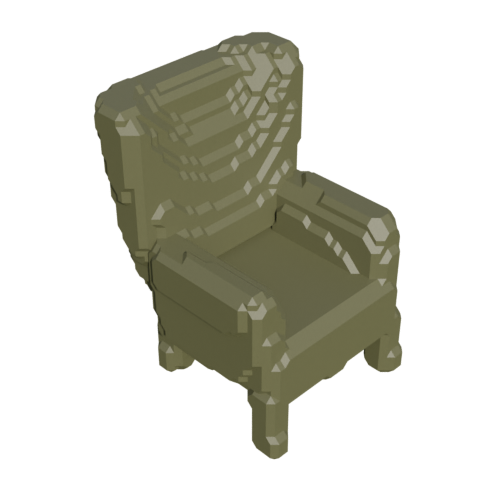}& 
\includegraphics[width=.14\linewidth,valign=m]{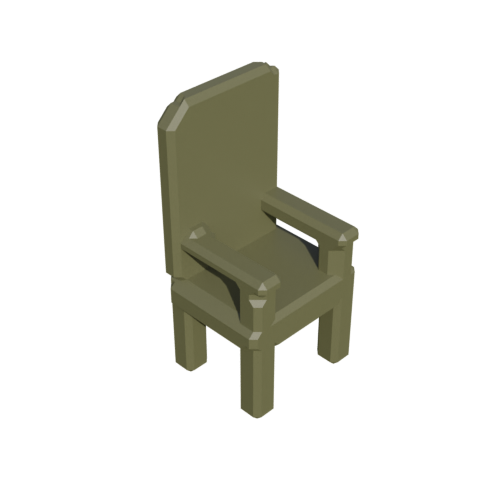}& 
\hspace{-5pt}
\includegraphics[width=.14\linewidth,valign=m]{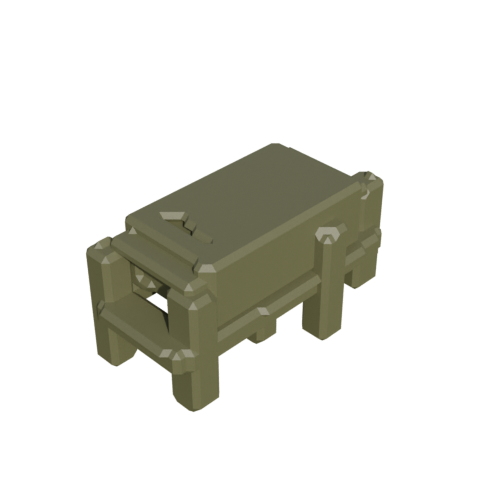} &
\includegraphics[width=.14\linewidth,valign=m]{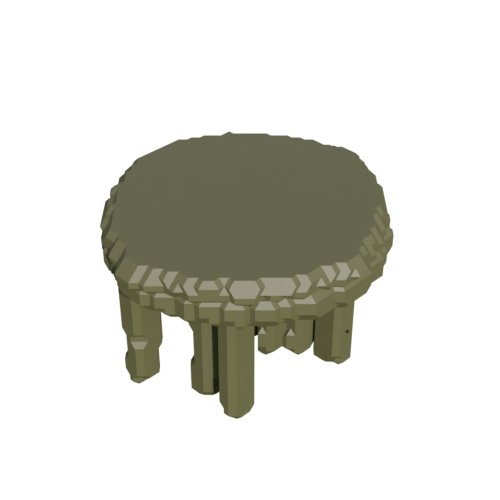} &
\includegraphics[width=.14\linewidth,valign=m]{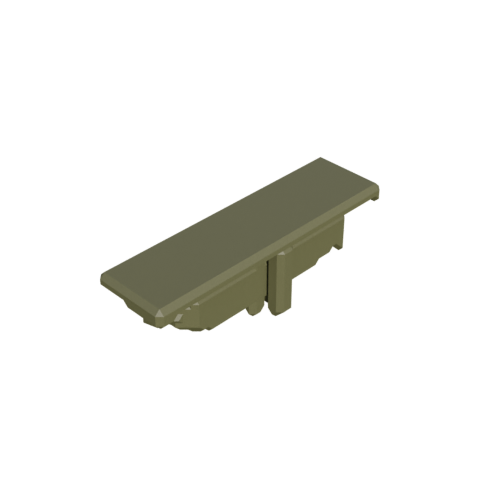} 
 \\
\rotatebox[origin=c]{90}{Ours}\vspace{-5pt} & \includegraphics[width=.14\linewidth,valign=m]{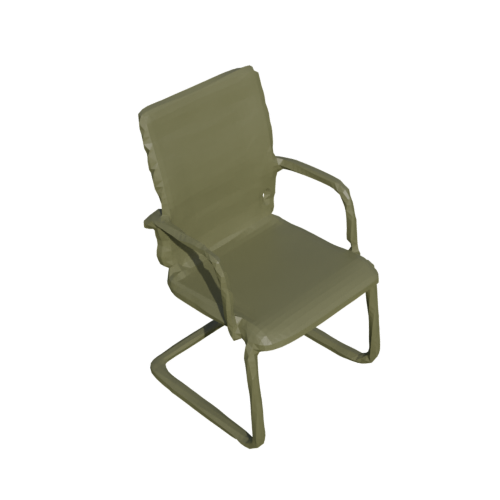} & 
\includegraphics[width=.14\linewidth,valign=m]{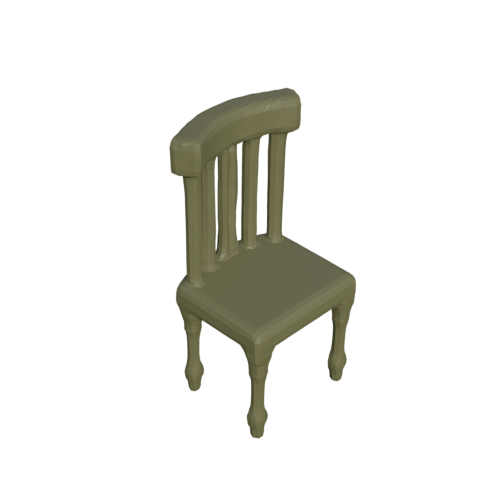} & 
\includegraphics[width=.14\linewidth,valign=m]{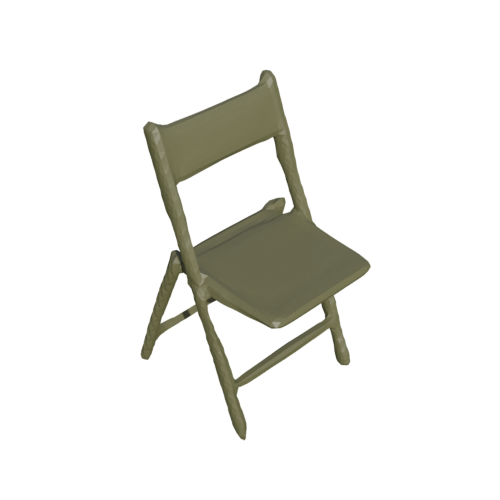} & 
\includegraphics[width=.14\linewidth,valign=m]{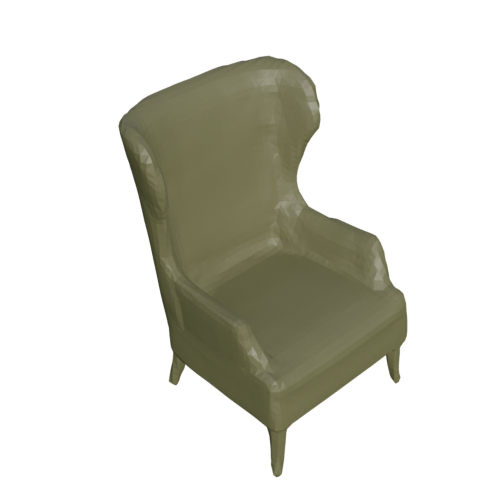} & 
\includegraphics[width=.14\linewidth,valign=m]{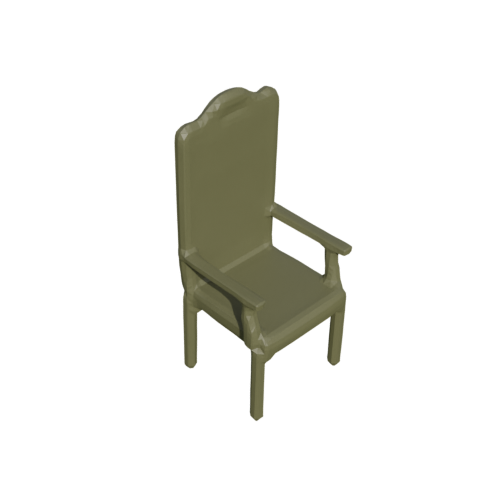} & 
\hspace{-5pt}
\includegraphics[width=.14\linewidth,valign=m]{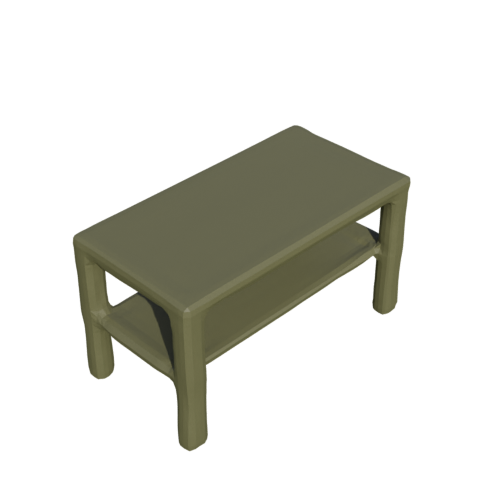} &
\includegraphics[width=.14\linewidth,valign=m]{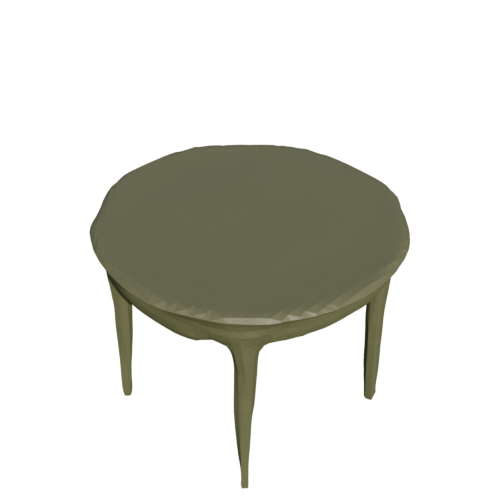} &
\includegraphics[width=.14\linewidth,valign=m]{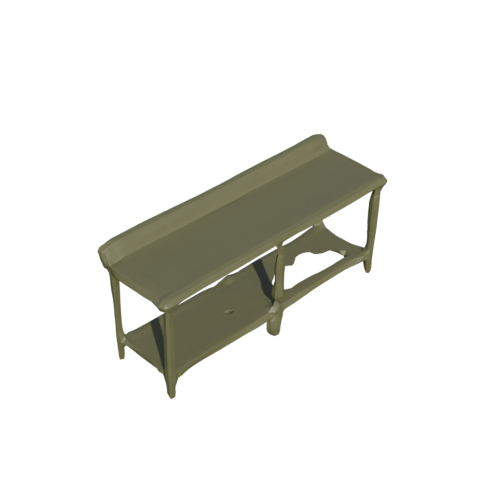} 
 \\
\rotatebox[origin=c]{90}{GT}\vspace{-5pt} & \includegraphics[width=.14\linewidth,valign=m]{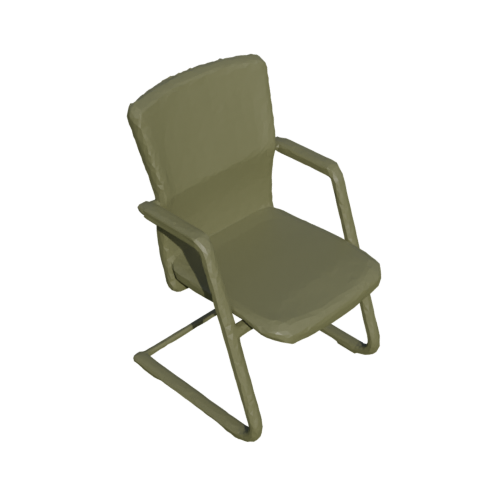} &
\includegraphics[width=.14\linewidth,valign=m]{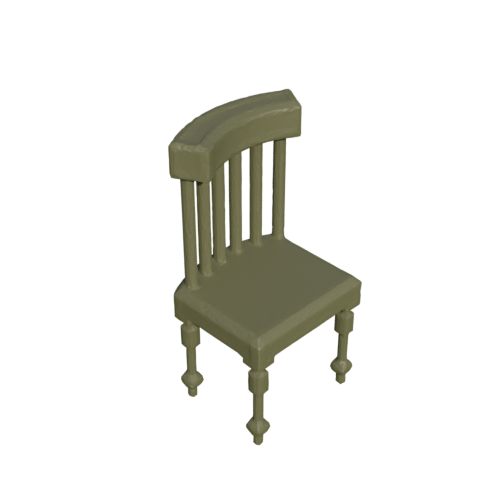} & 
\includegraphics[width=.14\linewidth,valign=m]{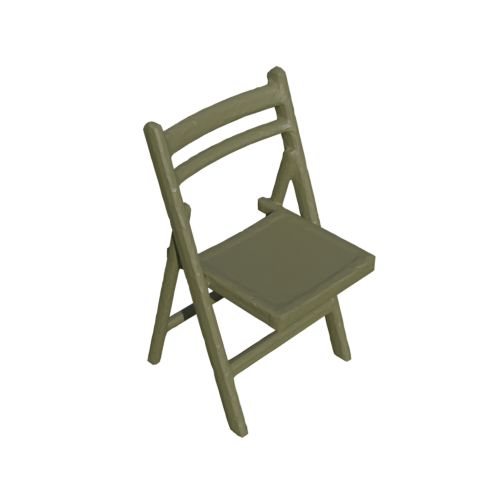} & 
\includegraphics[width=.14\linewidth,valign=m]{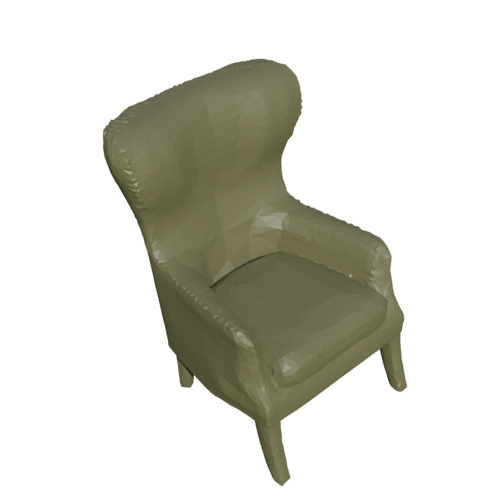} & 
\includegraphics[width=.14\linewidth,valign=m]{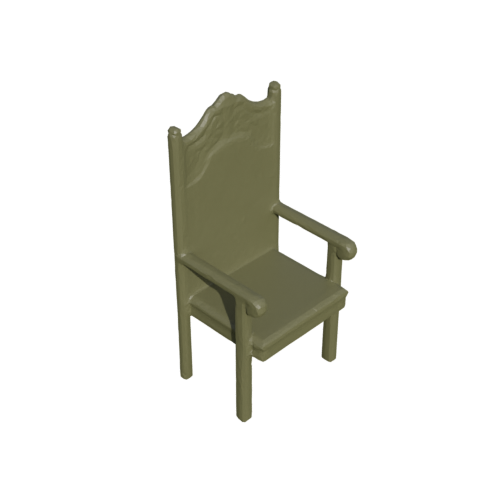} & 
\includegraphics[width=.14\linewidth,valign=m]{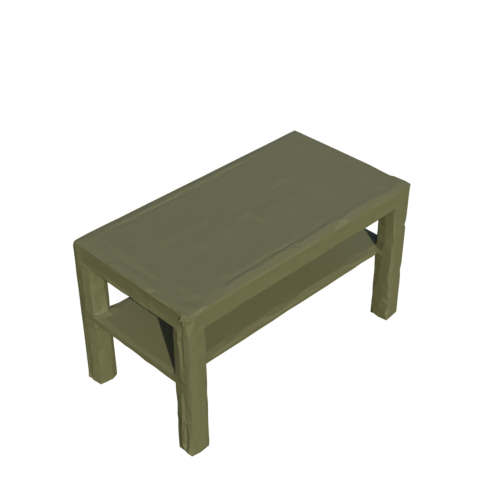} &
\includegraphics[width=.14\linewidth,valign=m]{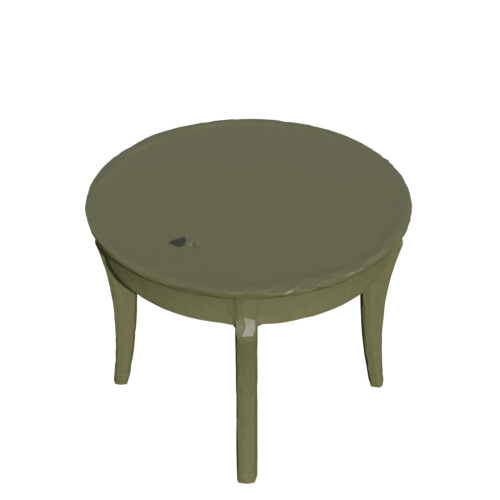} &
\includegraphics[width=.14\linewidth,valign=m]{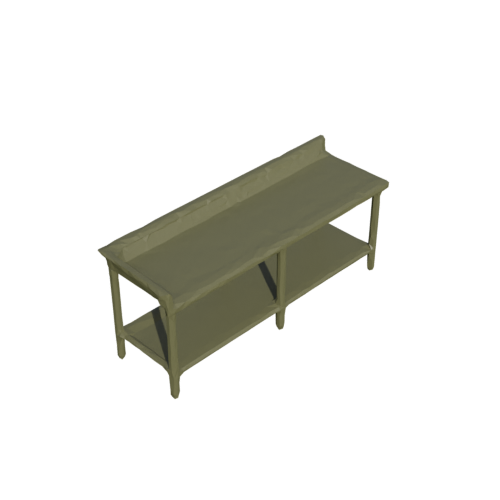} 
\hspace{-5pt}

 \\
\arrayrulecolor{black}
\end{tabular}

\end{center}
%\vspace{-16pt}
\caption{\textbf{ANISE  vs part-aware single view reconstruction methods.} We compare with JLRD and PQ-Net on PartNet. Compared to JLRD our model better models shape topology. Compared to PQ-Net our model better reconstructs small geometric details.}
\label{fig:comparison_svr_parts}
\end{figure*}

%% file: sec/4_experiments.tex
\section{Experiments and Results}

We experimented with our method in two different operational setups -- part retrieval \& assembly (PR\&A) and implicit function assembly.
The former allows constraining the parts of the estimated shape to be \emph{exactly} like one in a shape database.
The later leads to shapes whose parts are better connected to each other while also benefiting from additional training by fine-tuning the model on complete shapes.
Both modes of operation are quantitatively evaluated in benchmarks where shapes are reconstructed from single RGB images or point clouds. Below we discuss datasets, evaluation metrics, comparisons of our method with state-of-the-art models and its applications.

\paragraph{Datasets} 
We use the \emph{PartNet} dataset~\cite{mo2019partnet} for our experiments, a large-scale 3D mesh dataset with semantic segmentation labels on three levels of hierarchy. For our experiments we use the three large shape categories, Tables, Chairs and Lamps (8218, 6326 and 2203 meshes respectively). 
For point cloud reconstruction, we  train and evaluate our method and competing ones according to the PartNet provided splits, except for the  comparison case of our method with JLRD~\cite{jldr}: in this case, we use the splits provided by authors. For comparison with part-unaware single-view reconstruction methods we also use \emph{ShapeNetCore}~\cite{shapenet} with same categories as PartNet. 
The renderings used as input on the single-view reconstruction experiments are obtained from JLRD~\cite{jldr} for the part-aware comparison and from DISN~\cite{disn} for the part-agnostic comparison.

\input{fig/comparison_pc}

\paragraph{Part extraction and preprocessing} We use part Signed Distance Fields (SDF) as supervision in our experiments. Computing those requires watertight meshes and we compute those using code provided by Mescheder \etal~\cite{occnet} which is a modification of the code by Stutz \etal.~\cite{stutz2018learning}. 

 The PartNet dataset provides part meshes on three hierarchical levels of granularity: coarse-, middle- and fine-grained. For our experiments, we use the coarse level. The parts themselves are not watertight (like the full shapes). In addition, there are two additional complications: (a) there are holes in locations of part connections (e.g. holes in the chair seat where legs connect to it) and (b) there are shell-like parts which do not form a closed surface (e.g., flat sheets). To produce a watertight version of each part, we compute the convex hull of the part and intersect it with watertight version of the full shape. Since watertight meshes are slightly different from original ones (e.g. thin parts become slightly thicker), we align the mesh constructed from parts with its watertight version using rigid registration with Coherent Point Drift \cite{myronenko2010point}. Our preprocessing procedure fails in approximately $10\%$ of cases resulting in missing parts. We still use shapes with missing parts for training. After preprocessing, average numbers of parts per shape are 5.6/2.8/2.9 for Chair, Table and Lamp categories respectively. Following \cite{occnet} we scale and center all shapes and their parts to a unit cube with 0.1 padding. We save transformation parameters (isotropic scale and translation vector) as a supervision for part positions. For ShapeNetCore experiments we match its shapes with our preprocessed PartNet shapes matching provided by PartNet authors. We provide more details on preprocessing in supplementary.

 Characteristic results from our pre-processing pipeline are shown in Figure \ref{fig:preprocessing}. We can make the following observations. First, our approach is able to close holes in places where parts connect to each other. Second, our approach  deals with shell-like parts transforming them into box-like shapes. Third, due to small alignment issues, our approach sometimes creates thin-shell surfaces in places where parts connect. Fourth, if parts with the same semantic label are connected, they become the same part. Finally, if parts have complex geometry (like office chair handles in Figure \ref{fig:preprocessing}), our approach adds a additional (small) volume to parts.

\input{fig/cond_rec_one_column}

\paragraph{Model training and evaluation} 
To obtain part code supervision, we train the part encoder-decoder on normalized parts using SDF supervision. For point cloud reconstruction, we use the PointNet implementation with residual skip connections provided by Mescheder \etal \cite{occnet} and point cloud size is 2048. For single-view reconstruction we use ResNet18 as encoder for ANISE and PQ-Net; and pretrained JLRD \cite{jldr} model provided by the authors. For PQ-Net, we train single view and point cloud reconstruction models as a regression of shape codes produced by pretrained models provided by the authors. For all experiments, we train our models with $M=10$ part predictions to ensure fair comparison with PQ-Net. 

We reconstruct meshes from from implicit representations using marching cubes \cite{lorensen1987marching}. For surface reconstruction evaluation, we use shape IoU, Chamfer distance and F-score. To ensure fair comparison, we normalize outputs of all models to match unit cube across largest dimension.
We will release preprocessed data, training and evaluation code for reproducibility.

\input{fig/part_interpolation}
\input{fig/conditional_retrieval_vs_size}

%% file: fig/comparison_pc.tex
\begin{figure*}[!ht]
\vspace{-12pt}
\begin{center}
% \begin{overpic} 
% [width=\linewidth]
% {example-image-a}
% \end{overpic}
\setlength\tabcolsep{-3pt}
\begin{tabular}{c@{\hskip 5pt}lllllllll}
\rotatebox[origin=c]{90}{Input}\vspace{-5pt} & \includegraphics[width=.13\linewidth,valign=m]{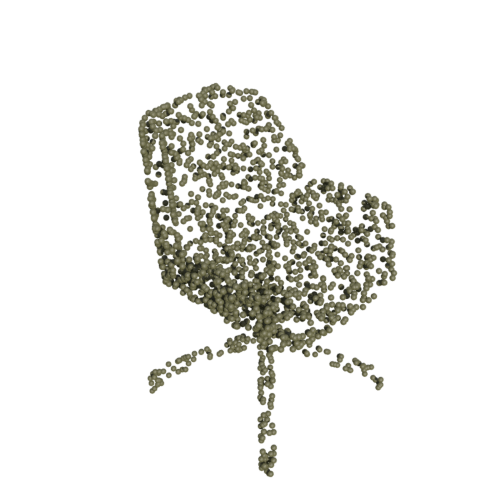} & 
\includegraphics[width=.13\linewidth,valign=m]{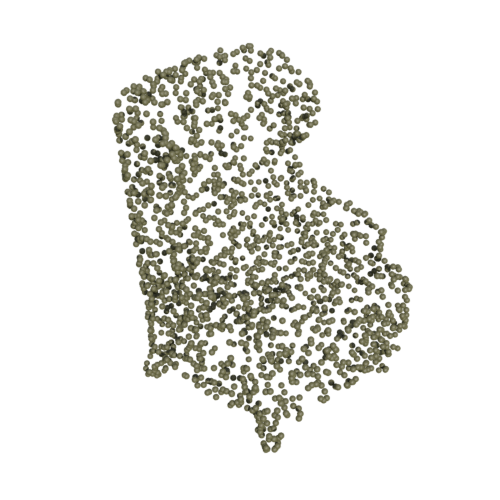} &
\includegraphics[width=.13\linewidth,valign=m]{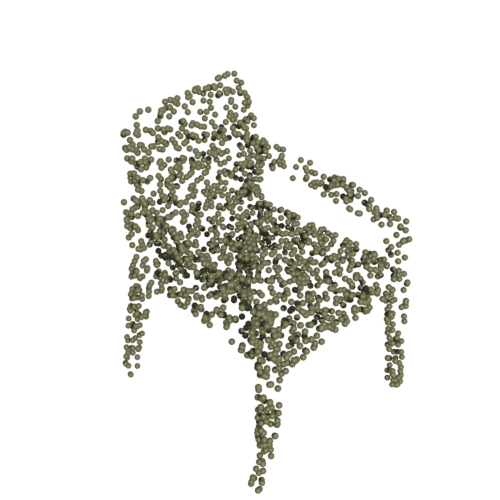} &
\includegraphics[width=.13\linewidth,valign=m]{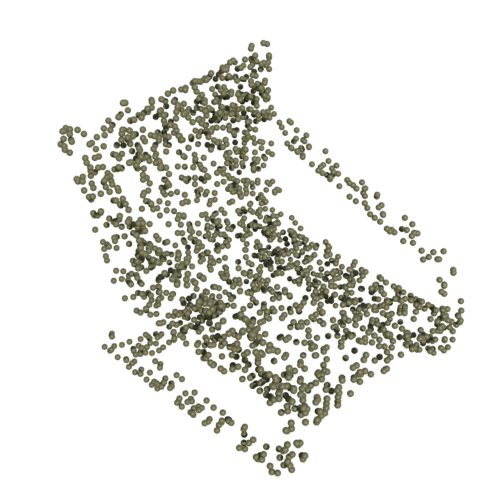} &
\includegraphics[width=.13\linewidth,valign=m]{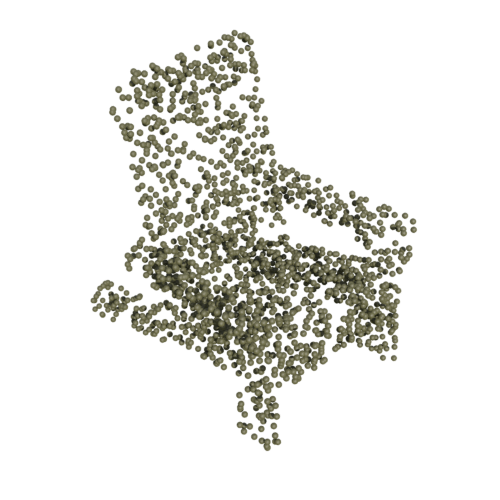} &
\hspace{-5pt}
\includegraphics[width=.13\linewidth,valign=m]{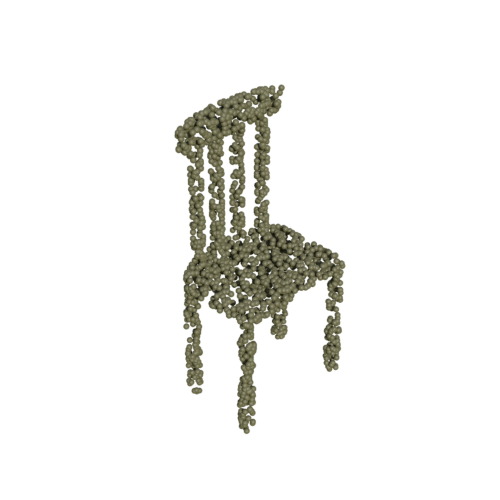} &
\includegraphics[width=.13\linewidth,valign=m]{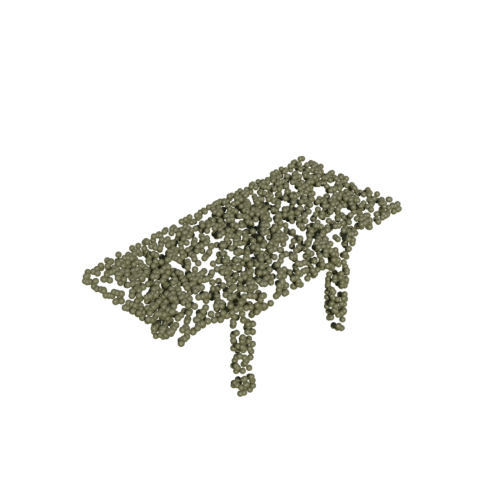} &
\includegraphics[width=.13\linewidth,valign=m]{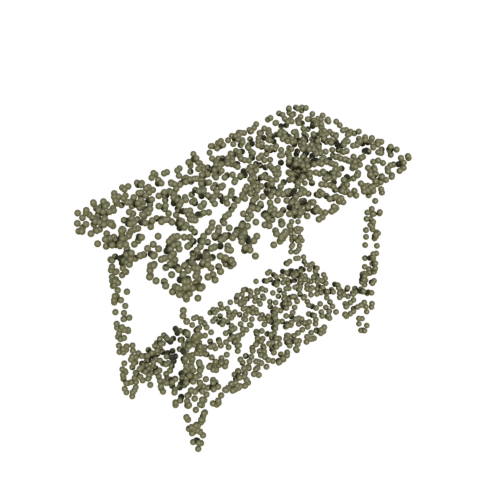} &
 \\
%\rotatebox[origin=c]{90}{JLRD} \vspace{-5pt}& \includegraphics[width=.13\linewidth,valign=m]{fig/img/comparison_pc/Chair/41744/jlrd0001.png} & 
%\includegraphics[width=.13\linewidth,valign=m]{fig/img/comparison_pc/Chair/38674/jlrd0001.png} & 
%\includegraphics[width=.13\linewidth,valign=m]{fig/img/comparison_pc/Chair/39023/jlrd0001.png} & 
%\includegraphics[width=.13\linewidth,valign=m]{fig/img/comparison_pc/Chair/39409/jlrd0001.png} & 
%\includegraphics[width=.13\linewidth,valign=m]{fig/img/comparison_pc/Chair/41284/jlrd0001.png} & 
%\hspace{-5pt}
%\includegraphics[width=.13\linewidth,valign=m]{fig/img/comparison_pc/Chair/41830/jlrd0001.png}  &
%\includegraphics[width=.13\linewidth,valign=m]{fig/img/comparison_pc/Table/26119/jlrd0001.png}  &
%\includegraphics[width=.13\linewidth,valign=m]{fig/img/comparison_pc/Table/27635/jlrd0001.png} 
 
\rotatebox[origin=c]{90}{PQ-Net}\vspace{-5pt} & \includegraphics[width=.13\linewidth,valign=m]{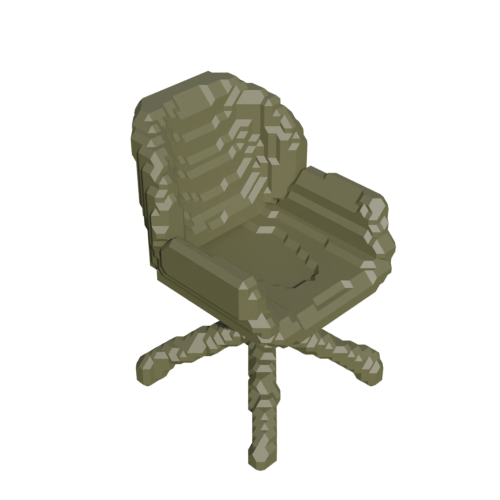}& 
\includegraphics[width=.13\linewidth,valign=m]{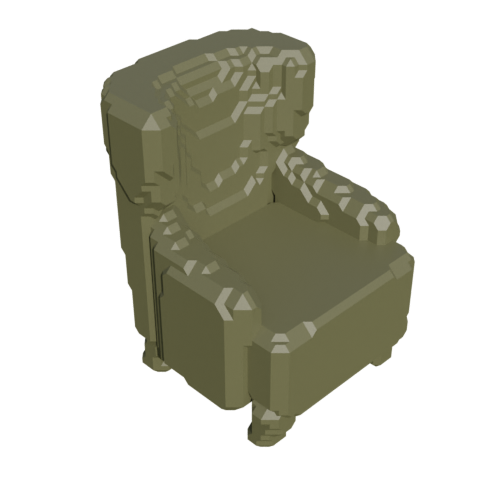}& 
\includegraphics[width=.13\linewidth,valign=m]{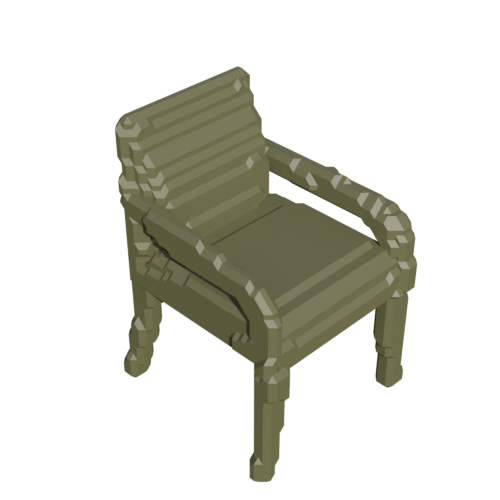}& 
\includegraphics[width=.13\linewidth,valign=m]{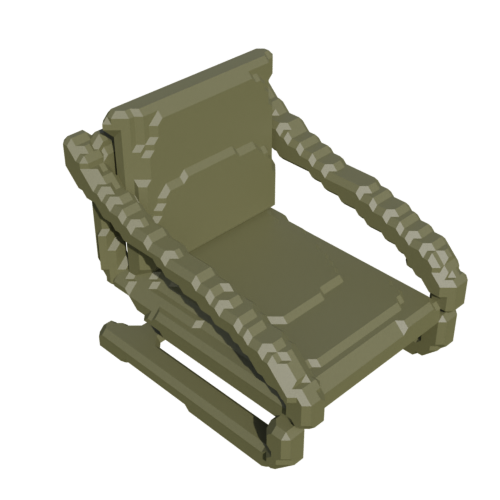}& 
\includegraphics[width=.13\linewidth,valign=m]{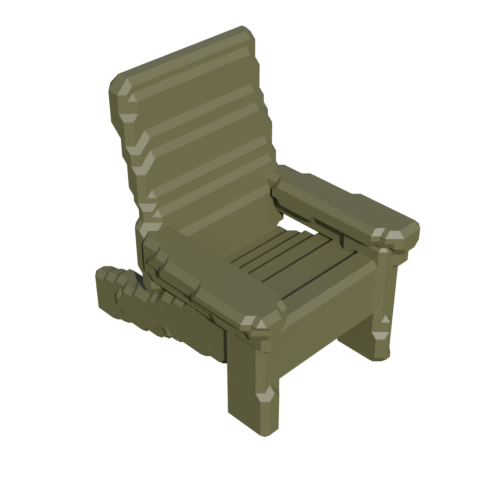}& 
\hspace{-5pt}
\includegraphics[width=.13\linewidth,valign=m]{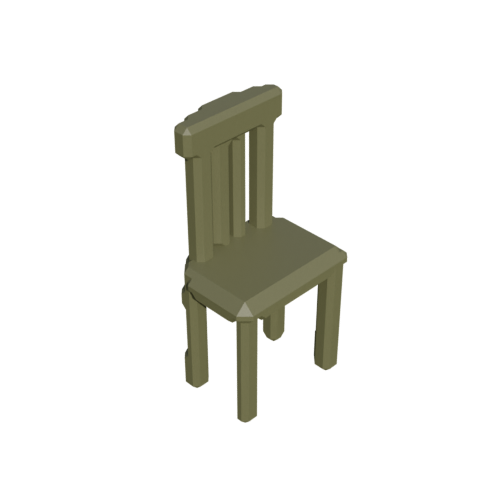}& 
\includegraphics[width=.13\linewidth,valign=m]{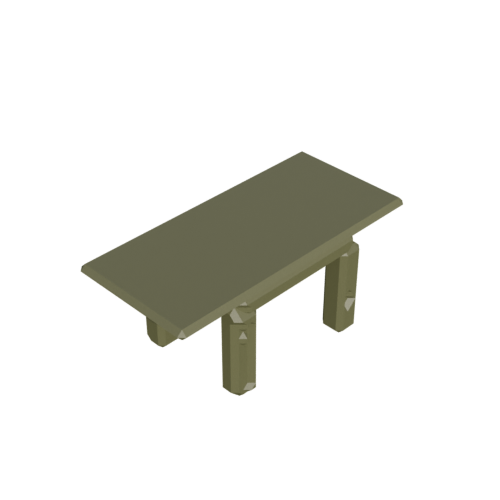} &
\includegraphics[width=.13\linewidth,valign=m]{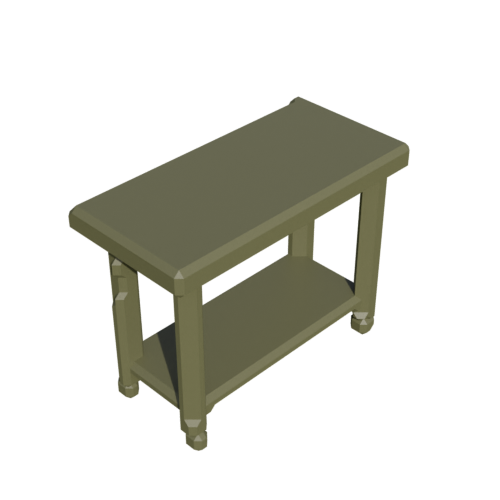} &

 \\
\rotatebox[origin=c]{90}{Ours}\vspace{-5pt} & \includegraphics[width=.13\linewidth,valign=m]{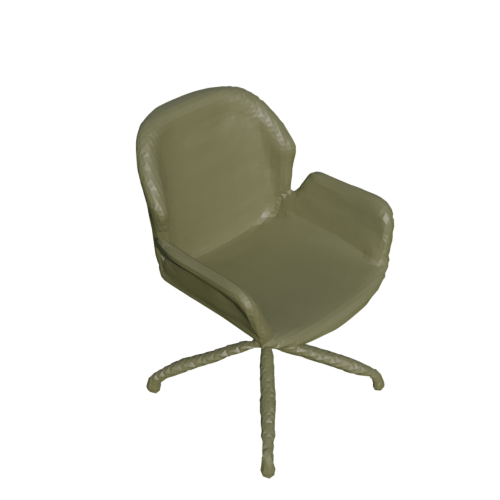} & 
\includegraphics[width=.13\linewidth,valign=m]{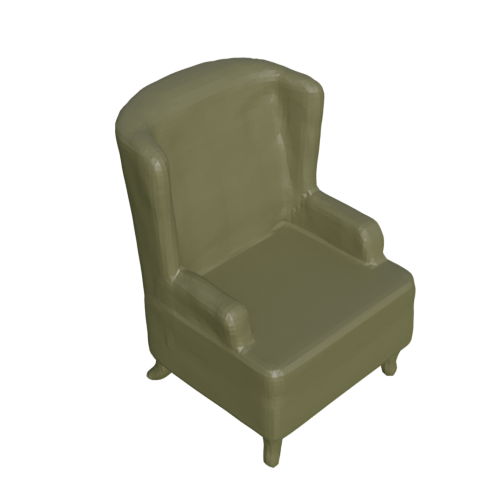} & 
\includegraphics[width=.13\linewidth,valign=m]{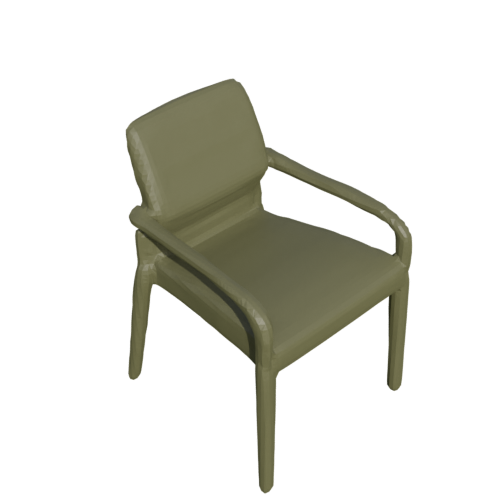} & 
\includegraphics[width=.13\linewidth,valign=m]{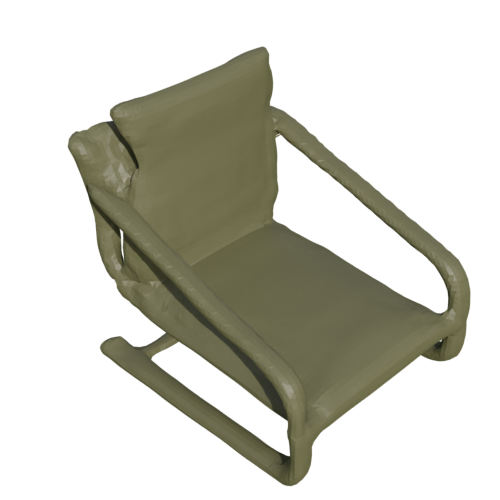} & 
\includegraphics[width=.13\linewidth,valign=m]{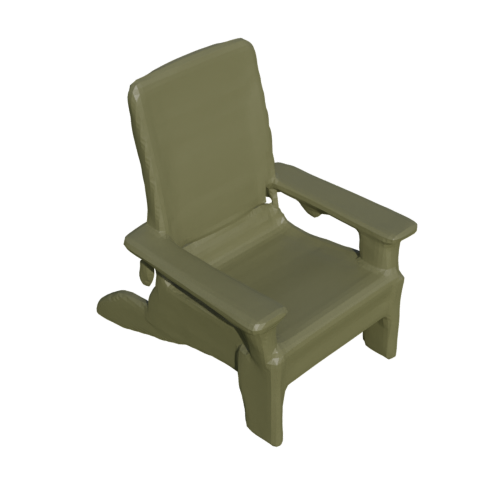} & 
\hspace{-5pt}
\includegraphics[width=.13\linewidth,valign=m]{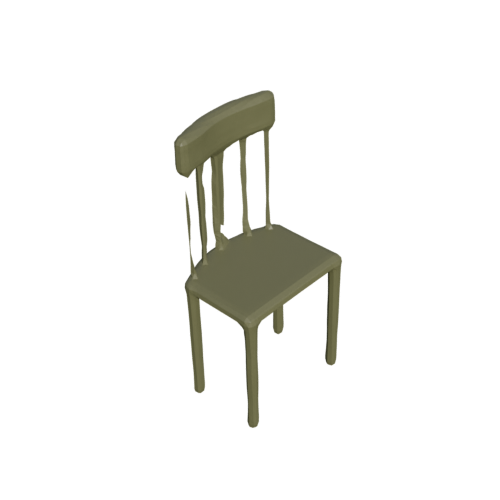} &
\includegraphics[width=.13\linewidth,valign=m]{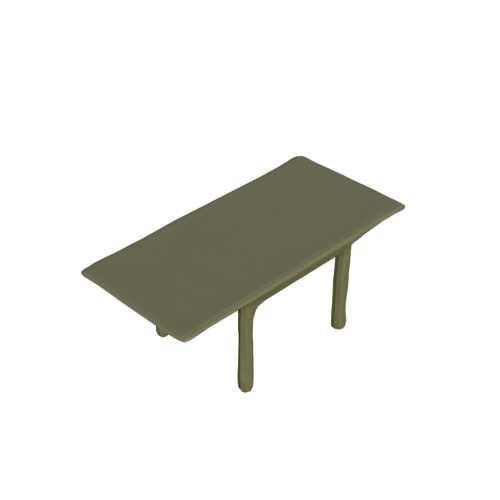} &
\includegraphics[width=.13\linewidth,valign=m]{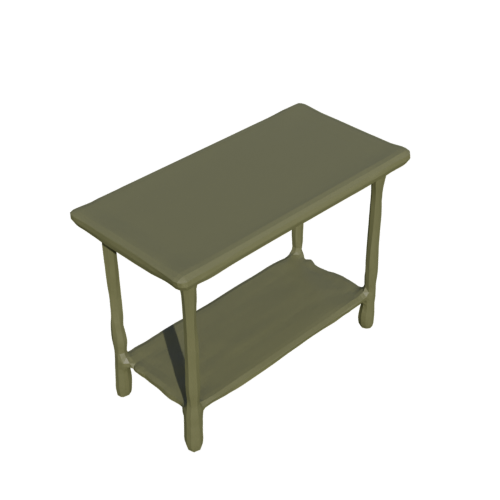} &

 \\
\rotatebox[origin=c]{90}{GT}\vspace{-5pt} & \includegraphics[width=.13\linewidth,valign=m]{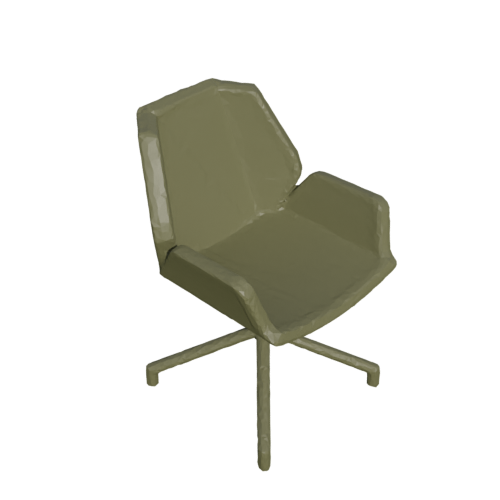} &
\includegraphics[width=.13\linewidth,valign=m]{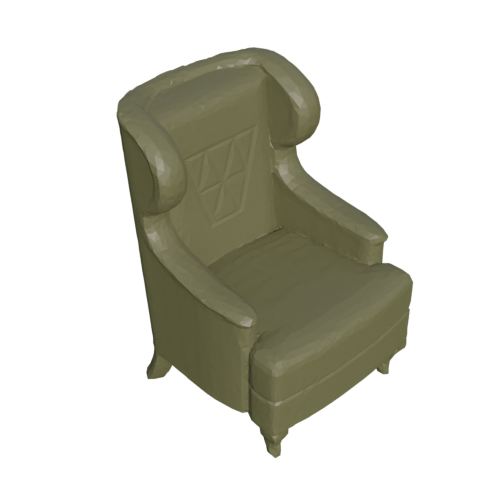} & 
\includegraphics[width=.13\linewidth,valign=m]{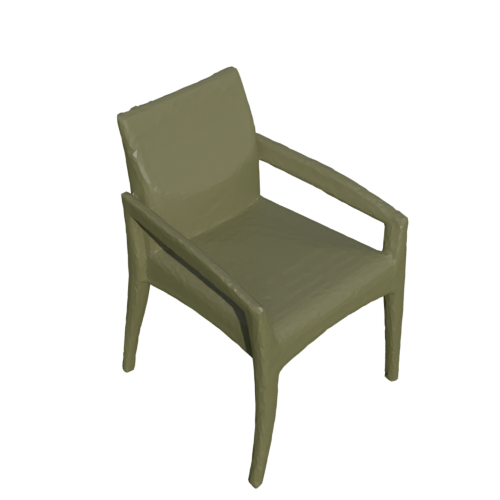} & 
\includegraphics[width=.13\linewidth,valign=m]{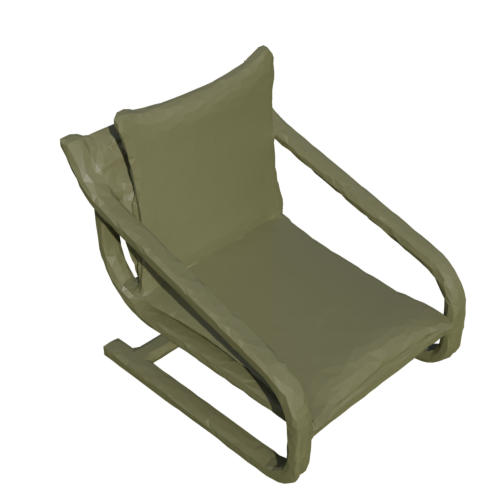} & 
\includegraphics[width=.13\linewidth,valign=m]{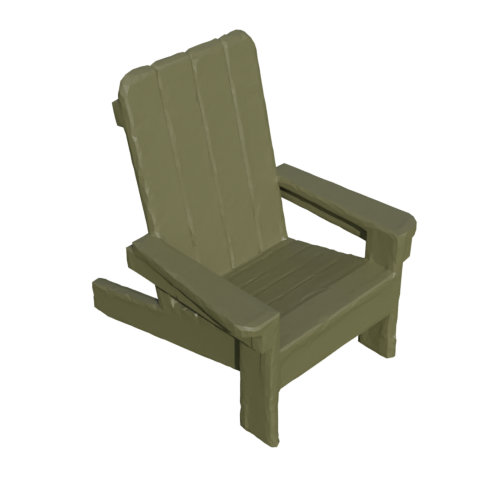} & 
\hspace{-5pt}
\includegraphics[width=.13\linewidth,valign=m]{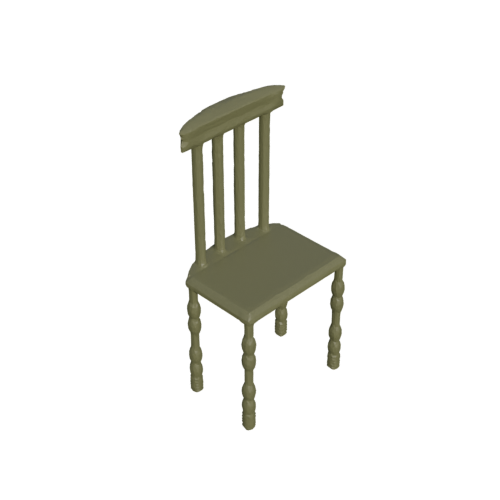} &
\includegraphics[width=.13\linewidth,valign=m]{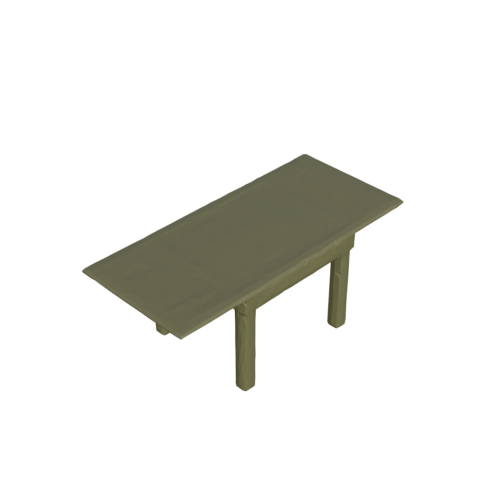} &
\includegraphics[width=.13\linewidth,valign=m]{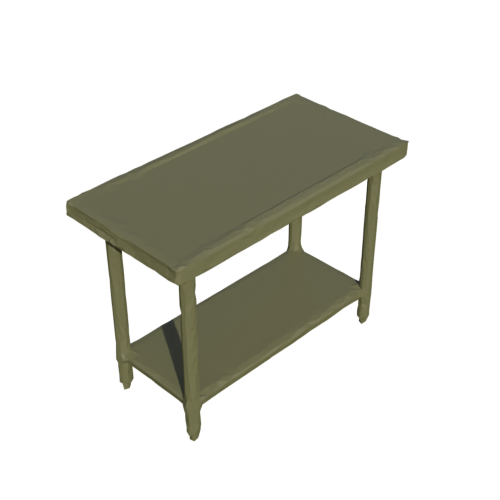} &

 \\

\end{tabular}

\end{center}
\vspace{-6pt}
\caption{
\textbf{Point cloud reconstruction.}
Note that our model is able to reconstruct important object regions: thin stripes on chair back, curved armchair handles, swivel chair legs, etc. PQ-Net reconstructions are shown for PQ-Net-voxel version (Section \ref{res:point_reconstruction}).}
%\vspace{-16pt}
\label{fig:comparison}
\end{figure*}

%% file: fig/cond_rec_one_column.tex
\begin{figure*}[t]
\begin{center}
% \begin{overpic} 
% [width=\linewidth]
% {example-image-a}
% \end{overpic}
\includegraphics[width=\linewidth]{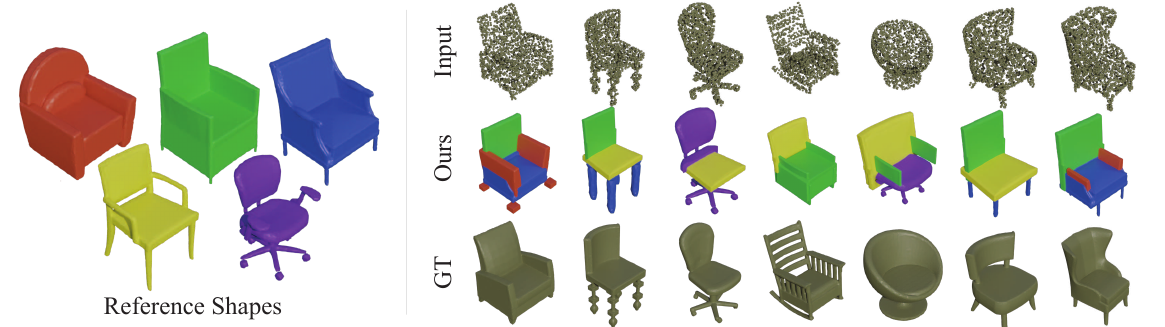}

\end{center}
%\vspace{-3mm}
\caption{
\textbf{Part-constrained shape assembly.}  
Our model is capable of reconstructing sparse point clouds by assembling parts from a user-provided reference database of shapes or parts. In this example, the user selects $5$ shapes (left) pre-segmented into their parts. Our model reconstructs shapes that 
closely approximate the input point clouds while \emph{only using parts from the reference shapes}.
Differently from traditional shape retrieval-based methods~\cite{jldr,productnet,imagepurification} our approach can mix parts
from different shapes to better approximate the target. Best viewed in color.
}
\label{fig:cond_rec2}
\end{figure*}

%% file: fig/part_interpolation.tex
\begin{figure}[t]
\begin{center}
% \begin{overpic} 
% [width=\linewidth]
% {example-image-a}
% \end{overpic}
\includegraphics[width=1.0\linewidth]{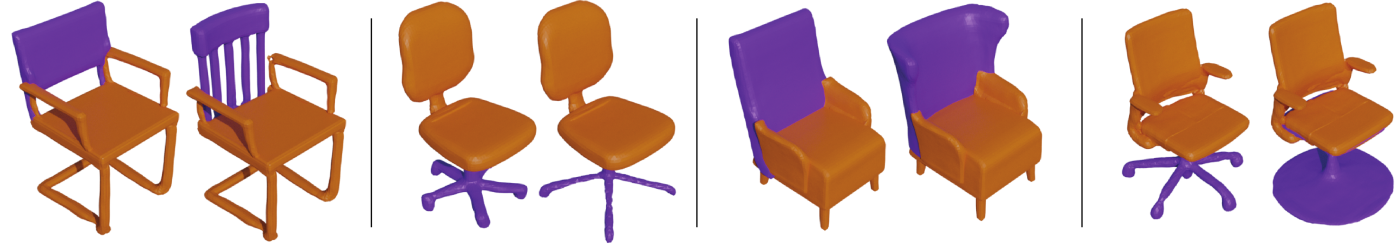}
\end{center}
\vspace{-12pt}
\caption{
\textbf{Part editing.} Given a query shape, users can replace or interpolate a particular part to produce a full shape with modified parts. Left shapes: original reconstruction. Right shapes: edited reconstruction.}
%All results are shown for non-finetuned model.}
\label{fig:part_interpolation}
\end{figure}

%% file: fig/conditional_retrieval_vs_size.tex
\begin{figure}[t]
\begin{center}
% \begin{overpic} 
% [width=\linewidth]
% {example-image-a}
% \end{overpic}
\includegraphics[width=0.85\linewidth]{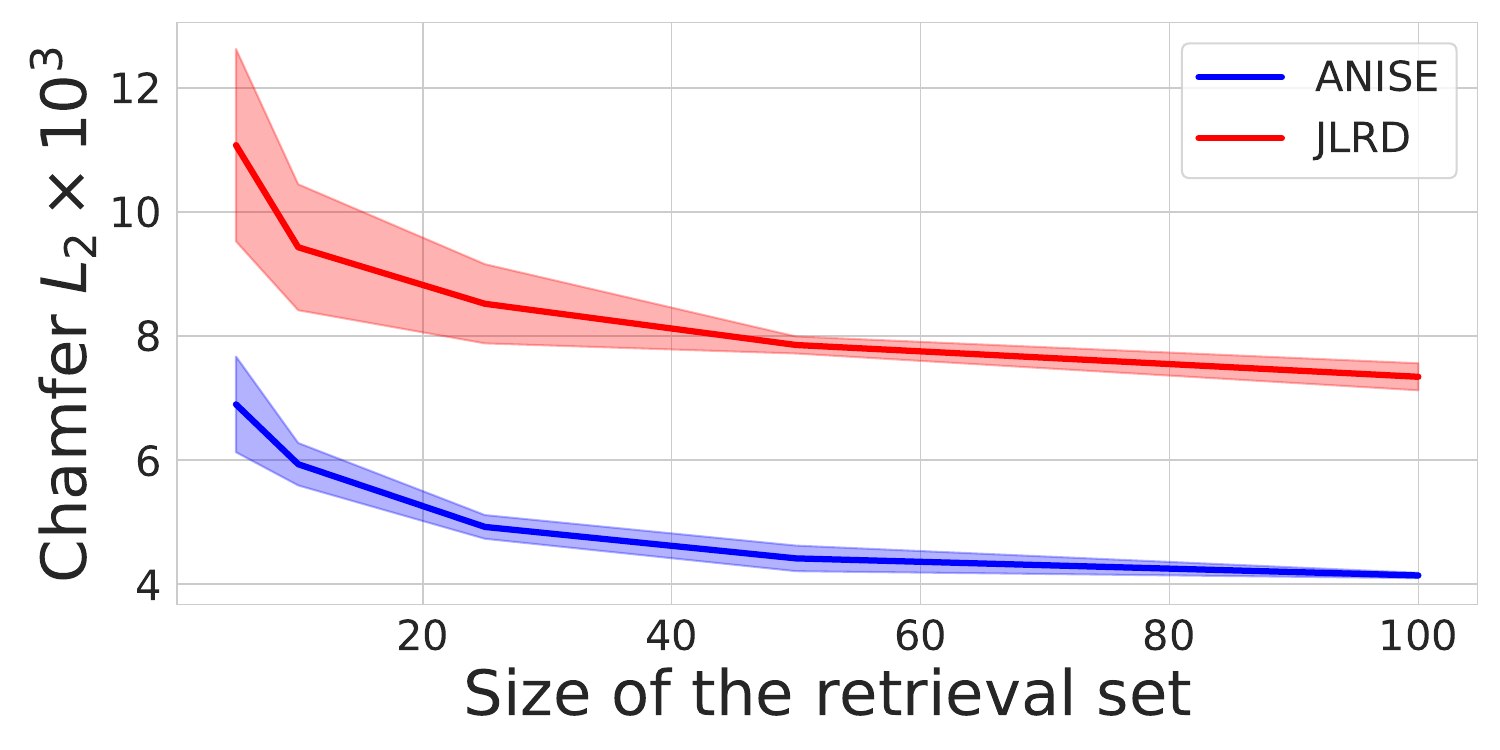}
\end{center}
\vspace{-3mm}
\caption{
\textbf{Retrieval-based SVR quality vs. size of the reference dataset.} We compare part retrieval-based (ANISE PR\&A) model to shape retrieval and deformation (JLRD) approach depending on the size of retrieval dataset ($K = 5, 10, 25, 50, 100$). For each $K$, we perform $5$ experiments and depict the mean value and standard deviation. Results are reported for Chair category.  
%All results are shown for non-finetuned model.
}
\label{fig:cond_retrieval_vs_size}
\end{figure}

%% file: sec/5_results.tex
\subsection{Single view reconstruction}
\label{res:svr}

\paragraph{ANISE vs part-agnostic methods} 
We compare ANISE to part-agnostic single view reconstruction methods: DISN \cite{disn} and D2IM-Net \cite{li2021d2im} in Table \ref{tab:d2im_comparison}. Quantitatively, our model outperforms DISN by a large margin and performs similarly to D2IM-Net. Qualitative results (Figure \ref{fig:comparison_svr_combined}) show that ANISE is better at small part accentuation but sometimes slightly incorrectly estimates proportions of the shape. 
Notice that the type of supervision for the methods compared here are very different.
Unlike ANISE, DISN and D2IM use the whole ShapeNet dataset and have access to object pose annotations.
On the other hand, our method is only trained on PartNet categories while using their part segmentations.
We believe that combining DISN and D2IM representations with ANISE might yield further performance gains for single view reconstruction.

\paragraph{ANISE vs. part-aware methods}
\label{res:svr_part_aware_comparison} We evaluate our method versus part-aware reconstruction methods: PQ-Net-regr (regression of PQ-Net embeddings from image encodings) \cite{pqnet} and JLRD \cite{jldr} in Table \ref{tab:rgb_reconstruction}. ANISE outperforms both approaches in terms of volumetric  (IoU) and surface-based (Chamfer) measures for Chair and Table categories. Poor performance of PQ-Net for Table category is due to the fact that it predicts extra legs that are badly positioned thus significantly increasing Chamfer distance. Qualitative analysis (Figure \ref{fig:comparison_svr_parts} ) shows that even though PQ-Net-regr and ANISE exhibit similar priors, our model is significantly better in recovering fine shape details. 

\paragraph{ANISE PR\&A vs. retrieval-based reconstruction}
We compare our PR\&A approach to a deformation and retrieval-based approach (JLRD). Our approach outperforms it by a significant margin (see Table~\ref{tab:rgb_reconstruction}). We demonstrate this in additional experiment where we limit size of the retrieval database to a certain size $K$ (see Figure~\ref{fig:cond_retrieval_vs_size}).
PR\&A based reconstruction achieves the same performance as JLRD with a reference dataset of only 5 shapes.   The gap in performance is due to fact that JLRD retrieves a full shape, then deforms its parts. Since it cannot adjust the shape structure, it is limited to the retrieved shape structure and topology. ANISE does not suffer from this limitation: it does part-based retrieval instead and can utilize all possible combinations of parts provided in the database.

\subsection{Point cloud reconstruction} 
\label{res:point_reconstruction}

\paragraph{ANISE vs part-aware methods}
We compare our approach with part-aware PQ-Net \cite{pqnet} that utilizes sequential part assembly. Since original model was trained on voxel grids of size 64, we train a separate model that regresses PQ-Net representations from point clouds (PQ-Net-regr). This regression model follows the same point cloud network architecture employed by ANISE. Our approach clearly outperforms PQ-Net-regr even when compared to their original voxel encoding approach (PQ-Net-voxel in Table~\ref{tab:sota}), both quantitatively (Table \ref{tab:sota}) and qualitatively (Figure \ref{fig:comparison}).

\label{res:retrieval_reconstruction}

\input{tab/d2im_comparison}

\input{tab/rgb_reconstruction}

\input{tab/sota}

\input{tab/hierarchy_quality}

%\input{tab/jlrd_comparison}

%\input{tab/conditional_reconstruction_numbers}

\input{tab/ablations}

\input{fig/ablation_quality}

\input{fig/supplementary/scratch_comparison}

\paragraph{Part editing and constrained shape assembly} Using transformations and part codes predicted by our model, we can also constrain the shape assembly to parts from a small database. This is done by replacing the predicted part codes with closest one in the database. In Figure~\ref{fig:cond_rec2}, we illustrate this application with an example where only the parts from $5$ reference shapes are used to perform point cloud reconstruction. Model correctly guesses the shape structure and selects the appropriate parts from reference shapes. In Figure \ref{fig:part_interpolation} we also demonstrate part editing capabilities of our model through modification of the part codes. 

\paragraph{ANISE vs hierarchical methods} 
As an alternative evaluation, we compare with the StructureNet method \cite{structurenet}. StructureNet encodes the shape structure as a hierarchical graph where nodes represent parts, and edges represent adjacency, symmetry, and hierarchical relationships between parts, thus, it encodes wealthier structural information compared to ANISE. One question is how structurally consistent ANISE is compared to such method. For structural consistency evaluation, we compare how well the reconstructed part set matches with the ground-truth one. To this end, we compute part precision (PP) and part recall (PR) as follows: 
$$
PP=\frac{|P_M|}{|P_{PRED}|}, PR=\frac{|P_M|}{|P_{GT}|}
$$ 
where $P_M$ is set of matching parts between ground truth and predicted parts; $P_{PRED}$ and $P_{GT}$ are sets of predicted and ground truth parts respectively. For part matching, we follow StructureNet \cite{structurenet}'s pipeline and compute it via Hungarian matching using the Chamfer distance between parts as similarity metric. The Chamfer distance is computed using point clouds of size $1000$ points that are sampled from $P_{PRED}$ and $P_{GT}$. For fair comparisons, we  use StructureNet's \cite{structurenet} test split which is downsampled from the PartNet split due to the fact the StructureNet cannot process shapes that have more than 10 children in any node of part hierarchy. We use the pre-trained models provided by the authors in point cloud auto-encoding setting.

Table \ref{tab:hier_quality} reports the comparison with StructureNet. ANISE shows significantly better surface reconstruction quality in terms of Chamfer distance for both chairs and tables. For chairs, the precision is better, yet the recall is lower. For tables, its precision is comparable to StructureNet, yet its recall is lower. The results indicate that ANISE benefits from the its use of implicit functions to capture surface geometry compared to StructureNet. Both methods have comparable precision i.e., they tend not to generate spurious parts that are unmatched with the ground-truth. Yet, compared to StructureNet, ANISE may miss more ground-truth parts during reconstruction. We suspect that this is due to the issue that parts with significantly overlapping volumes might end up being merged. A more hierarchical approach, as used in StructureNet, could alleviate this problem.

\label{res:ablation}

\paragraph{Ablation study} Justifications for the design choices in ANISE are summarized in Table \ref{tab:ablations}. Part geometry prediction conditioned on 
part transformations used by ANISE (``Cond. pred.'' column)
is compared to its unconditional counterpart that predicts parts and transformation in parallel, similar to PQ-Net. Our results indicate that this small changes provides a significant boost to reconstruction quality. Furthermore, supervision from SDF of fully assembled shapes gives significant boost to reconstruction quality according to all evaluated metrics (``Fine-tuning'' column).

We provide qualitative evaluation of fine-tuning stage importance in Figure \ref{fig:ablation_quality}. Even though model without SDF fine-tuning produces reasonable looking shapes, their proportions are usually way off which especially hurts reconstruction of smaller parts. It is also interesting that our model is able to generalize over preprocessing artifacts (row 3 in Figure \ref{fig:ablation_quality}).

We also compare our multi-stage training described in Section \ref{sec:training} to end-to-end ANISE training from ``scratch'' without relying on the intermediate stages.
%which doesn't use intermediate training for part and position codes' predictions. 
When training from scratch, ANISE ``collapses'' to predicting only one part representing the whole shape while the rest of the parts result in either tiny or no surfaces.
%to model which consistently uses one of $M$ part codes to encode full shape while turning other parts into negligible surfaces.
Figure \ref{fig:scratch_comparison} demonstrates this effect.

%% file: tab/d2im_comparison.tex
\begin{table}
\centering
\caption{
\textbf{Comparison with part-agnostic single view reconstruction.} Our model achieves similar performance to state-of-the-art part unaware methods. IoU is computed using $64^3$ resolution and multiplied by $10^3$. Chamfer distance (CD) is multiplied by $10^3$.}
\resizebox{\linewidth}{!}{ %< auto-adjusts font size to fill line
\begin{tabular}{@{}lccccccc@{}}
\toprule
& \multicolumn{2}{c}{Chair} & \multicolumn{2}{c}{Table} & \multicolumn{2}{c}{Lamp}\\
Method & IoU ($\uparrow$) & CD($\downarrow$) & IoU ($\uparrow$) & CD ($\downarrow$) & IoU ($\uparrow$) & CD ($\downarrow$)  \\
\midrule
%PQ-Net 256 && - & 2.86 & - & 4.50\\
%IMNET &54.1&   3.76 & 50.1 & 4.32 & 41.7 & 5.57 \\
%DISN & 55.4 & 3.26 &  56.3 & 3.16 & 37.9 & 7.70 \\ 
%Ours (retr) & 50.1 & 3.74 &  33.4 & 45.3 & 3.91 & 53.3 \\ 
%D2IM-Net & 56.1 & 3.29 & 53.7 & 3.56 & 42.1 & 5.57 \\ 
%D2IM-Net GL  & 54.4 & 3.08 & 54.8 & 3.10 & 45.0 & 5.49 \\ 
%\midrule
DISN \cite{disn} & 46.6 & 4.21 & 41.0 & 6.58 & 19.54 & 16.29 \\ 

D2IM-Net \cite{li2021d2im} & 54.3 & \textbf{2.81} & 47.4 & 3.96 & 31.9 & \textbf{6.66} \\ 

\midrule
Ours (no f-tuning) & \textbf{56.3} & 3.04 &  \textbf{48.9} & \textbf{3.59} & \textbf{38.5} & 8.09\\ 
\bottomrule
\end{tabular}
} % \resizebox

% }
 % \caption
%\vspace{-7mm}
\label{tab:d2im_comparison}
\end{table}

%% file: tab/rgb_reconstruction.tex
\begin{table}
\centering
\caption{
\textbf{Part-aware single-view reconstruction.} Our method significantly outperforms both part-based (PQ-Net) and retrieval and deformation based (JLRD) approaches. IoU is computed using 64$^3$ resolution and multiplied by $10^2$. Chamfer distance (CD) is multiplied by $10^3$. 
} 
\resizebox{\linewidth}{!}{ %< auto-adjusts font size to fill line
\begin{tabular}{@{}lccccccccc@{}}
\toprule
& \multicolumn{3}{c}{Chair} & \multicolumn{3}{c}{Table} \\
Method & IoU ($\uparrow$) & CD ($\downarrow$) & F1 ($\uparrow$)&  IoU ($\uparrow$) & CD ($\downarrow$) & F1 ($\uparrow$)  \\
\midrule
%PQ-Net 256 && - & 2.86 & - & 4.50\\
JLRD~\cite{jldr} & 31.1 &  5.09 & 48.1 &  29.4 & 4.92 & 58.0 \\
Ours (PR\&A)  & 50.6 & 3.55 & 60.5 &  44.3 & 5.35 & 64.2\\ 
\midrule
PQ-Net-regr~\cite{pqnet} & 47.0 & 4.27 & 50.2 & 29.7 & 18.41 & 32.1  \\ 
Ours  & \textbf{56.7} & \textbf{2.99} & \textbf{67.0} &   \textbf{57.4} & \textbf{2.54} & \textbf{77.8} \\ 
\bottomrule
\end{tabular}
} % \resizebox
% \caption
%\vspace{-7mm}
\label{tab:rgb_reconstruction}
\end{table}

%% file: tab/sota.tex
\begin{table}
\centering
\caption{
\textbf{Part-aware point cloud reconstruction results.} IoU is computed using 64$^3$ resolution and multiplied by $10^2$. Chamfer distance (\textbf{CD}) is multiplied by $10^3$.} % \caption

\resizebox{\linewidth}{!}{ %< auto-adjusts font size to fill line
\begin{tabular}{@{}lcccccc@{}}
\toprule
& \multicolumn{2}{c}{Chair} & \multicolumn{2}{c}{Table}  \\
Method & IoU ($\uparrow$) & CD ($\downarrow$) & IoU ($\uparrow$) & CD  ($\downarrow$) \\
\midrule
%IM-Net  Auto Enc & 62.9 & 3.64 & 56.1 & 6.79 & 39.56 & 12.43\\
%IM-Net 256 && - & 3.59 & - & 6.31 \\
PQ-Net-voxel~\cite{pqnet}  & 67.3 & 3.38 & 47.4 & 5.49 \\ % \\&\\ \textbf{41.29} & 11.49\\
\midrule

PQ-Net-regr~\cite{pqnet}  & 46.0 & 4.21 & 29.8 & 27.26 \\
%PQ-Net (recon) & 45.9 & 4.21 & - & - \\

%PQ-Net 256 && - & 2.86 & - & 4.50\\
Ours & \textbf{69.8} & \textbf{1.76} & \textbf{72.7} & \textbf{2.35} \\ 
\bottomrule
\end{tabular}
} % \resizebox
%\vspace{-5mm}
\label{tab:sota}
\end{table}

%10 parts
%chamfer_l2_rescaled 0.0017639687710895905
%fscore 0.8125034630377445
%fscore_rescaled 0.7099743492111897
%iou_voxel_64 0.7327848803613374
%iou_voxel_64_rescaled 0.6981010514466569

%% file: tab/hierarchy_quality.tex
{\color{red}\begin{table}
\centering
\caption{
\textbf{Structural consistency comparison for point cloud reconstruction.} Part precision (\textbf{PP}) and part recall (\textbf{PR}) are multiplied by $10^2$. Chamfer distance (\textbf{CD}) is multiplied by $10^3$.} % \caption

\resizebox{\linewidth}{!}{ %< auto-adjusts font size to fill line
\begin{tabular}{@{}lcccccccc@{}}
\toprule
& \multicolumn{3}{c}{Chair} & \multicolumn{3}{c}{Table}  \\
Method & PR ($\uparrow$) &\ PP ($\uparrow$) & CD ($\downarrow$) & PR ($\uparrow$) & PP ($\uparrow$) &  CD  ($\downarrow$) \\
\midrule
%IM-Net  Auto Enc & 62.9 & 3.64 & 56.1 & 6.79 & 39.56 & 12.43\\
%IM-Net 256 && - & 3.59 & - & 6.31 \\
StructureNet PC-AE~\cite{structurenet}  &  94.3 & \textbf{95.3}  & 5.31 & \textbf{92.2} & \textbf{91.6} & 12.8 \\ % \\&\\ \textbf{41.29} & 11.49\\
%\midrule
%PQ-Net-regr~\cite{pqnet}  & 96.0 & 96.8 & 4.03 & 64.3 & 92.2 & 23.67\\
%PQ-Net (recon) & 45.9 & 4.21 & - & - \\

%PQ-Net 256 && - & 2.86 & - & 4.50\\
Ours & \textbf{96.5} & 92.5 & \textbf{1.49} & 92.0  &  83.9 & \textbf{3.31} \\ 
\bottomrule
\end{tabular}
} % \resizebox
%\vspace{-5mm}
\label{tab:hier_quality}
\end{table}}

%10 parts
%chamfer_l2_rescaled 0.0017639687710895905
%fscore 0.8125034630377445
%fscore_rescaled 0.7099743492111897
%iou_voxel_64 0.7327848803613374
%iou_voxel_64_rescaled 0.6981010514466569

%% file: tab/ablations.tex
%\begin{table}
%\centering
%\resizebox{\linewidth}{!}{ %< auto-adjusts font size to fill %line
%\begin{tabular}{@{}lcc|ccc@{}}
%\toprule
%& IoU & Chamfer & Precision & Coverage & F-Score \\
%\midrule
%-cond. prediction & - & - & - \\
%-fine-tuning & - & - & - & - \\
%-learned latents & - & - & - & - \\
%\midrule
%Occupancy & - & - & - \\ 
%SDF & - & - & - \\ 
%\bottomrule
%\end{tabular}
%} %< \resizebox
%\caption{
% 
%\textbf{Ablation studies}. Results for the chairs category. %Let's add F-Score here. 
%IMO that is the more reliable metric and having
%it in the ablation will allow other people to compare with us.
%We can add/remove comparisons, these are some I think are worth discussing.
% 
%} % \caption
%\label{tab:ablations}
%\end{table}

\begin{table}
\caption{\textbf{Ablation}. Results are based on Chair category experiments for point cloud reconstruction. IoU and F-score are multiplied by $10^2$, Chamfer distance is multiplied by $10^3$.
}
\centering
\resizebox{\linewidth}{!}{ %< auto-adjusts font size to fill line
\begin{tabular}{@{}cc|ccc@{}}
\toprule
%\multicolumn{2}{c|}{Design choice} & \multicolumn{3}{c}{Metric} \\
 Cond. pred. & 
% AE latents & 
 Finetuning & IOU ($\uparrow$) & Chamfer ($\downarrow$) & F-score($\uparrow$) \\
\midrule
 - & -  & 64.4 & 2.81 & 57.4 \\
\checkmark & - & 61.4 & 2.44 & 62.0\\
%\checkmark & - & -  & Maybe later \\
\checkmark & \checkmark  & 70.9 & 1.69 & 73.9 \\
\bottomrule
\end{tabular}
} %< \resizebox
 % \caption
%\vspace{-5mm}
\label{tab:ablations}
\end{table}

%% file: fig/ablation_quality.tex
\begin{figure}[t]
\begin{center}
% \begin{overpic} 
% [width=\linewidth]
% {example-image-a}
% \end{overpic}
\includegraphics[width=\linewidth]{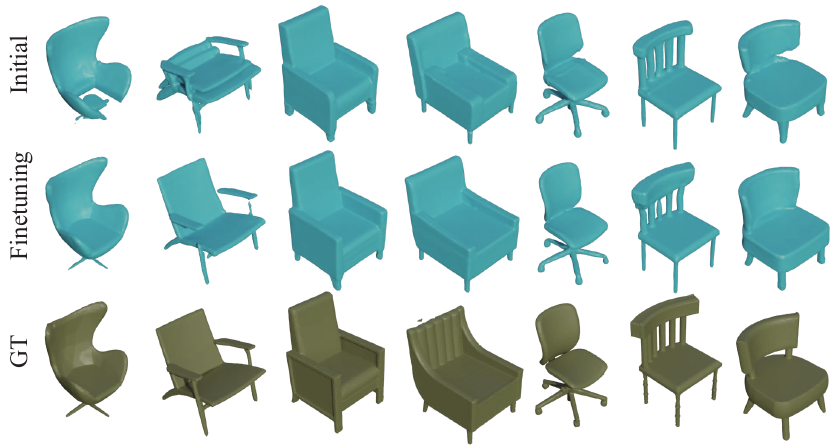}\end{center}
\vspace{-3mm}
\caption{
\textbf{Implicit function fine-tuning stage:} The supervision from the full shape implicit function helps our model to better capture part proportions and fine details of geometry. 
}
\vspace{-3mm}
\label{fig:ablation_quality}
\end{figure}

%% file: fig/supplementary/scratch_comparison.tex
\begin{figure*}[!ht]
\begin{center}
% \begin{overpic} 
% [width=\linewidth]
% {example-image-a}
% \end{overpic}
\setlength\tabcolsep{-10pt}
\begin{tabular}{c@{\hskip 5pt}llllllllll}
\rotatebox[origin=c]{90}{Input}\vspace{-7pt} & \includegraphics[width=.15\linewidth,valign=m]{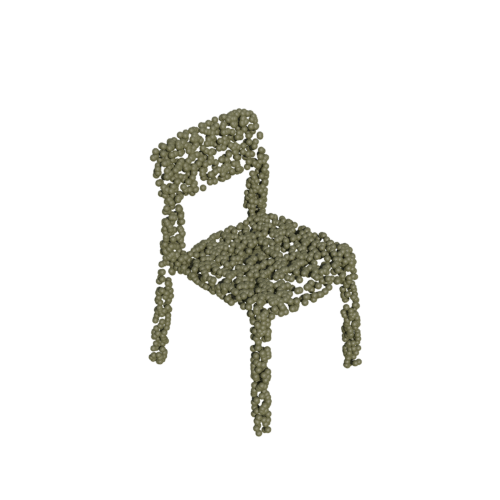} & \includegraphics[width=.15\linewidth,valign=m]{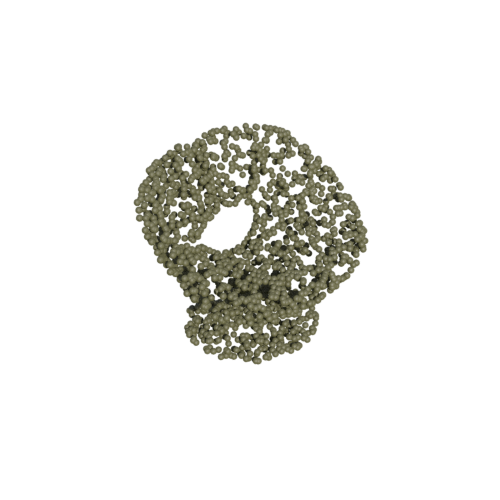} & 
\includegraphics[width=.15\linewidth,valign=m]{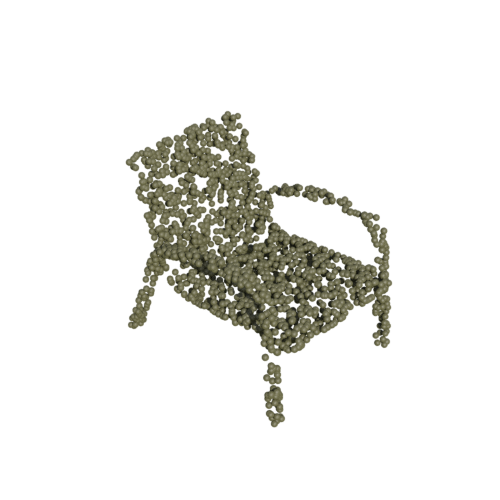} & 
\includegraphics[width=.15\linewidth,valign=m]{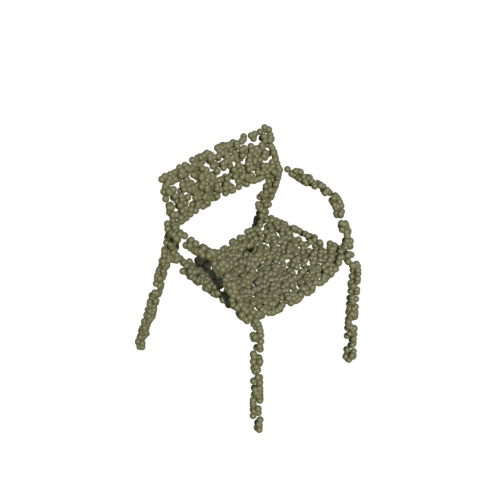} &
\includegraphics[width=.15\linewidth,valign=m]{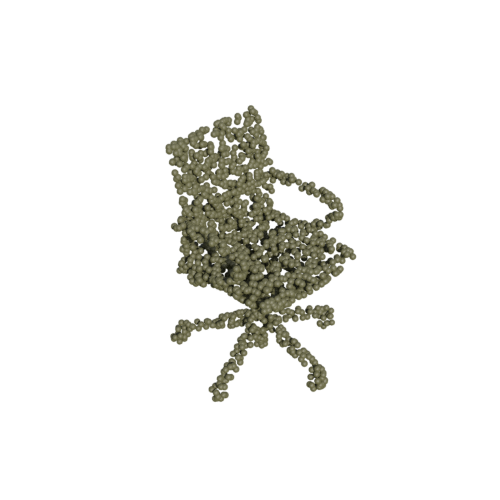} &
\includegraphics[width=.15\linewidth,valign=m]{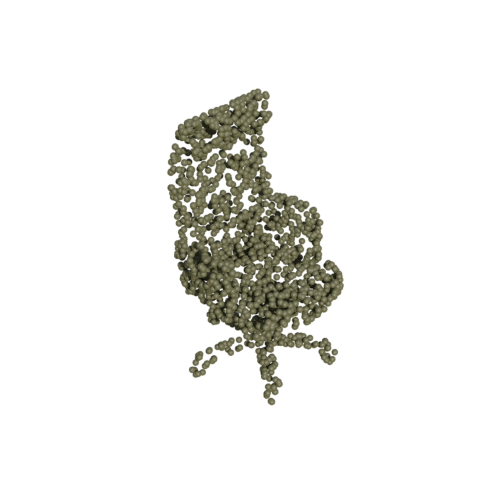} &
\includegraphics[width=.15\linewidth,valign=m]{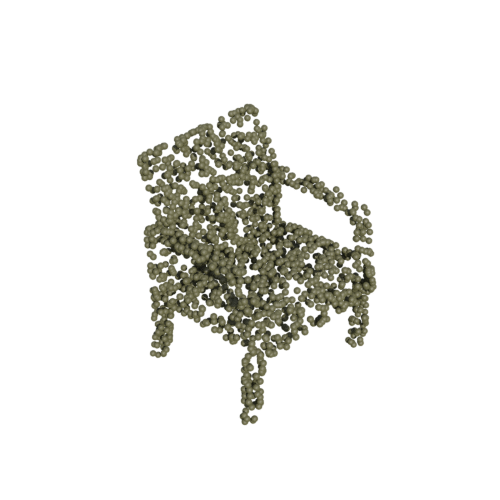} &
\includegraphics[width=.15\linewidth,valign=m]{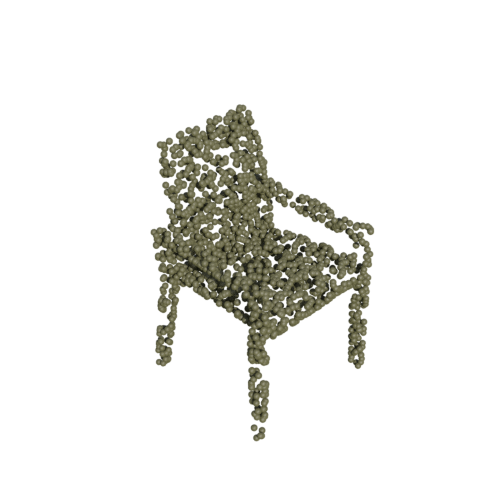} &
\includegraphics[width=.15\linewidth,valign=m]{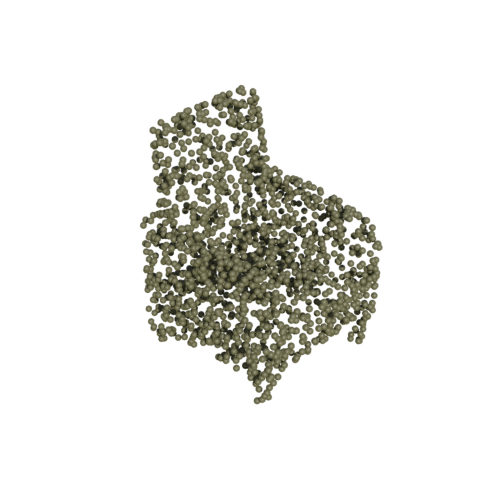} &
 \\
\rotatebox[origin=c]{90}{Scratch}\vspace{-7pt} & \includegraphics[width=.15\linewidth,valign=m]{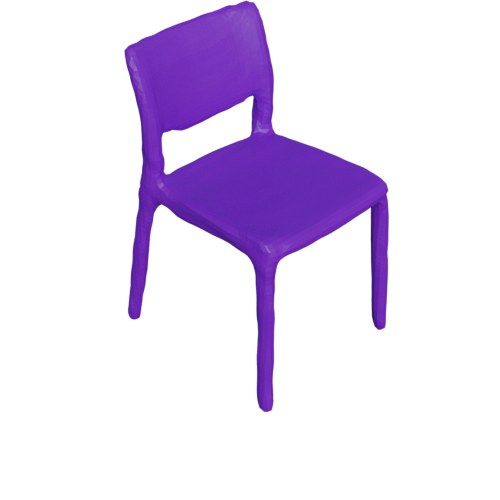} & \includegraphics[width=.15\linewidth,valign=m]{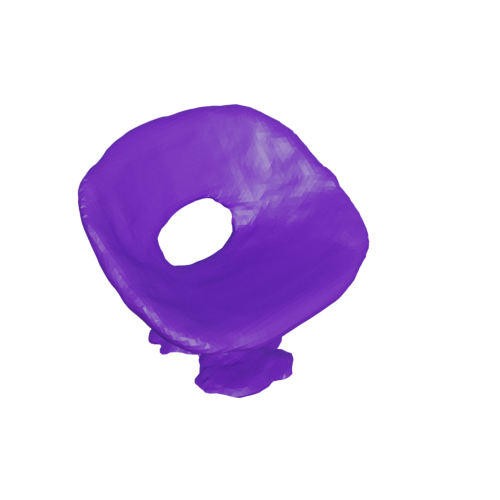} & 
\includegraphics[width=.15\linewidth,valign=m]{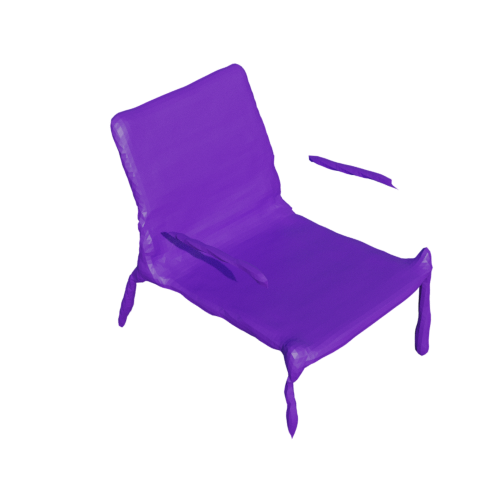} & 
\includegraphics[width=.15\linewidth,valign=m]{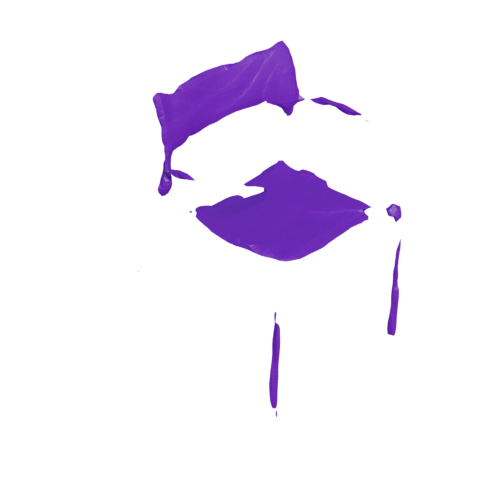} &
\includegraphics[width=.15\linewidth,valign=m]{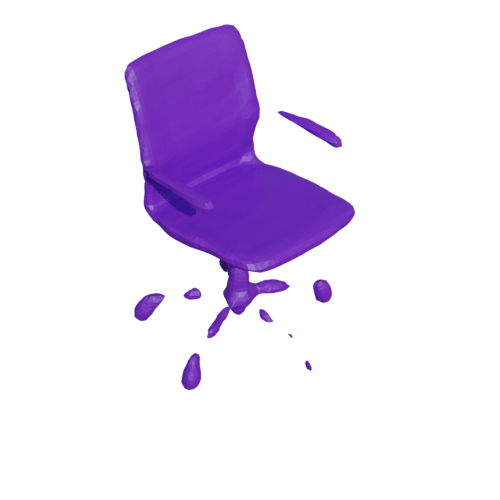} &
\includegraphics[width=.15\linewidth,valign=m]{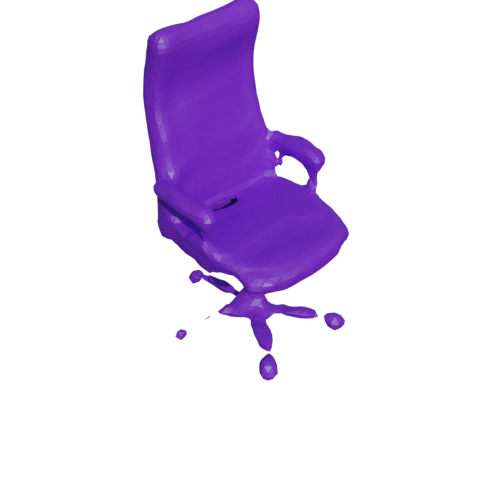} &
\includegraphics[width=.15\linewidth,valign=m]{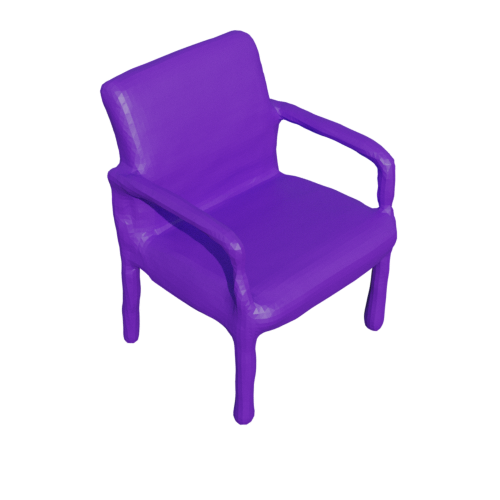} &
\includegraphics[width=.15\linewidth,valign=m]{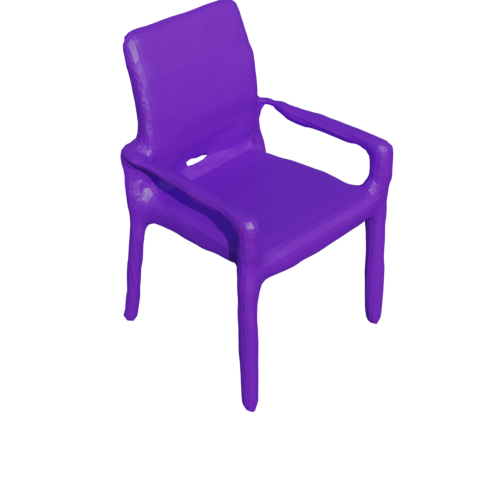} &
\includegraphics[width=.15\linewidth,valign=m]{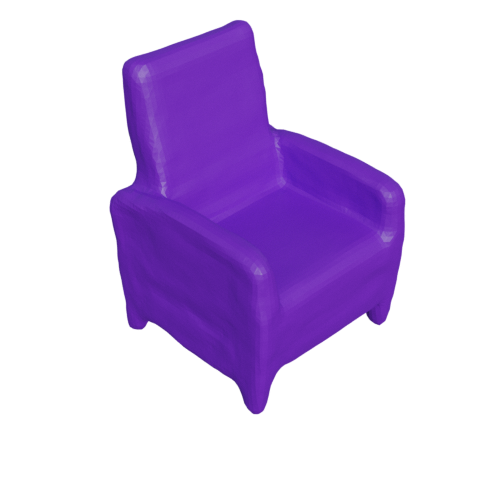} &
 \\
 \rotatebox[origin=c]{90}{Multi-stage}\vspace{-15pt} & \includegraphics[width=.15\linewidth,valign=m]{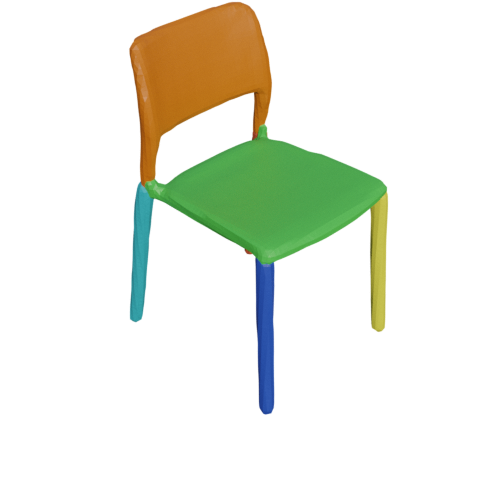} & \includegraphics[width=.15\linewidth,valign=m]{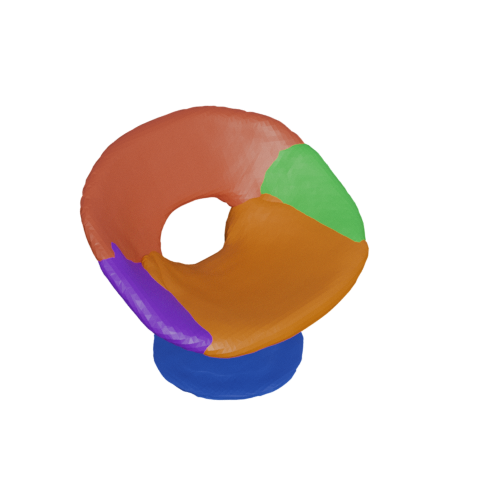} & 
\includegraphics[width=.15\linewidth,valign=m]{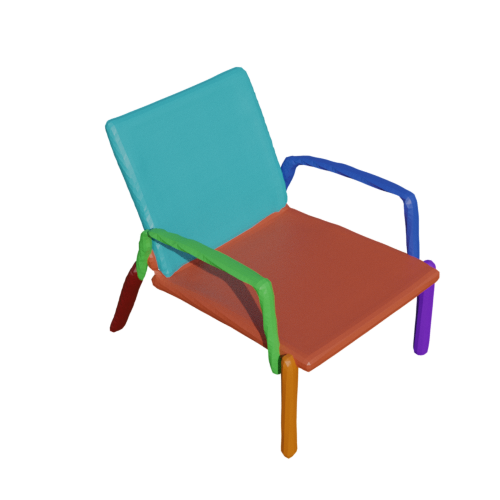} & 
\includegraphics[width=.15\linewidth,valign=m]{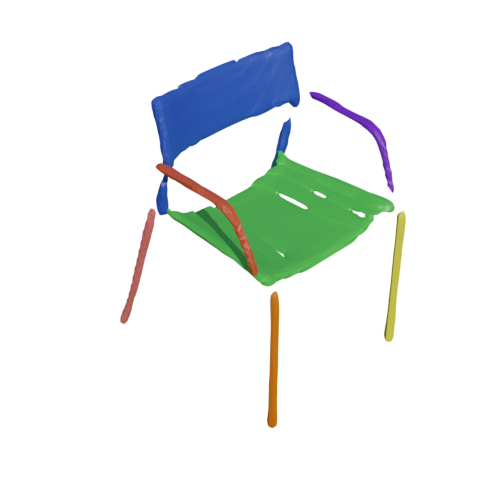} &
\includegraphics[width=.15\linewidth,valign=m]{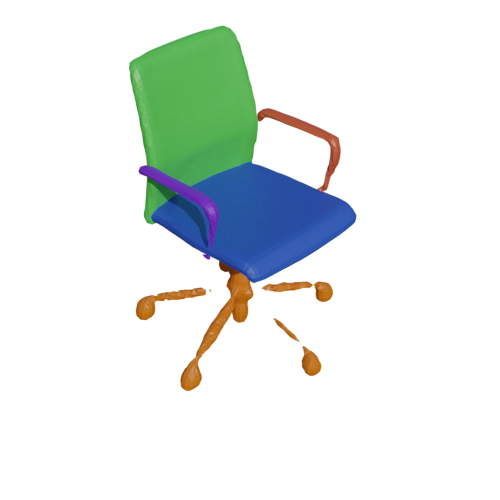} &
\includegraphics[width=.15\linewidth,valign=m]{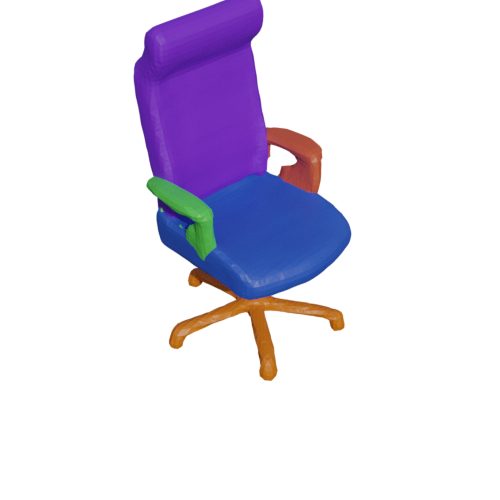} &
\includegraphics[width=.15\linewidth,valign=m]{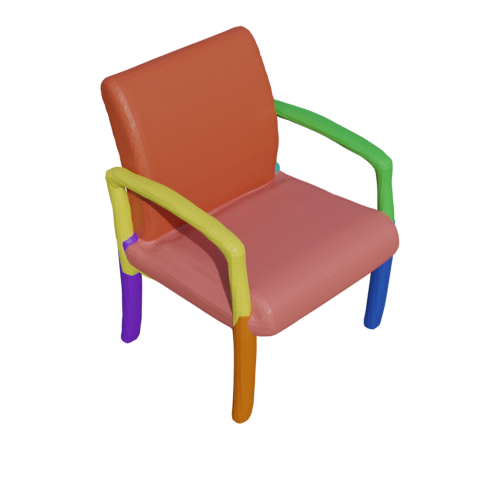} &
\includegraphics[width=.15\linewidth,valign=m]{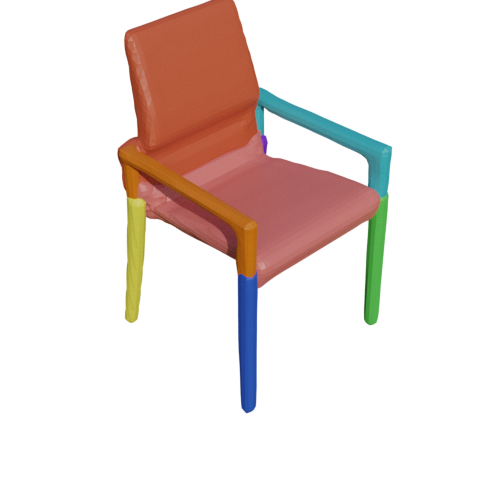} &
\includegraphics[width=.15\linewidth,valign=m]{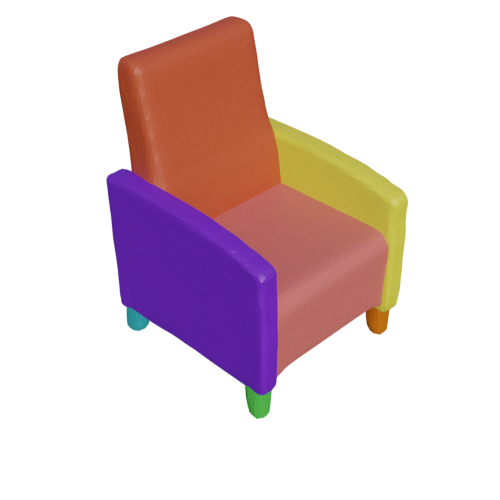} &
 \\
 \rotatebox[origin=c]{90}{GT}\vspace{-5pt} & \includegraphics[width=.15\linewidth,valign=m]{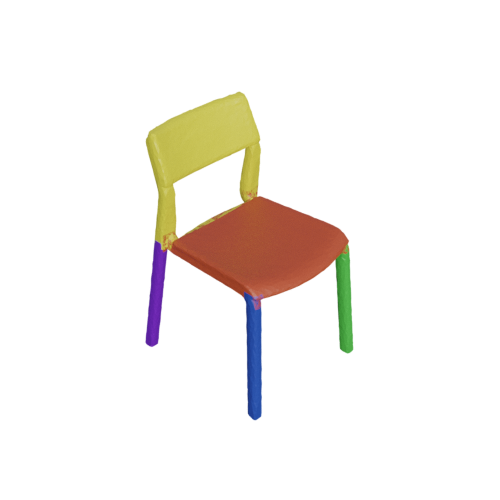} & \includegraphics[width=.15\linewidth,valign=m]{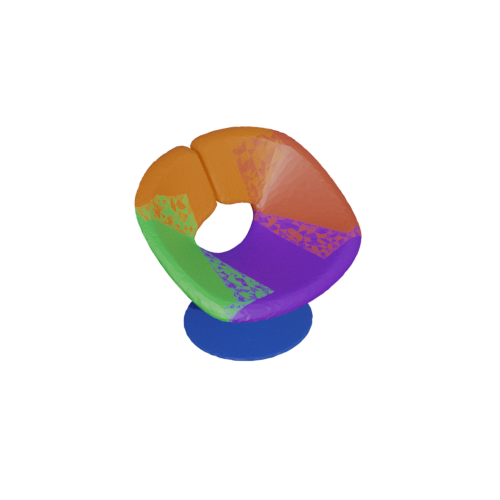} & 
\includegraphics[width=.15\linewidth,valign=m]{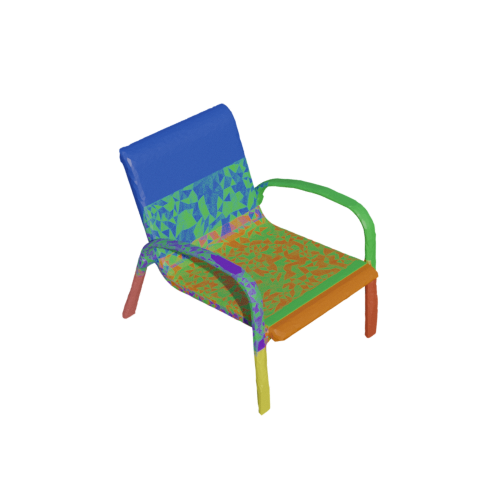} & 
\includegraphics[width=.15\linewidth,valign=m]{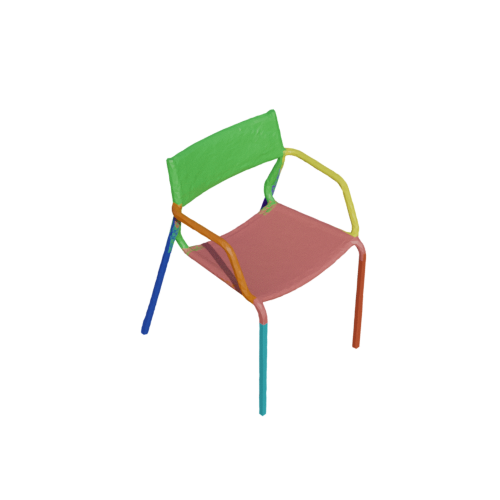} &
\includegraphics[width=.15\linewidth,valign=m]{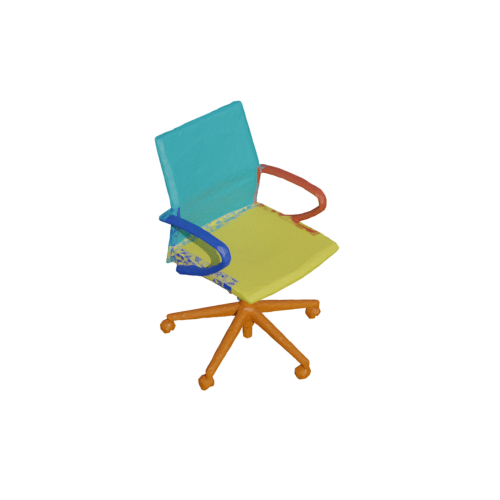} &
\includegraphics[width=.15\linewidth,valign=m]{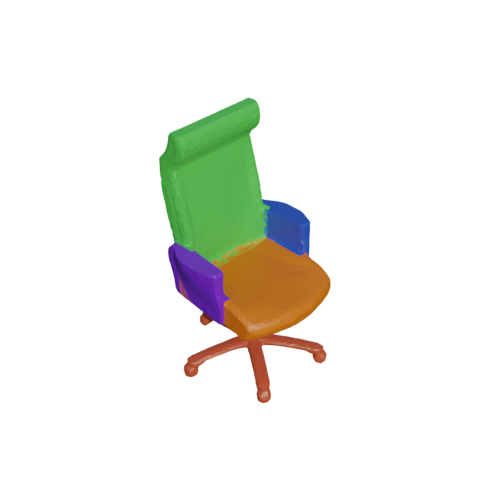} &
\includegraphics[width=.15\linewidth,valign=m]{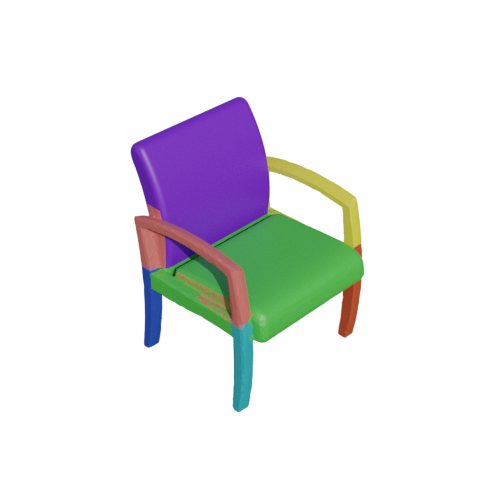} &
\includegraphics[width=.15\linewidth,valign=m]{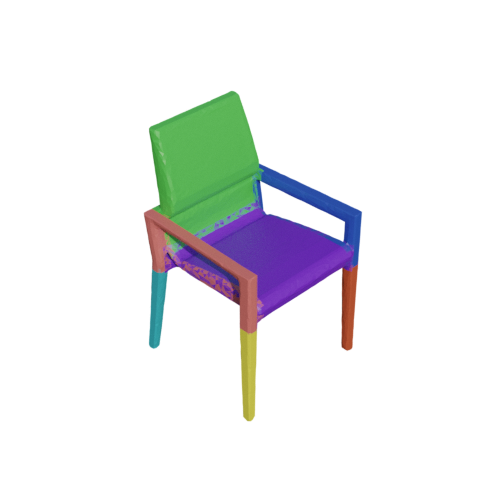} &
\includegraphics[width=.15\linewidth,valign=m]{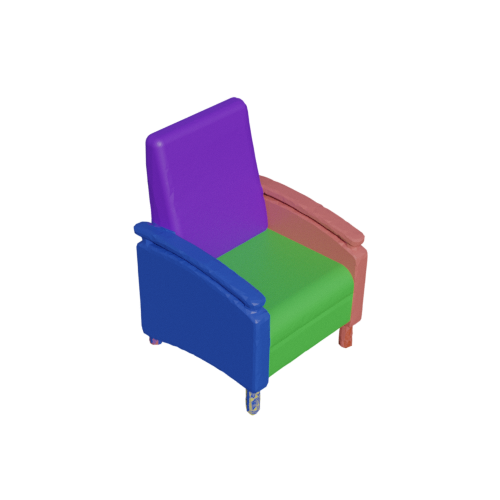} &
 \\
\end{tabular}

\end{center}
\caption{
\textbf{Multi-stage training vs training from scratch.} Training from scratch results in shapes with significant artifacts. In addition, the whole shape is represented by a single output part. Visual artifacts in some ground truth shapes colorings are due to fact that some ground truth parts overlap after preprocessing. When overlapping regions are perfectly aligned it causes rendering issues. We still decided to include such shapes in examples since most of them have interesting geometry. Best viewed in color and zoomed in on-screen.
}
\label{fig:scratch_comparison}
\end{figure*}

%% file: sec/6_conclusions.tex
\section{Conclusion}
%\lorem{3}
%\lorem{3}
We presented a novel shape reconstruction model formulated as an assembly of a set of neural implicits. Our network first predicts part-specific transformations which are then associated with part codes that encode geometry. After that, parts have their geometry decoded and transformed to assemble the final shape. We show that our model outperforms part-agnostic and sequential assembly models. We demonstrate its useful applications: single-view and point cloud reconstruction, selective editing of parts of reconstructed shapes, and reconstruction based on part retrieval.

Some aspects worth investigating are described as follows.
Currently, our model requires pre-training of category-specific part representations which might be unpractical in many scenarios. Exploring how to learn generic part codes, suitable to multiple shape categories, is left as future work. 
Our part transformations are currently limited: we only explore part positioning and uniform scaling. While we achieve state-of-art-results compared to other part-aware approaches, we believe adding other transformations (e.g. free-form deformations) or representing the implicit assembly as a hierarchical set 
are interesting topics for further investigation.

%% file: sec/X_supplementary.tex
% --- PDF will be split by an editor (e.g. macOS preview), so need to restart from page 1

% --- repeat the title (AT: haven't found a more elegant way to do this...)
\twocolumn[
\centering
\Large
%\textbf{Supplementary material} \\

%\vspace{0.5em}Supplementary Material \\
\vspace{1.0em}
] %< twocolumn
\appendix
\setcounter{page}{1}

\paragraph{Watertight mesh generation} 
Our approach requires watertight meshes of the full shapes for training. Since the shapes in the PartNet and ShapeNet datasets are not watertight, we generate their watertight versions using the code provided by Mescheder \etal \cite{occnet} which is itself is a modification of the code by Stutz \etal. \cite{stutz2018learning}. This approach uses 100 depth maps of 640 $\times$ 640 pixels resolution, distributed uniformly on a viewing sphere around the shape, and performs Truncated SDF fusion at a voxel resolution of 256$^3$.

\paragraph{Part extraction} The PartNet dataset provides part meshes on three hierarchical levels of granularity: coarse-, middle- and fine-grained. For our experiments, we use the coarse level. The parts themselves are not watertight (like the full shapes). In addition, there are two additional complications: (a) there are holes in locations of part connections (e.g. holes in the chair seat where legs connect to it) and (b) there are shell-like parts which do not form a closed surface (e.g., flat sheets). To produce a watertight version of each part, we compute the convex hull of the part and intersect it with watertight version of the full shape. Since watertight meshes are slightly different from original ones (e.g. thin parts become slightly thicker), we align the mesh constructed from parts with its watertight version using rigid registration with Coherent Point Drift \cite{myronenko2010point} (specifically, the Probreg library \cite{probreg}). We perform part and shape intersection using Trimesh \cite{trimesh} and a Blender \cite{blender} backend for boolean operations. Our preprocessing procedure fails in approximately $10\%$ of cases resulting in missing parts. We still use shapes with missing parts for training. After preprocessing, average numbers of parts per shape are 5.6/2.8/2.9 for Chair, Table and Lamp categories respectively. Following \cite{occnet} we scale and center all shapes and their parts to a unit cube with 0.1 padding. We save transformation parameters (isotropic scale and translation vector) as a supervision for part positions.

\paragraph{Part connection} Another issue is that some parts share the same semantic label while still being different instances (e.g. chair legs). Other parts are sub-parts of the same instance (e.g. different parts of the seat). We resolve this ambiguity using connectivity constraints: we treat all parts of the shape with same semantic label as the same part if they are connected, 
(i.e., the distance between their closest points is less than $0.05$, assuming shapes are scaled in unit cubes) 
and as different parts otherwise. We do this before performing the intersection process described in the above paragraph.  
%In the end, we scale and center all shapes and their parts to a unit cube with 0.1 padding. We save transformation parameters (isotropic scale and translation vector) to use as supervision for part positions. 

%\paragraph{Preprocessing examples}  Characteristic results from our pre-processing pipeline are shown in Figure \ref{fig:preprocessing}. We can make the following observations. First, our approach is able to close holes in places where parts connect to each other. Second, our approach  deals with shell-like parts transforming them into box-like shapes. Third, due to small alignment issues, our approach sometimes creates thin-shell surfaces in places where parts connect. This does not seem to affect the final assembly model (see qualitative results in Figure \ref{fig:chair_parts}). Fourth, if parts with the same semantic label are connected, they become the same part. Finally, if parts have complex geometry (like office chair handles in Figure \ref{fig:preprocessing}), our approach adds a additional (small) volume to parts. 
%Again, qualitative and quantitative results suggest that our model
%is  influenced by these preprocessing artifacts.   
%We will release the preprocessing pipeline code and its resulting data. 

\paragraph{Shape renderings}
For single view reconstruction experiments we use shape renders provided by authors of JLRD \cite{jldr} (PartNet shapes) and DISN \cite{disn} (ShapeNetCore). Details on rendering process can be found in the respective papers. 

%\section{Qualitative ablation of fine-tuning}

%\input{fig/ablation_quality}

%We provide qualitative evaluation of fine-tuning stage importance in Figure \ref{fig:ablation_quality}. Even though model without SDF fine-tuning produces reasonable looking shapes, their proportions are usually way off which especially hurts reconstruction of smaller parts. It is also interesting that our model is able to generalize over preprocessing artifacts (row 3 in Figure \ref{fig:ablation_quality}).

\paragraph{Part geometry encoder-decoder} We use a point cloud encoder based on PointNet with skip connections described in Table \ref{tab:encoder}. 
%This encoder produces two latent vectors: the first one corresponds to the mean of the part code distribution and the second one corresponds to its log variance. Using these vectors we sample a part code from the associated normal distribution using the reparametrization trick  (as done in variational autoencoders).
The encoded part codes are passed to the implicit function decoder (Table~\ref{tab:decoder}) along with query points to produce implicit function values.

\paragraph{ANISE} The layers of our main architecture are described in Table~\ref{tab:anise}. ANISE takes point clouds as input. It first produces a shape code using a PointNet-based encoder with skip connections (see Table \ref{tab:encoder}). Based on the shape codes, ANISE produces part positions and part codes which are passed to the implicit function decoder (Table \ref{tab:decoder}) along with query points. Before passing query points through the decoder, we transform them for each part according to Equation \ref{eq:combine_implicits_simple} of our main text. 
%The decoder is not used during the first stage of main training (see Section \ref{sec:training}). For fine-tuning, the decoder is initialized using the decoder from pre-trained part geometry encoder-decoder model. 

\input{tab/anise}
\input{tab/encoder}
\input{tab/decoder}
\input{tab/encoder_block}
\input{tab/pointresblock}

\paragraph{Geometrical auto-encoder training} To get part latents supervision, we train Variational Auto-Encoder on normalized parts using SDF supervision. As input, we use part point clouds of size 2048 sampled from watertight part meshes using Poisson sampling. As input encoder, we use PointNet with residual skip connections which produces latent representation of 256 dimensions. Similar to Variational Autoencoders (VAE), encoder produces two latent vectors: mean and log variance of latent distribution. We use them to sample part latent vectors for the decoder. Unlike VAE, we don't regularize our loss with KL-divergence, so part codes end up unnormalized (we experimented with regularization but it yielded worse reconstruction results). When we compute part codes for subsequent training, we use predicted latent vectors of means as part codes. 

\paragraph{Pre-training stage details} For our implicit part surface decoder pre-training (Section 4.1), we use a MLP which outputs SDF values for query points. To obtain SDF supervision, we uniformly sample $100,000$ points inside the bounding box of each part and compute SDF values for them. Due to memory constraints, during training we randomly subsample $10,000$ points from the full point sample set. We train the part geometry encoder-decoder model for $2,000$ epochs on the training split of the data using the Adam optimizer with learning rate of $0.0001$ and batch size $48$.  
This pre-training stage takes a week on a NVIDIA Quadro RTX 8000. This is the most computationally intensive part of our pipeline training. 

\paragraph{Main training stage details} For single view and point cloud reconstruction during training (Section 4.2) we train our model for 200 epochs using validation set loss to save model checkpoints. For all models we use Adam optimizer with learning rates $0.0001$ (point cloud reconstruction) and $0.0002$ (SVR). For single view reconstruction training, for each shape in the batch we use randomly select 12 views out of all views available for the shape. Batch size depends on amount of available GPU memory (around 20 for NVIDIA RTX 2080 Ti). The main training stage takes about 24 hours to converge, depending on the size of category on a NVIDIA RTX 2080 Ti. For experiments on ShapeNetCore we train our models for fixed number of epochs since it doesn't have a predefined validation split (140 epochs for Chair and Table, 120 epochs for Lamp). We strongly recommend using validation set for our training whenever possible, especially for fine-tuning stage (below).

For PQ-Net regression training, we first use code provided by the authors to generate shape codes. Then we use either single view encoder (ResNet-18) or point cloud encoder (PointNet) combined with MLP decoder to regress shape codes. We train these models 1000 epochs using validation loss to checkpoint the best model. 

%We do checkpoint model based on validation loss (we use train/val/test splits), so the final model is the one which performs the best on validation set. 

\paragraph{Fine-tuning stage} During this stage (last paragraph of Section 4.2), we fine-tune for additional $100-200$ epochs using the Adam optimizer with learning rate $0.0001-0.0002$ and batch size depending on amount of available GPU memory. %For implicit function based finetuning we use signed distance field representation (SDF). 
Due to memory constraints, for each shape and for each mini-batch iteration we subsample 500 point queries and associated SDF values used for supervision. For the full shapes we use sample of 100,000 points sampled inside shape bounding box plus 10,000 points sampled inside unit cube to ensure that model interpolates SDF values well inside marching cube reconstruction bounds. 
%We train model with Adam optimizer with batch size of $19$ and learning rate $0.0001$. 
%Model takes about 36 hours to converge on NVIDIA Quadro RTX 8000.

\paragraph{Evaluation details} For ANISE, we reconstruct meshes with marching cubes \cite{lorensen1987marching} using the predicted implicit function values. We restrict the marching cubes reconstruction grid to the unit cube or, in case of point cloud, to a bounding box of the input point cloud with a margin of 0.02. For DISN \cite{disn}, D2IM-Net \cite{li2021d2im} and PQ-Net \cite{pqnet} reconstruction from images or shape codes we use code provided by the authors.
%latents and our SDF decoder (either from finetuning stage or from part geometry encoder-decoder model).

%and use contour level $0$ for all reconstructions. 
%We align reconstructed meshes with reconstructed meshes using Coherent Point Drift \cite{myronenko2010point} since marching cube implementation we use takes only grid SDF values without absolute coordinates which causes resulting shapes to be misaligned with ground truth shapes. 
To compute the IoU and Chamfer distance, we normalize reconstructed meshes to have unit length across the largest dimension. To evaluate Chamfer distance, we randomly sample 2000 points from the reconstructed mesh (same number was used by Wu et al. \cite{pqnet} whose method is included in our comparisons). To evaluate IoU, we voxelize each mesh using a 64$\times$64$\times$64 grid to ensure fair comparison with Wu et al. \cite{pqnet}. To evaluate F-score, we sample $100,000$ points on the ground truth and reconstructed meshes and use $0.02$ threshold to compute precision and recall. 
%For ablation and qualitative results, we report F-score on unnormalized meshes. 

%\paragraph{Comparison with JLRD \cite{jldr}} To perform retrieval baseline, we use non-finetuned model. We retrieve parts from the source set based on our model predictions and L2-distance. We use predicted part positions to assemble shapes from retrieved source parts. Due to preprocessing issues, our source database is a subset of source database used by Uy \etal (around $90\%$ of it).

%$$\section{Training from scratch vs\\multi-stage training}

%$As an additional ablation study, we compare our multi-stage training described in Section \ref{sec:training} to end-to-end ANISE training from ``scratch'' without relying on the intermediate stages.
%which doesn't use intermediate training for part and position codes' predictions. 
%When training from scratch, ANISE ``collapses'' to predicting only one part representing the whole shape while the rest of the parts result in either tiny or no surfaces.
%to model which consistently uses one of $M$ part codes to encode full shape while turning other parts into negligible surfaces.
%Figure \ref{fig:scratch_comparison} demonstrates this effect.
%: model trained from scratch produces only one part representing the whole shape and ANISE produces parts consistently resembling ones used for supervision. 
%Table \ref{tab:end_to_end_comparison} shows that this model  is significantly worse compared to  the multi-stage trained ANISE according to all our evaluation measures.

%\input{fig/supplementary/scratch_comparison}
%\input{tab/supplementary/end_to_end_comparison}

\input{fig/supplementary/distributions}

\Paragraph{Additional results and failure cases} Figures \ref{fig:chair_parts} and \ref{fig:table_parts} show qualitative ANISE reconstruction results for the Chair and Table categories respectively. In both cases ANISE is able to reconstruct plausible parts and assemble them into a shape which  closely resembles the ground truth one. In both figures, colors are used to show distinct parts. Note that the colors are not consistent for  parts of the same category -- ANISE does not output semantically labeled shapes.
%This reflects the fact that parts produces by ANISE are not semantically aware (i.e. first part for Chair category produced by ANISE is not always the same semantic class). 
%Since our model produces duplicate parts, for this visualization we combined parts produced by our model using agglomerative clustering based on Chamfer distance between parts (clustering threshold was the same for all shapes and categories).

Figures~\ref{fig:misc_cases} and \ref{fig:misc_cases_table} show additional qualitative reconstruction examples compared to ground truth meshes for the Chair and Table categories respectively. To reduce the selection bias, results are grouped in three categories: high quality reconstruction (F-score greater than 0.9), medium quality reconstruction (F-score between 0.7 and 0.9) and low quality reconstruction (F-score between 0.45 and 0.7). In most cases, ANISE recovers fine part geometry and produces coherent shapes even though we do not provide explicit connectivity constraints to the model. The part-aware nature of ANISE also leads to an interesting property: in case of out-of-distribution test cases, the model tends to produce parts resembling the most common parts encountered at shapes in their given position and scale. We also provide distributions of evaluation metrics in Figure~\ref{fig:distvis}.

Figure~\ref{fig:failure_cases} shows failure cases (F-score lower than 0.45) for the Chair category. They broadly fall into three failure modes. First, ANISE has trouble generalizing to out-of-distribution shapes and tends to reconstruct them using parts closely resembling training ones. Second, the model struggles with thin parts and highly detailed geometry, which could be due to the sparsity of the input point clouds. Finally, in the case of unusual test parts, model again tends to use 'average' parts and position them to approximate the input point cloud.

\input{fig/supplementary/chair_parts}
\input{fig/supplementary/misc_cases}

\input{fig/supplementary/failure_cases}

\input{fig/supplementary/table_parts}
\input{fig/supplementary/misc_cases_table}

\section{Code and data information}

\paragraph{Code release} We will release our code and data for reproducibility. 

\paragraph{Used code, packages and data licenses} Here we provide links and license information for the data and code we used. 
\begin{itemize}
\item PartNet dataset \cite{partnet}, MIT license, \href{https://partnet.cs.stanford.edu/}{link}; 
\item Occupancy Networks\cite{occnet} code, MIT license, \href{https://github.com/autonomousvision/occupancy_networks}{link};
\item Mesh fusion code \cite{stutz2018learning}, license info provided at \href{https://github.com/davidstutz/mesh-fusion}{project page} (no commercial use);
\item Joint learning of retrieval and deformation codebase \cite{jldr} code, MIT license, \href{https://github.com/mikacuy/joint_learning_retrieval_deformation}{link};
\item DISN \cite{jldr} code (license in link),  \href{https://github.com/laughtervv/DISN}{link};

\item D2IM \cite{li2021d2im} code (license is not provided),  \href{https://github.com/ManyiLi12345/D2IM-Net}{link};

\item PQ-Net \cite{pqnet} code, MIT license,  \href{https://github.com/ChrisWu1997/PQ-NET}{link};

\item Python libraries: PyTorch \cite{NEURIPS2019_9015} (BSD), Open3D \cite{Zhou2018} (MIT), Trimesh \cite{trimesh} (MIT), ProbReg \cite{probreg} (MIT), Scikit-Image \cite{scikit-image} (BSD), NumPy \cite{harris2020array} (\href{https://numpy.org/doc/stable/license.html}{license link}), SciPy \cite{2020SciPy-NMeth} (BSD), Scikit-learn \cite{scikit-learn} (BSD), pymeshlab \cite{pymeshlab} (GPL-3.0), Matplotlib \cite{hunter2007matplotlib} (BSD), Seaborn \cite{Waskom2021} (BSD 3-Clause "New" or "Revised" License);
\item Software: Blender \cite{blender} (GPL 3.0), MeshLab \cite{LocalChapterEvents:ItalChap:ItalianChapConf2008:129-136} (GPL 3.0).
\end{itemize}

%% file: tab/anise.tex
\begin{table}[t!]
  \centering
\caption{
  \label{tab:anise}
  \textbf{ANISE architecture.} The input to ANISE is a batch of point clouds (batch size is $B$ and number of points in the cloud is $N$).
  The outputs are part codes 
  and part positions (outputs of the 6th and 9th layer, denoted as ``out(6)'' and ``out(9)'' respectively). These are used by the implicit function decoder (Table \ref{tab:decoder}). The ``Encoder'' is the point cloud encoder explained in Table \ref{tab:encoder}. The MLP is a multi-layer perceptron with $4$ layers and LeakyRELU activation with negative slope $0.2$. ``Linear'' means linear layer. ``Reshape'' is a reshape layer. ``CAT'' means concatenation along the last axis. ``EXPAND'' repeats the tensor along newly created second dimension.}
\resizebox{\linewidth}{!}{  \def\arraystretch{1}
\begin{tabular}{c|l|l}
Index & Layer                                            & out       \\
\hline
1 & $\text{Input}$                            & B $\otimes$ N $\otimes$ 3\\
\hline
2 & Encoder(out(1))                                               & B $\otimes$ 256    \\
\hline
3 & MLP(out(2), 4, 2048)                                                & B $\otimes$  2048    \\
\hline
4 & Linear(out(3), 2048, $\text{N}_{\text{parts}} \times 128$)                                                & B $\otimes$ $\text{N}_{\text{parts}} \times 128$    \\
\hline
5 & Reshape(out(4))                                                & B  $\otimes$ $\text{N}_{\text{parts}} \otimes 128$    \\
\hline
6 & Linear(out(5), 128, 4)                                                & B $\otimes$ $\text{N}_{\text{parts}} \otimes 4$    \\
\hline
7 & CAT(out(5), EXPAND(out(2),N))                                                 & B $\otimes$ $\text{N}_{\text{parts}}$ $\otimes$ 368    \\
\hline
8 & MLP(out(7), 4, 2048)                                                & B $\otimes$ $\text{N}_{\text{parts}} \otimes 2048$    \\
\hline
9 & Linear(out(8), 2048, 256)                                                 & B $\otimes$ $\text{N}_{\text{parts}}$ $\otimes$ 256    \\

\hline

\hline
\hline

\end{tabular}
\vspace{2mm}}

\end{table}

%% file: tab/encoder.tex
\begin{table}[t!]
  \centering
\caption{
  \label{tab:encoder}
  \textbf{Point cloud encoder.} The input to the encoder is a batch of point clouds. Its output are part codes. The ``EncoderBlock'' is  explained in Table \ref{tab:encoderblock}. Regarding the rest of the layer names: Linear: linear layer, MaxPool: max pooling along second dimension; LeakyRELU: leaky RELU with negative slope $0.2$. }
\resizebox{\linewidth}{!}{  \def\arraystretch{1}
\begin{tabular}{c|l|l}
Index & Layer                                            & out       \\
\hline
1 & $\text{Input}$                            & B $\otimes$ N $\otimes$ 3\\
\hline
2 & Linear(out(1), 3,1024)                                               & B $\otimes$ N $\otimes$ 1024    \\
\hline
3 & EncoderBlock(out(2), 1024, 1024)                                                & B $\otimes$ N $\otimes$ 1024    \\
\hline
4 & EncoderBlock(out(3), 1024, 1024)                                                & B $\otimes$ N $\otimes$ 1024    \\
\hline
5 & EncoderBlock(out(4), 1024, 1024)                                                & B $\otimes$ N $\otimes$ 1024    \\
\hline
6 & EncoderBlock(out(5), 1024, 1024)                                                & B $\otimes$ N $\otimes$ 1024    \\
\hline
7 & PointResBlock(out(6), 1024, 512)                                                & B $\otimes$ N $\otimes$ 512    \\

\hline
8 & MaxPool(out(7))                                             &  B $\otimes$ 512   \\
\hline
9 & Linear(LeakyRELU(out(8)), 512, 256) &                                    B $\otimes$ 256   \\

\hline
\hline

\end{tabular}
\vspace{2mm}}

\end{table}

%% file: tab/decoder.tex
\begin{table}[t!]
  \centering
\caption{
  \label{tab:decoder}
  \textbf{Implicit function decoder.} This decoder takes a batch of shape or part codes (size $B$) and $N$ 3D query points and outputs implicit function values for each code and corresponding query point. The layer/operation names are the following: Linear: linear layer; CAT: concatenation along the last dimension; EXPAND: repeat tensor $N$ times across newly created second dimension; PointResBlock: explained in Table \ref{tab:pointresblock} ; RELU: rectified linear unit \cite{agarap2018deep}. 
  }
\resizebox{\linewidth}{!}{  \def\arraystretch{1}
\begin{tabular}{c|l|l}
Index & Layer                                            & out       \\
\hline
1 & $\text{Input latents}$                            & B $\otimes$ 256\\
\hline
2 & $\text{Query points}$                            & B $\otimes$ N $\otimes$ 3\\
\hline

3 & Linear(out(1), 256, 512)                                               & B $\otimes$ 512    \\
\hline
4 & Linear(out(2), 3, 512)                                                & B $\otimes$ N $\otimes$ 512    \\
\hline
5 & CAT(out(4), EXPAND(out(2),N))                                                 & B $\otimes$ N $\otimes$ 1024    \\
\hline
6 & Linear(out(5), 1024, 512)                                                & B $\otimes$ N $\otimes$ 512    \\
\hline
7 & PointResBlock(out(6), 512, 512)                                                  & B $\otimes$ N $\otimes$ 512    \\
\hline
8 & PointResBlock(out(7), 512, 512)                                                & B $\otimes$ N $\otimes$ 512    \\
\hline
9 & PointResBlock(out(8), 512, 512)                                                & B $\otimes$ N $\otimes$ 512    \\
\hline
10 & PointResBlock(out(9), 512, 512)                                                & B $\otimes$ N $\otimes$ 512    \\
\hline
11 & PointResBlock(out(10), 512, 512)                                                & B $\otimes$ N $\otimes$ 512    \\
\hline
12 & Linear(RELU(out(11)), 512, 1) &                                    B $\otimes$ N $\otimes$ 1   \\

\hline
\hline

\end{tabular}
\vspace{2mm}}

\end{table}

%% file: tab/encoder_block.tex
\begin{table}[t!]
  \centering
\caption{
  \label{tab:encoderblock}
  \textbf{EncoderBlock.}
  PointResBlock: explained in Table \ref{tab:pointresblock}; Linear: linear layer; CAT: concatenate tensors along the last dimension, EXPAND:  repeat tensor N times over newly created second dimension; MaxPool: max pooling over second dimension.}
\resizebox{\linewidth}{!}{  \def\arraystretch{1}
\begin{tabular}{c|l|l}
Index & Layer                                            & out       \\
\hline
1 & $\text{Input}$                            & B $\otimes$ N $\otimes$ 2D\\
\hline
%2 & Linear(out(1), D,1024)                                               & B $\otimes$ N $\otimes$ 1024    \\
%\hline
2 & PointResBlock(out(1), 2D, D)                                                & B $\otimes$ N $\otimes$ D    \\
\hline
3 & MaxPool(out(2))                                             &  B $\otimes$ D   \\
\hline
4 & CAT(out(2), EXPAND(out(3),N))                                   & B $\otimes$ N $\otimes$ 2D   \\
\hline

\hline

\end{tabular}
\vspace{2mm}}

\end{table}

%% file: tab/pointresblock.tex
\begin{table}[t!]
  \centering
\caption{
  \label{tab:pointresblock}
  \textbf{PointResBlock.} This is the building block of the point cloud encoder and implicit function decoder. 
  EdgeConv: edge convolution, FC: fully connected layer, CAT: concatenate tensors along the second dimension, AMP:  adaptive max-pooling along the first dimension, AAP:  adaptive average-pooling along the first dimension.}
\resizebox{\linewidth}{!}{  \def\arraystretch{1}
\begin{tabular}{c|l|l}
Index & Layer                                            & out       \\
\hline
1 & $\text{Input}$                            & B $\otimes$ N $\otimes$ D\\
\hline
%2 & Linear(out(1), D,1024)                                               & B $\otimes$ N $\otimes$ 1024    \\
%\hline
2 & Linear(RELU(out(1)), D, $\text{D}_{\text{hidden}}$)                                                & B $\otimes$ N $\otimes$ $\text{D}_{\text{hidden}}$    \\
\hline
3 &  Linear(RELU(out(2)), $\text{D}_{\text{hidden}}$, $\text{D}_{\text{out}}$)                           & B $\otimes$ N $\otimes$ $\text{D}_{\text{out}}$  \\
\hline
4 & Linear(out(1), D, $\text{D}_{\text{out}})$  + out(3)                                   & B $\otimes$ N $\otimes$ $\text{D}_{\text{out}}$    \\
\hline

\hline

\end{tabular}
\vspace{2mm}}

\end{table}

%% file: fig/supplementary/distributions.tex
\begin{figure*}[!h]
\begin{center}
  \includegraphics[width=.32\linewidth]{fig/supplementary/dists/table_chair_iou.png}
  \includegraphics[width=.32\linewidth]{fig/supplementary/dists/table_chair_chamfer.png}
  \includegraphics[width=.32\linewidth]{fig/supplementary/dists/table_chair_fscore.png}

\end{center}
\caption{\textbf{Distributions of test metrics.} Histograms of IOU (\textbf{left}), Chamfer distance (\textbf{center}) and F-score (\textbf{right}) for the Chair and Table categories. Histograms are normalized so that the values of bins sum to $1$. For IoU and F-score we used $20$ bins with $0.05$ step. For Chamfer distance we used $30$ bins with $0.5$ step.}
\label{fig:distvis}
\end{figure*}

%% file: fig/supplementary/chair_parts.tex
\newcommand{\partfigsize}{.14}
\begin{figure*}[!ht]

\begin{center}
% \begin{overpic} 
% [width=\linewidth]
% {example-image-a}
% \end{overpic}
\setlength\tabcolsep{-10pt}
\begin{tabular}{c@{\hskip 5pt}llllllllll}
\rotatebox[origin=c]{90}{Input}\vspace{-7pt} & \includegraphics[width=\partfigsize\linewidth,valign=m]{fig/supplementary/chair_parts/36243/point_cloud0001.png} 
& \includegraphics[width=\partfigsize\linewidth,valign=m]{fig/supplementary/chair_parts/36309/point_cloud0001.png} & 
\includegraphics[width=\partfigsize\linewidth,valign=m]{fig/supplementary/chair_parts/36489/point_cloud0001.png} & 
\includegraphics[width=\partfigsize\linewidth,valign=m]{fig/supplementary/chair_parts/36535/point_cloud0001.png} &
\includegraphics[width=\partfigsize\linewidth,valign=m]{fig/supplementary/chair_parts/36538/point_cloud0001.png} &
\includegraphics[width=\partfigsize\linewidth,valign=m]{fig/supplementary/chair_parts/36548/point_cloud0001.png} &
\includegraphics[width=\partfigsize\linewidth,valign=m]{fig/supplementary/chair_parts/36558/point_cloud0001.png} &
\includegraphics[width=\partfigsize\linewidth,valign=m]{fig/supplementary/chair_parts/36702/point_cloud0001.png} &
\includegraphics[width=\partfigsize\linewidth,valign=m]{fig/supplementary/chair_parts/36703/point_cloud0001.png} &
 \\
 \rotatebox[origin=c]{90}{Recon}\vspace{-7pt} & \includegraphics[width=\partfigsize\linewidth,valign=m]{fig/supplementary/chair_parts/36243/col_fine0001.png} & \includegraphics[width=\partfigsize\linewidth,valign=m]{fig/supplementary/chair_parts/36309/col_fine0001.png} & 
\includegraphics[width=\partfigsize\linewidth,valign=m]{fig/supplementary/chair_parts/36489/col_fine0001.png} & 
\includegraphics[width=\partfigsize\linewidth,valign=m]{fig/supplementary/chair_parts/36535/col_fine0001.png} &
\includegraphics[width=\partfigsize\linewidth,valign=m]{fig/supplementary/chair_parts/36538/col_fine0001.png} &
\includegraphics[width=\partfigsize\linewidth,valign=m]{fig/supplementary/chair_parts/36548/col_fine0001.png} &
\includegraphics[width=\partfigsize\linewidth,valign=m]{fig/supplementary/chair_parts/36558/col_fine0001.png} &
\includegraphics[width=\partfigsize\linewidth,valign=m]{fig/supplementary/chair_parts/36702/col_fine0001.png} &
\includegraphics[width=\partfigsize\linewidth,valign=m]{fig/supplementary/chair_parts/36703/col_fine0001.png} &
 \\
 \rotatebox[origin=c]{90}{GT}\vspace{0pt} & \includegraphics[width=\partfigsize\linewidth,valign=m]{fig/supplementary/chair_parts/36243/col_gt0001.png} & \includegraphics[width=\partfigsize\linewidth,valign=m]{fig/supplementary/chair_parts/36309/col_gt0001.png} & 
\includegraphics[width=\partfigsize\linewidth,valign=m]{fig/supplementary/chair_parts/36489/col_gt0001.png} & 
\includegraphics[width=\partfigsize\linewidth,valign=m]{fig/supplementary/chair_parts/36535/col_gt0001.png} &
\includegraphics[width=\partfigsize\linewidth,valign=m]{fig/supplementary/chair_parts/36538/col_gt0001.png} &
\includegraphics[width=\partfigsize\linewidth,valign=m]{fig/supplementary/chair_parts/36548/col_gt0001.png} &
\includegraphics[width=\partfigsize\linewidth,valign=m]{fig/supplementary/chair_parts/36558/col_gt0001.png} &
\includegraphics[width=\partfigsize\linewidth,valign=m]{fig/supplementary/chair_parts/36702/col_gt0001.png} &
\includegraphics[width=\partfigsize\linewidth,valign=m]{fig/supplementary/chair_parts/36703/col_gt0001.png} &
 \\
\end{tabular}
\noindent\rule[0.5ex]{\linewidth}{0.5pt}
%+==============================================================
%================================================================
\begin{tabular}{c@{\hskip 5pt}llllllllll}
\rotatebox[origin=c]{90}{Input}\vspace{-7pt} & \includegraphics[width=\partfigsize\linewidth,valign=m]{fig/supplementary/chair_parts/36724/point_cloud0001.png} & \includegraphics[width=\partfigsize\linewidth,valign=m]{fig/supplementary/chair_parts/36845/point_cloud0001.png} & 
\includegraphics[width=\partfigsize\linewidth,valign=m]{fig/supplementary/chair_parts/36861/point_cloud0001.png} & 
\includegraphics[width=\partfigsize\linewidth,valign=m]{fig/supplementary/chair_parts/36898/point_cloud0001.png} &
\includegraphics[width=\partfigsize\linewidth,valign=m]{fig/supplementary/chair_parts/37838/point_cloud0001.png} &
\includegraphics[width=\partfigsize\linewidth,valign=m]{fig/supplementary/chair_parts/37026/point_cloud0001.png} &
\includegraphics[width=\partfigsize\linewidth,valign=m]{fig/supplementary/chair_parts/37114/point_cloud0001.png} &
\includegraphics[width=\partfigsize\linewidth,valign=m]{fig/supplementary/chair_parts/37079/point_cloud0001.png} &
\includegraphics[width=\partfigsize\linewidth,valign=m]{fig/supplementary/chair_parts/37099/point_cloud0001.png} &
 \\
 \rotatebox[origin=c]{90}{Recon}\vspace{-7pt} & \includegraphics[width=\partfigsize\linewidth,valign=m]{fig/supplementary/chair_parts/36724/col_fine0001.png} & \includegraphics[width=\partfigsize\linewidth,valign=m]{fig/supplementary/chair_parts/36845/col_fine0001.png} & 
\includegraphics[width=\partfigsize\linewidth,valign=m]{fig/supplementary/chair_parts/36861/col_fine0001.png} & 
\includegraphics[width=\partfigsize\linewidth,valign=m]{fig/supplementary/chair_parts/36898/col_fine0001.png} &
\includegraphics[width=\partfigsize\linewidth,valign=m]{fig/supplementary/chair_parts/37838/col_fine0001.png} &
\includegraphics[width=\partfigsize\linewidth,valign=m]{fig/supplementary/chair_parts/37026/col_fine0001.png} &
\includegraphics[width=\partfigsize\linewidth,valign=m]{fig/supplementary/chair_parts/37114/col_fine0001.png} &
\includegraphics[width=\partfigsize\linewidth,valign=m]{fig/supplementary/chair_parts/37079/col_fine0001.png} &
\includegraphics[width=\partfigsize\linewidth,valign=m]{fig/supplementary/chair_parts/37099/col_fine0001.png} &
 \\
 \rotatebox[origin=c]{90}{GT}\vspace{0pt} & \includegraphics[width=\partfigsize\linewidth,valign=m]{fig/supplementary/chair_parts/36724/col_gt0001.png} & \includegraphics[width=\partfigsize\linewidth,valign=m]{fig/supplementary/chair_parts/36845/col_gt0001.png} & 
\includegraphics[width=\partfigsize\linewidth,valign=m]{fig/supplementary/chair_parts/36861/col_gt0001.png} & 
\includegraphics[width=\partfigsize\linewidth,valign=m]{fig/supplementary/chair_parts/36898/col_gt0001.png} &
\includegraphics[width=\partfigsize\linewidth,valign=m]{fig/supplementary/chair_parts/37838/col_gt0001.png} &
\includegraphics[width=\partfigsize\linewidth,valign=m]{fig/supplementary/chair_parts/37026/col_gt0001.png} &
\includegraphics[width=\partfigsize\linewidth,valign=m]{fig/supplementary/chair_parts/37114/col_gt0001.png} &
\includegraphics[width=\partfigsize\linewidth,valign=m]{fig/supplementary/chair_parts/37079/col_gt0001.png} &
\includegraphics[width=\partfigsize\linewidth,valign=m]{fig/supplementary/chair_parts/37099/col_gt0001.png} &
 \\
\end{tabular}
\noindent\rule[0.5ex]{\linewidth}{0.5pt}
%+==============================================================
%================================================================
\begin{tabular}{c@{\hskip 5pt}llllllllll}
\rotatebox[origin=c]{90}{Input}\vspace{-7pt} & \includegraphics[width=\partfigsize\linewidth,valign=m]{fig/supplementary/chair_parts/37700/point_cloud0001.png} & \includegraphics[width=\partfigsize\linewidth,valign=m]{fig/supplementary/chair_parts/37116/point_cloud0001.png} & 
\includegraphics[width=\partfigsize\linewidth,valign=m]{fig/supplementary/chair_parts/37585/point_cloud0001.png} & 
\includegraphics[width=\partfigsize\linewidth,valign=m]{fig/supplementary/chair_parts/37541/point_cloud0001.png} &
\includegraphics[width=\partfigsize\linewidth,valign=m]{fig/supplementary/chair_parts/37378/point_cloud0001.png} &
\includegraphics[width=\partfigsize\linewidth,valign=m]{fig/supplementary/chair_parts/37304/point_cloud0001.png} &
\includegraphics[width=\partfigsize\linewidth,valign=m]{fig/supplementary/chair_parts/37277/point_cloud0001.png} &
\includegraphics[width=\partfigsize\linewidth,valign=m]{fig/supplementary/chair_parts/37206/point_cloud0001.png} &
\includegraphics[width=\partfigsize\linewidth,valign=m]{fig/supplementary/chair_parts/37593/point_cloud0001.png} &
 \\
 \rotatebox[origin=c]{90}{Recon}\vspace{-7pt} & \includegraphics[width=\partfigsize\linewidth,valign=m]{fig/supplementary/chair_parts/37700/col_fine0001.png} & \includegraphics[width=\partfigsize\linewidth,valign=m]{fig/supplementary/chair_parts/37116/col_fine0001.png} & 
\includegraphics[width=\partfigsize\linewidth,valign=m]{fig/supplementary/chair_parts/37585/col_fine0001.png} & 
\includegraphics[width=\partfigsize\linewidth,valign=m]{fig/supplementary/chair_parts/37541/col_fine0001.png} &
\includegraphics[width=\partfigsize\linewidth,valign=m]{fig/supplementary/chair_parts/37378/col_fine0001.png} &
\includegraphics[width=\partfigsize\linewidth,valign=m]{fig/supplementary/chair_parts/37304/col_fine0001.png} &
\includegraphics[width=\partfigsize\linewidth,valign=m]{fig/supplementary/chair_parts/37277/col_fine0001.png} &
\includegraphics[width=\partfigsize\linewidth,valign=m]{fig/supplementary/chair_parts/37206/col_fine0001.png} &
\includegraphics[width=\partfigsize\linewidth,valign=m]{fig/supplementary/chair_parts/37593/col_fine0001.png} &
 \\
 \rotatebox[origin=c]{90}{GT}\vspace{-7pt} &
\includegraphics[width=\partfigsize\linewidth,valign=m]{fig/supplementary/chair_parts/37700/col_gt0001.png} & \includegraphics[width=\partfigsize\linewidth,valign=m]{fig/supplementary/chair_parts/37116/col_gt0001.png} & 
\includegraphics[width=\partfigsize\linewidth,valign=m]{fig/supplementary/chair_parts/37585/col_gt0001.png} & 
\includegraphics[width=\partfigsize\linewidth,valign=m]{fig/supplementary/chair_parts/37541/col_gt0001.png} &
\includegraphics[width=\partfigsize\linewidth,valign=m]{fig/supplementary/chair_parts/37378/col_gt0001.png} &
\includegraphics[width=\partfigsize\linewidth,valign=m]{fig/supplementary/chair_parts/37304/col_gt0001.png} &
\includegraphics[width=\partfigsize\linewidth,valign=m]{fig/supplementary/chair_parts/37277/col_gt0001.png} &
\includegraphics[width=\partfigsize\linewidth,valign=m]{fig/supplementary/chair_parts/37206/col_gt0001.png} &
\includegraphics[width=\partfigsize\linewidth,valign=m]{fig/supplementary/chair_parts/37593/col_gt0001.png} &
 \\
\end{tabular}

\end{center}
\caption{
\textbf{Qualitative examples of part-aware reconstruction for Chair category.} All presented shapes have reconstruction F-score greater than 0.75. Visual artifacts in some ground truth shapes colorings are due to fact that some ground truth parts overlap after preprocessing. When overlapping regions are perfectly aligned it causes rendering issues. We still decided to include such shapes in examples since most of them have interesting geometry. Best viewed in color and zoomed in on-screen.
}
\label{fig:chair_parts}
\end{figure*}

%% file: fig/supplementary/misc_cases.tex
\begin{figure*}[!ht]
\begin{center}
% \begin{overpic} 
% [width=\linewidth]
% {example-image-a}
% \end{overpic}
\setlength\tabcolsep{-10pt}
\begin{tabular}{c@{\hskip 5pt}llllllllll}
\rotatebox[origin=c]{90}{Input}\vspace{-15pt} & \includegraphics[width=.15\linewidth,valign=m]{fig/supplementary/good_cases/2470/point_cloud0001.png} & \includegraphics[width=.15\linewidth,valign=m]{fig/supplementary/good_cases/2488/point_cloud0001.png} & \includegraphics[width=.15\linewidth,valign=m]{fig/supplementary/good_cases/2506/point_cloud0001.png} & 
\includegraphics[width=.15\linewidth,valign=m]{fig/supplementary/good_cases/2609/point_cloud0001.png} &
\includegraphics[width=.15\linewidth,valign=m]{fig/supplementary/good_cases/2669/point_cloud0001.png} &
\includegraphics[width=.15\linewidth,valign=m]{fig/supplementary/good_cases/2858/point_cloud0001.png} &
\includegraphics[width=.15\linewidth,valign=m]{fig/supplementary/good_cases/3037/point_cloud0001.png} &
\includegraphics[width=.15\linewidth,valign=m]{fig/supplementary/good_cases/3078/point_cloud0001.png} &
\includegraphics[width=.15\linewidth,valign=m]{fig/supplementary/good_cases/3143/point_cloud0001.png} &
 \\
\rotatebox[origin=c]{90}{Recon} \vspace{-15pt}& \includegraphics[width=.15\linewidth,valign=m]{fig/supplementary/good_cases/2470/recon0001.png} & \includegraphics[width=.15\linewidth,valign=m]{fig/supplementary/good_cases/2488/recon0001.png} & \includegraphics[width=.15\linewidth,valign=m]{fig/supplementary/good_cases/2506/recon0001.png} & 
\includegraphics[width=.15\linewidth,valign=m]{fig/supplementary/good_cases/2609/recon0001.png} &
\includegraphics[width=.15\linewidth,valign=m]{fig/supplementary/good_cases/2669/recon0001.png} &
\includegraphics[width=.15\linewidth,valign=m]{fig/supplementary/good_cases/2858/recon0001.png} &
\includegraphics[width=.15\linewidth,valign=m]{fig/supplementary/good_cases/3037/recon0001.png} &
\includegraphics[width=.15\linewidth,valign=m]{fig/supplementary/good_cases/3078/recon0001.png} &
\includegraphics[width=.15\linewidth,valign=m]{fig/supplementary/good_cases/3143/recon0001.png} &
 \\
\rotatebox[origin=c]{90}{GT}\vspace{-8pt} & \includegraphics[width=.15\linewidth,valign=m]{fig/supplementary/good_cases/2470/orig0001.png} & \includegraphics[width=.15\linewidth,valign=m]{fig/supplementary/good_cases/2488/orig0001.png} & \includegraphics[width=.15\linewidth,valign=m]{fig/supplementary/good_cases/2506/orig0001.png} & 
\includegraphics[width=.15\linewidth,valign=m]{fig/supplementary/good_cases/2609/orig0001.png} &
\includegraphics[width=.15\linewidth,valign=m]{fig/supplementary/good_cases/2669/orig0001.png} &
\includegraphics[width=.15\linewidth,valign=m]{fig/supplementary/good_cases/2858/orig0001.png} &
\includegraphics[width=.15\linewidth,valign=m]{fig/supplementary/good_cases/3037/orig0001.png} &
\includegraphics[width=.15\linewidth,valign=m]{fig/supplementary/good_cases/3078/orig0001.png} &
\includegraphics[width=.15\linewidth,valign=m]{fig/supplementary/good_cases/3143/orig0001.png} &
 \\
\end{tabular}
\noindent\rule[0.5ex]{\linewidth}{0.5pt}
%+==============================================================
%================================================================
\begin{tabular}{c@{\hskip 5pt}llllllllll}
\rotatebox[origin=c]{90}{Input}\vspace{-15pt} & \includegraphics[width=.15\linewidth,valign=m]{fig/supplementary/average_cases/2231/point_cloud0001.png} & \includegraphics[width=.15\linewidth,valign=m]{fig/supplementary/average_cases/2298/point_cloud0001.png} & \includegraphics[width=.15\linewidth,valign=m]{fig/supplementary/average_cases/2368/point_cloud0001.png} & 
\includegraphics[width=.15\linewidth,valign=m]{fig/supplementary/average_cases/2402/point_cloud0001.png} &
\includegraphics[width=.15\linewidth,valign=m]{fig/supplementary/average_cases/2410/point_cloud0001.png} &
\includegraphics[width=.15\linewidth,valign=m]{fig/supplementary/average_cases/2465/point_cloud0001.png} &
\includegraphics[width=.15\linewidth,valign=m]{fig/supplementary/average_cases/2965/point_cloud0001.png} &
\includegraphics[width=.15\linewidth,valign=m]{fig/supplementary/average_cases/2648/point_cloud0001.png} &
\includegraphics[width=.15\linewidth,valign=m]{fig/supplementary/average_cases/2861/point_cloud0001.png} &
 \\
\rotatebox[origin=c]{90}{Recon} \vspace{-15pt} & \includegraphics[width=.15\linewidth,valign=m]{fig/supplementary/average_cases/2231/recon0001.png} & \includegraphics[width=.15\linewidth,valign=m]{fig/supplementary/average_cases/2298/recon0001.png} & \includegraphics[width=.15\linewidth,valign=m]{fig/supplementary/average_cases/2368/recon0001.png} & 
\includegraphics[width=.15\linewidth,valign=m]{fig/supplementary/average_cases/2402/recon0001.png} &
\includegraphics[width=.15\linewidth,valign=m]{fig/supplementary/average_cases/2410/recon0001.png} &
\includegraphics[width=.15\linewidth,valign=m]{fig/supplementary/average_cases/2465/recon0001.png} &
\includegraphics[width=.15\linewidth,valign=m]{fig/supplementary/average_cases/2965/recon0001.png} &
\includegraphics[width=.15\linewidth,valign=m]{fig/supplementary/average_cases/2648/recon0001.png} &
\includegraphics[width=.15\linewidth,valign=m]{fig/supplementary/average_cases/2861/recon0001.png} &
 \\
\rotatebox[origin=c]{90}{GT}\vspace{-8pt} & \includegraphics[width=.15\linewidth,valign=m]{fig/supplementary/average_cases/2231/orig0001.png} & \includegraphics[width=.15\linewidth,valign=m]{fig/supplementary/average_cases/2298/orig0001.png} & \includegraphics[width=.15\linewidth,valign=m]{fig/supplementary/average_cases/2368/orig0001.png} & 
\includegraphics[width=.15\linewidth,valign=m]{fig/supplementary/average_cases/2402/orig0001.png} &
\includegraphics[width=.15\linewidth,valign=m]{fig/supplementary/average_cases/2410/orig0001.png} &
\includegraphics[width=.15\linewidth,valign=m]{fig/supplementary/average_cases/2465/orig0001.png} &
\includegraphics[width=.15\linewidth,valign=m]{fig/supplementary/average_cases/2965/orig0001.png} &
\includegraphics[width=.15\linewidth,valign=m]{fig/supplementary/average_cases/2648/orig0001.png} &
\includegraphics[width=.15\linewidth,valign=m]{fig/supplementary/average_cases/2861/orig0001.png} &
 \\
\end{tabular}

\noindent\rule[0.5ex]{\linewidth}{0.5pt}
%+==============================================================
%================================================================
\begin{tabular}{c@{\hskip 5pt}llllllllll}
\rotatebox[origin=c]{90}{Input}\vspace{-15pt} & \includegraphics[width=.15\linewidth,valign=m]{fig/supplementary/poor_cases/2197/point_cloud0001.png} & \includegraphics[width=.15\linewidth,valign=m]{fig/supplementary/poor_cases/2523/point_cloud0001.png} & 
\includegraphics[width=.15\linewidth,valign=m]{fig/supplementary/poor_cases/2615/point_cloud0001.png} & 
\includegraphics[width=.15\linewidth,valign=m]{fig/supplementary/poor_cases/2622/point_cloud0001.png} &
\includegraphics[width=.15\linewidth,valign=m]{fig/supplementary/poor_cases/2636/point_cloud0001.png} &
\includegraphics[width=.15\linewidth,valign=m]{fig/supplementary/poor_cases/2738/point_cloud0001.png} &
\includegraphics[width=.15\linewidth,valign=m]{fig/supplementary/poor_cases/2743/point_cloud0001.png} &
\includegraphics[width=.15\linewidth,valign=m]{fig/supplementary/poor_cases/2886/point_cloud0001.png} &
\includegraphics[width=.15\linewidth,valign=m]{fig/supplementary/poor_cases/2990/point_cloud0001.png} &
 \\
\rotatebox[origin=c]{90}{Recon} \vspace{-15pt} & \includegraphics[width=.15\linewidth,valign=m]{fig/supplementary/poor_cases/2197/recon0001.png} & \includegraphics[width=.15\linewidth,valign=m]{fig/supplementary/poor_cases/2523/recon0001.png} & 
\includegraphics[width=.15\linewidth,valign=m]{fig/supplementary/poor_cases/2615/recon0001.png} & 
\includegraphics[width=.15\linewidth,valign=m]{fig/supplementary/poor_cases/2622/recon0001.png} &
\includegraphics[width=.15\linewidth,valign=m]{fig/supplementary/poor_cases/2636/recon0001.png} &
\includegraphics[width=.15\linewidth,valign=m]{fig/supplementary/poor_cases/2738/recon0001.png} &
\includegraphics[width=.15\linewidth,valign=m]{fig/supplementary/poor_cases/2743/recon0001.png} &
\includegraphics[width=.15\linewidth,valign=m]{fig/supplementary/poor_cases/2886/recon0001.png} &
\includegraphics[width=.15\linewidth,valign=m]{fig/supplementary/poor_cases/2990/recon0001.png} &
 \\
\rotatebox[origin=c]{90}{GT}\vspace{-15pt} & \includegraphics[width=.15\linewidth,valign=m]{fig/supplementary/poor_cases/2197/orig0001.png} & \includegraphics[width=.15\linewidth,valign=m]{fig/supplementary/poor_cases/2523/orig0001.png} & 
\includegraphics[width=.15\linewidth,valign=m]{fig/supplementary/poor_cases/2615/orig0001.png} & 
\includegraphics[width=.15\linewidth,valign=m]{fig/supplementary/poor_cases/2622/orig0001.png} &
\includegraphics[width=.15\linewidth,valign=m]{fig/supplementary/poor_cases/2636/orig0001.png} &
\includegraphics[width=.15\linewidth,valign=m]{fig/supplementary/poor_cases/2738/orig0001.png} &
\includegraphics[width=.15\linewidth,valign=m]{fig/supplementary/poor_cases/2743/orig0001.png} &
\includegraphics[width=.15\linewidth,valign=m]{fig/supplementary/poor_cases/2886/orig0001.png} &
\includegraphics[width=.15\linewidth,valign=m]{fig/supplementary/poor_cases/2990/orig0001.png} &
 \\
\end{tabular}

\end{center}
\caption{
\textbf{Various reconstruction cases for Chair category.} \emph{Top group:} results with F-score greater than $0.9$. \emph{Middle group:} results with F-score between $0.7$ and $0.9$. \emph{Bottom group:} results with F-score between $0.45$ and $0.7$. Best viewed zoomed in on-screen.
}
\label{fig:misc_cases}
\end{figure*}

%% file: fig/supplementary/failure_cases.tex
\begin{figure*}[!ht]
\begin{center}
% \begin{overpic} 
% [width=\linewidth]
% {example-image-a}
% \end{overpic}
\setlength\tabcolsep{-10pt}
\begin{tabular}{c@{\hskip 5pt}llllllllll}
\rotatebox[origin=c]{90}{Input}\vspace{-15pt} & \includegraphics[width=.15\linewidth,valign=m]{fig/supplementary/failure_cases/35867/point_cloud0001.png} & \includegraphics[width=.15\linewidth,valign=m]{fig/supplementary/failure_cases/2314/point_cloud0001.png} & \includegraphics[width=.15\linewidth,valign=m]{fig/supplementary/failure_cases/36625/point_cloud0001.png} & 
\includegraphics[width=.15\linewidth,valign=m]{fig/supplementary/failure_cases/2480/point_cloud0001.png} &
\includegraphics[width=.15\linewidth,valign=m]{fig/supplementary/failure_cases/2685/point_cloud0001.png} &
\includegraphics[width=.15\linewidth,valign=m]{fig/supplementary/failure_cases/2753/point_cloud0001.png} &
\includegraphics[width=.15\linewidth,valign=m]{fig/supplementary/failure_cases/2798/point_cloud0001.png} &
\includegraphics[width=.15\linewidth,valign=m]{fig/supplementary/failure_cases/3175/point_cloud0001.png} &
\includegraphics[width=.15\linewidth,valign=m]{fig/supplementary/failure_cases/35764/point_cloud0001.png} &
 \\
\rotatebox[origin=c]{90}{Recon} \vspace{-15pt}& \includegraphics[width=.15\linewidth,valign=m]{fig/supplementary/failure_cases/35867/recon0001.png} & \includegraphics[width=.15\linewidth,valign=m]{fig/supplementary/failure_cases/2314/recon0001.png} & \includegraphics[width=.15\linewidth,valign=m]{fig/supplementary/failure_cases/36625/recon0001.png} & 
\includegraphics[width=.15\linewidth,valign=m]{fig/supplementary/failure_cases/2480/recon0001.png} &
\includegraphics[width=.15\linewidth,valign=m]{fig/supplementary/failure_cases/2685/recon0001.png} &
\includegraphics[width=.15\linewidth,valign=m]{fig/supplementary/failure_cases/2753/recon0001.png} &
\includegraphics[width=.15\linewidth,valign=m]{fig/supplementary/failure_cases/2798/recon0001.png} &
\includegraphics[width=.15\linewidth,valign=m]{fig/supplementary/failure_cases/3175/recon0001.png} &
\includegraphics[width=.15\linewidth,valign=m]{fig/supplementary/failure_cases/35764/recon0001.png} &
 \\
\rotatebox[origin=c]{90}{GT}\vspace{-8pt} & \includegraphics[width=.15\linewidth,valign=m]{fig/supplementary/failure_cases/35867/orig0001.png} & \includegraphics[width=.15\linewidth,valign=m]{fig/supplementary/failure_cases/2314/orig0001.png} & \includegraphics[width=.15\linewidth,valign=m]{fig/supplementary/failure_cases/36625/orig0001.png} & 
\includegraphics[width=.15\linewidth,valign=m]{fig/supplementary/failure_cases/2480/orig0001.png} &
\includegraphics[width=.15\linewidth,valign=m]{fig/supplementary/failure_cases/2685/orig0001.png} &
\includegraphics[width=.15\linewidth,valign=m]{fig/supplementary/failure_cases/2753/orig0001.png} &
\includegraphics[width=.15\linewidth,valign=m]{fig/supplementary/failure_cases/2798/orig0001.png} &
\includegraphics[width=.15\linewidth,valign=m]{fig/supplementary/failure_cases/3175/orig0001.png} &
\includegraphics[width=.15\linewidth,valign=m]{fig/supplementary/failure_cases/35764/orig0001.png} &
\\
\end{tabular}
\noindent\rule[0.5ex]{\linewidth}{0.5pt}
%+==============================================================
%================================================================
\begin{tabular}{c@{\hskip 5pt}llllllllll}
\rotatebox[origin=c]{90}{Input}\vspace{-15pt} & \includegraphics[width=.15\linewidth,valign=m]{fig/supplementary/failure_cases/36748/point_cloud0001.png} & \includegraphics[width=.15\linewidth,valign=m]{fig/supplementary/failure_cases/37095/point_cloud0001.png} & \includegraphics[width=.15\linewidth,valign=m]{fig/supplementary/failure_cases/36885/point_cloud0001.png} & 
\includegraphics[width=.15\linewidth,valign=m]{fig/supplementary/failure_cases/36908/point_cloud0001.png} &
\includegraphics[width=.15\linewidth,valign=m]{fig/supplementary/failure_cases/37128/point_cloud0001.png} &
\includegraphics[width=.15\linewidth,valign=m]{fig/supplementary/failure_cases/37173/point_cloud0001.png} &
\includegraphics[width=.15\linewidth,valign=m]{fig/supplementary/failure_cases/37276/point_cloud0001.png} &
\includegraphics[width=.15\linewidth,valign=m]{fig/supplementary/failure_cases/37485/point_cloud0001.png} &
\includegraphics[width=.15\linewidth,valign=m]{fig/supplementary/failure_cases/37643/point_cloud0001.png} &
 \\
\rotatebox[origin=c]{90}{Recon} \vspace{-15pt}& \includegraphics[width=.15\linewidth,valign=m]{fig/supplementary/failure_cases/36748/recon0001.png} & \includegraphics[width=.15\linewidth,valign=m]{fig/supplementary/failure_cases/37095/recon0001.png} & \includegraphics[width=.15\linewidth,valign=m]{fig/supplementary/failure_cases/36885/recon0001.png} & 
\includegraphics[width=.15\linewidth,valign=m]{fig/supplementary/failure_cases/36908/recon0001.png} &
\includegraphics[width=.15\linewidth,valign=m]{fig/supplementary/failure_cases/37128/recon0001.png} &
\includegraphics[width=.15\linewidth,valign=m]{fig/supplementary/failure_cases/37173/recon0001.png} &
\includegraphics[width=.15\linewidth,valign=m]{fig/supplementary/failure_cases/37276/recon0001.png} &
\includegraphics[width=.15\linewidth,valign=m]{fig/supplementary/failure_cases/37485/recon0001.png} &
\includegraphics[width=.15\linewidth,valign=m]{fig/supplementary/failure_cases/37643/recon0001.png} &
 \\
\rotatebox[origin=c]{90}{GT}\vspace{-8pt} & \includegraphics[width=.15\linewidth,valign=m]{fig/supplementary/failure_cases/36748/orig0001.png} & \includegraphics[width=.15\linewidth,valign=m]{fig/supplementary/failure_cases/37095/orig0001.png} & \includegraphics[width=.15\linewidth,valign=m]{fig/supplementary/failure_cases/36885/orig0001.png} & 
\includegraphics[width=.15\linewidth,valign=m]{fig/supplementary/failure_cases/36908/orig0001.png} &
\includegraphics[width=.15\linewidth,valign=m]{fig/supplementary/failure_cases/37128/orig0001.png} &
\includegraphics[width=.15\linewidth,valign=m]{fig/supplementary/failure_cases/37173/orig0001.png} &
\includegraphics[width=.15\linewidth,valign=m]{fig/supplementary/failure_cases/37276/orig0001.png} &
\includegraphics[width=.15\linewidth,valign=m]{fig/supplementary/failure_cases/37485/orig0001.png} &
\includegraphics[width=.15\linewidth,valign=m]{fig/supplementary/failure_cases/37643/orig0001.png} &
\\
\end{tabular}

\noindent\rule[0.5ex]{\linewidth}{0.5pt}
%+==============================================================
%================================================================
\begin{tabular}{c@{\hskip 5pt}llllllllll}
\rotatebox[origin=c]{90}{Input}\vspace{-15pt} & \includegraphics[width=.15\linewidth,valign=m]{fig/supplementary/failure_cases/37688/point_cloud0001.png} & \includegraphics[width=.15\linewidth,valign=m]{fig/supplementary/failure_cases/37790/point_cloud0001.png} & \includegraphics[width=.15\linewidth,valign=m]{fig/supplementary/failure_cases/38124/point_cloud0001.png} & 
\includegraphics[width=.15\linewidth,valign=m]{fig/supplementary/failure_cases/38219/point_cloud0001.png} &
\includegraphics[width=.15\linewidth,valign=m]{fig/supplementary/failure_cases/38308/point_cloud0001.png} &
\includegraphics[width=.15\linewidth,valign=m]{fig/supplementary/failure_cases/38353/point_cloud0001.png} &
\includegraphics[width=.15\linewidth,valign=m]{fig/supplementary/failure_cases/38700/point_cloud0001.png} &
\includegraphics[width=.15\linewidth,valign=m]{fig/supplementary/failure_cases/38732/point_cloud0001.png} &
\includegraphics[width=.15\linewidth,valign=m]{fig/supplementary/failure_cases/39304/point_cloud0001.png} &
 \\
\rotatebox[origin=c]{90}{Recon} \vspace{-15pt}& \includegraphics[width=.15\linewidth,valign=m]{fig/supplementary/failure_cases/37688/recon0001.png} & \includegraphics[width=.15\linewidth,valign=m]{fig/supplementary/failure_cases/37790/recon0001.png} & \includegraphics[width=.15\linewidth,valign=m]{fig/supplementary/failure_cases/38124/recon0001.png} & 
\includegraphics[width=.15\linewidth,valign=m]{fig/supplementary/failure_cases/38219/recon0001.png} &
\includegraphics[width=.15\linewidth,valign=m]{fig/supplementary/failure_cases/38308/recon0001.png} &
\includegraphics[width=.15\linewidth,valign=m]{fig/supplementary/failure_cases/38353/recon0001.png} &
\includegraphics[width=.15\linewidth,valign=m]{fig/supplementary/failure_cases/38700/recon0001.png} &
\includegraphics[width=.15\linewidth,valign=m]{fig/supplementary/failure_cases/38732/recon0001.png} &
\includegraphics[width=.15\linewidth,valign=m]{fig/supplementary/failure_cases/39304/recon0001.png} &
 \\
\rotatebox[origin=c]{90}{GT}\vspace{-15pt} & \includegraphics[width=.15\linewidth,valign=m]{fig/supplementary/failure_cases/37688/orig0001.png} & \includegraphics[width=.15\linewidth,valign=m]{fig/supplementary/failure_cases/37790/orig0001.png} & \includegraphics[width=.15\linewidth,valign=m]{fig/supplementary/failure_cases/38124/orig0001.png} & 
\includegraphics[width=.15\linewidth,valign=m]{fig/supplementary/failure_cases/38219/orig0001.png} &
\includegraphics[width=.15\linewidth,valign=m]{fig/supplementary/failure_cases/38308/orig0001.png} &
\includegraphics[width=.15\linewidth,valign=m]{fig/supplementary/failure_cases/38353/orig0001.png} &
\includegraphics[width=.15\linewidth,valign=m]{fig/supplementary/failure_cases/38700/orig0001.png} &
\includegraphics[width=.15\linewidth,valign=m]{fig/supplementary/failure_cases/38732/orig0001.png} &
\includegraphics[width=.15\linewidth,valign=m]{fig/supplementary/failure_cases/39304/orig0001.png} &
 \\
\end{tabular}

\end{center}
\caption{
\textbf{Failure cases.} Results with F-score lower than $0.45$. All failure cases broadly fall into the following modes: reconstruction of out-of-distribution shapes using common parts; issues with thin parts and fine geometry due to sparsity of point clouds; a tendency of our model to use 'average' parts for the reconstruction. Best viewed zoomed in on-screen.
}
\label{fig:failure_cases}
\end{figure*}

%% file: fig/supplementary/table_parts.tex
\newcommand{\tablepartfigsize}{.135}
\begin{figure*}[!ht]

\begin{center}
% \begin{overpic} 
% [width=\linewidth]
% {example-image-a}
% \end{overpic}
\setlength\tabcolsep{-6pt}
\begin{tabular}{c@{\hskip 5pt}llllllllll}
\rotatebox[origin=c]{90}{Input}\vspace{-7pt} & \includegraphics[width=\tablepartfigsize\linewidth,valign=m]{fig/supplementary/table_parts/18969/point_cloud0001.png} & \includegraphics[width=\tablepartfigsize\linewidth,valign=m]{fig/supplementary/table_parts/18991/point_cloud0001.png} & 
\includegraphics[width=\tablepartfigsize\linewidth,valign=m]{fig/supplementary/table_parts/19139/point_cloud0001.png} & 
\includegraphics[width=\tablepartfigsize\linewidth,valign=m]{fig/supplementary/table_parts/19162/point_cloud0001.png} &
\includegraphics[width=\tablepartfigsize\linewidth,valign=m]{fig/supplementary/table_parts/19347/point_cloud0001.png} &
\includegraphics[width=\tablepartfigsize\linewidth,valign=m]{fig/supplementary/table_parts/19413/point_cloud0001.png} &
\includegraphics[width=\tablepartfigsize\linewidth,valign=m]{fig/supplementary/table_parts/19642/point_cloud0001.png} &
\includegraphics[width=\tablepartfigsize\linewidth,valign=m]{fig/supplementary/table_parts/19644/point_cloud0001.png} &
\includegraphics[width=\tablepartfigsize\linewidth,valign=m]{fig/supplementary/table_parts/19387/point_cloud0001.png} &
 \\
 \rotatebox[origin=c]{90}{Recon}\vspace{-7pt} & \includegraphics[width=\tablepartfigsize\linewidth,valign=m]{fig/supplementary/table_parts/18969/col_fine0001.png} & \includegraphics[width=\tablepartfigsize\linewidth,valign=m]{fig/supplementary/table_parts/18991/col_fine0001.png} & 
\includegraphics[width=\tablepartfigsize\linewidth,valign=m]{fig/supplementary/table_parts/19139/col_fine0001.png} & 
\includegraphics[width=\tablepartfigsize\linewidth,valign=m]{fig/supplementary/table_parts/19162/col_fine0001.png} &
\includegraphics[width=\tablepartfigsize\linewidth,valign=m]{fig/supplementary/table_parts/19347/col_fine0001.png} &
\includegraphics[width=\tablepartfigsize\linewidth,valign=m]{fig/supplementary/table_parts/19413/col_fine0001.png} &
\includegraphics[width=\tablepartfigsize\linewidth,valign=m]{fig/supplementary/table_parts/19642/col_fine0001.png} &
\includegraphics[width=\tablepartfigsize\linewidth,valign=m]{fig/supplementary/table_parts/19644/col_fine0001.png} &
\includegraphics[width=\tablepartfigsize\linewidth,valign=m]{fig/supplementary/table_parts/19387/col_fine0001.png} &
 \\
 \rotatebox[origin=c]{90}{GT}\vspace{0pt} & \includegraphics[width=\tablepartfigsize\linewidth,valign=m]{fig/supplementary/table_parts/18969/col_gt0001.png} & \includegraphics[width=\tablepartfigsize\linewidth,valign=m]{fig/supplementary/table_parts/18991/col_gt0001.png} & 
\includegraphics[width=\tablepartfigsize\linewidth,valign=m]{fig/supplementary/table_parts/19139/col_gt0001.png} & 
\includegraphics[width=\tablepartfigsize\linewidth,valign=m]{fig/supplementary/table_parts/19162/col_gt0001.png} &
\includegraphics[width=\tablepartfigsize\linewidth,valign=m]{fig/supplementary/table_parts/19347/col_gt0001.png} &
\includegraphics[width=\tablepartfigsize\linewidth,valign=m]{fig/supplementary/table_parts/19413/col_gt0001.png} &
\includegraphics[width=\tablepartfigsize\linewidth,valign=m]{fig/supplementary/table_parts/19642/col_gt0001.png} &
\includegraphics[width=\tablepartfigsize\linewidth,valign=m]{fig/supplementary/table_parts/19644/col_gt0001.png} &
\includegraphics[width=\tablepartfigsize\linewidth,valign=m]{fig/supplementary/table_parts/19387/col_gt0001.png} &
 \\
 \\
\end{tabular}
\noindent\rule[0.5ex]{\linewidth}{0.5pt}
%+==============================================================
%================================================================
\begin{tabular}{c@{\hskip 5pt}llllllllll}
\rotatebox[origin=c]{90}{Input}\vspace{-7pt} & \includegraphics[width=\tablepartfigsize\linewidth,valign=m]{fig/supplementary/table_parts/19661/point_cloud0001.png} & \includegraphics[width=\tablepartfigsize\linewidth,valign=m]{fig/supplementary/table_parts/19705/point_cloud0001.png} & 
\includegraphics[width=\tablepartfigsize\linewidth,valign=m]{fig/supplementary/table_parts/21126/point_cloud0001.png} & 
\includegraphics[width=\tablepartfigsize\linewidth,valign=m]{fig/supplementary/table_parts/21134/point_cloud0001.png} &
\includegraphics[width=\tablepartfigsize\linewidth,valign=m]{fig/supplementary/table_parts/21189/point_cloud0001.png} &
\includegraphics[width=\tablepartfigsize\linewidth,valign=m]{fig/supplementary/table_parts/21329/point_cloud0001.png} &
\includegraphics[width=\tablepartfigsize\linewidth,valign=m]{fig/supplementary/table_parts/21331/point_cloud0001.png} &
\includegraphics[width=\tablepartfigsize\linewidth,valign=m]{fig/supplementary/table_parts/21407/point_cloud0001.png} &
\includegraphics[width=\tablepartfigsize\linewidth,valign=m]{fig/supplementary/table_parts/21544/point_cloud0001.png} &
 \\
 \rotatebox[origin=c]{90}{Recon}\vspace{-7pt} & \includegraphics[width=\tablepartfigsize\linewidth,valign=m]{fig/supplementary/table_parts/19661/col_fine0001.png} & \includegraphics[width=\tablepartfigsize\linewidth,valign=m]{fig/supplementary/table_parts/19705/col_fine0001.png} & 
\includegraphics[width=\tablepartfigsize\linewidth,valign=m]{fig/supplementary/table_parts/21126/col_fine0001.png} & 
\includegraphics[width=\tablepartfigsize\linewidth,valign=m]{fig/supplementary/table_parts/21134/col_fine0001.png} &
\includegraphics[width=\tablepartfigsize\linewidth,valign=m]{fig/supplementary/table_parts/21189/col_fine0001.png} &
\includegraphics[width=\tablepartfigsize\linewidth,valign=m]{fig/supplementary/table_parts/21329/col_fine0001.png} &
\includegraphics[width=\tablepartfigsize\linewidth,valign=m]{fig/supplementary/table_parts/21331/col_fine0001.png} &
\includegraphics[width=\tablepartfigsize\linewidth,valign=m]{fig/supplementary/table_parts/21407/col_fine0001.png} &
\includegraphics[width=\tablepartfigsize\linewidth,valign=m]{fig/supplementary/table_parts/21544/col_fine0001.png} &
 \\
 \rotatebox[origin=c]{90}{GT}\vspace{0pt} & \includegraphics[width=\tablepartfigsize\linewidth,valign=m]{fig/supplementary/table_parts/19661/col_gt0001.png} & \includegraphics[width=\tablepartfigsize\linewidth,valign=m]{fig/supplementary/table_parts/19705/col_gt0001.png} & 
\includegraphics[width=\tablepartfigsize\linewidth,valign=m]{fig/supplementary/table_parts/21126/col_gt0001.png} & 
\includegraphics[width=\tablepartfigsize\linewidth,valign=m]{fig/supplementary/table_parts/21134/col_gt0001.png} &
\includegraphics[width=\tablepartfigsize\linewidth,valign=m]{fig/supplementary/table_parts/21189/col_gt0001.png} &
\includegraphics[width=\tablepartfigsize\linewidth,valign=m]{fig/supplementary/table_parts/21329/col_gt0001.png} &
\includegraphics[width=\tablepartfigsize\linewidth,valign=m]{fig/supplementary/table_parts/21331/col_gt0001.png} &
\includegraphics[width=\tablepartfigsize\linewidth,valign=m]{fig/supplementary/table_parts/21407/col_gt0001.png} &
\includegraphics[width=\tablepartfigsize\linewidth,valign=m]{fig/supplementary/table_parts/21544/col_gt0001.png} &
 \\
 
\end{tabular}
\noindent\rule[0.5ex]{\linewidth}{0.5pt}
%+==============================================================
%================================================================
\begin{tabular}{c@{\hskip 5pt}llllllllll}
\rotatebox[origin=c]{90}{Input}\vspace{-7pt} & \includegraphics[width=\tablepartfigsize\linewidth,valign=m]{fig/supplementary/table_parts/21653/point_cloud0001.png} & \includegraphics[width=\tablepartfigsize\linewidth,valign=m]{fig/supplementary/table_parts/22151/point_cloud0001.png} & 
\includegraphics[width=\tablepartfigsize\linewidth,valign=m]{fig/supplementary/table_parts/22235/point_cloud0001.png} & 
\includegraphics[width=\tablepartfigsize\linewidth,valign=m]{fig/supplementary/table_parts/22420/point_cloud0001.png} &
\includegraphics[width=\tablepartfigsize\linewidth,valign=m]{fig/supplementary/table_parts/22536/point_cloud0001.png} &
\includegraphics[width=\tablepartfigsize\linewidth,valign=m]{fig/supplementary/table_parts/22591/point_cloud0001.png} &
\includegraphics[width=\tablepartfigsize\linewidth,valign=m]{fig/supplementary/table_parts/22993/point_cloud0001.png} &
\includegraphics[width=\tablepartfigsize\linewidth,valign=m]{fig/supplementary/table_parts/23111/point_cloud0001.png} &
\includegraphics[width=\tablepartfigsize\linewidth,valign=m]{fig/supplementary/table_parts/23140/point_cloud0001.png} &
 \\
 \rotatebox[origin=c]{90}{Recon}\vspace{-7pt} & \includegraphics[width=\tablepartfigsize\linewidth,valign=m]{fig/supplementary/table_parts/21653/col_fine0001.png} & \includegraphics[width=\tablepartfigsize\linewidth,valign=m]{fig/supplementary/table_parts/22151/col_fine0001.png} & 
\includegraphics[width=\tablepartfigsize\linewidth,valign=m]{fig/supplementary/table_parts/22235/col_fine0001.png} & 
\includegraphics[width=\tablepartfigsize\linewidth,valign=m]{fig/supplementary/table_parts/22420/col_fine0001.png} &
\includegraphics[width=\tablepartfigsize\linewidth,valign=m]{fig/supplementary/table_parts/22536/col_fine0001.png} &
\includegraphics[width=\tablepartfigsize\linewidth,valign=m]{fig/supplementary/table_parts/22591/col_fine0001.png} &
\includegraphics[width=\tablepartfigsize\linewidth,valign=m]{fig/supplementary/table_parts/22993/col_fine0001.png} &
\includegraphics[width=\tablepartfigsize\linewidth,valign=m]{fig/supplementary/table_parts/23111/col_fine0001.png} &
\includegraphics[width=\tablepartfigsize\linewidth,valign=m]{fig/supplementary/table_parts/23140/col_fine0001.png} &
 \\
 \rotatebox[origin=c]{90}{GT}\vspace{0pt} & \includegraphics[width=\tablepartfigsize\linewidth,valign=m]{fig/supplementary/table_parts/21653/col_gt0001.png} & \includegraphics[width=\tablepartfigsize\linewidth,valign=m]{fig/supplementary/table_parts/22151/col_gt0001.png} & 
\includegraphics[width=\tablepartfigsize\linewidth,valign=m]{fig/supplementary/table_parts/22235/col_gt0001.png} & 
\includegraphics[width=\tablepartfigsize\linewidth,valign=m]{fig/supplementary/table_parts/22420/col_gt0001.png} &
\includegraphics[width=\tablepartfigsize\linewidth,valign=m]{fig/supplementary/table_parts/22536/col_gt0001.png} &
\includegraphics[width=\tablepartfigsize\linewidth,valign=m]{fig/supplementary/table_parts/22591/col_gt0001.png} &
\includegraphics[width=\tablepartfigsize\linewidth,valign=m]{fig/supplementary/table_parts/22993/col_gt0001.png} &
\includegraphics[width=\tablepartfigsize\linewidth,valign=m]{fig/supplementary/table_parts/23111/col_gt0001.png} &
\includegraphics[width=\tablepartfigsize\linewidth,valign=m]{fig/supplementary/table_parts/23140/col_gt0001.png} &
 \\
 
\end{tabular}

\end{center}
\caption{
\textbf{Qualitative examples of part-aware reconstruction for Table category.} All presented shapes have reconstruction F-score greater than 0.75. Best viewed in color and zoomed in on-screen.
%Results are shown for non-finetuned model. 
}
\label{fig:table_parts}
\end{figure*}

%% file: fig/supplementary/misc_cases_table.tex
\begin{figure*}[!ht]
\begin{center}
% \begin{overpic} 
% [width=\linewidth]
% {example-image-a}
% \end{overpic}
\setlength\tabcolsep{-10pt}
\begin{tabular}{c@{\hskip 5pt}llllllllll}
\rotatebox[origin=c]{90}{Input}\vspace{-15pt} & \includegraphics[width=.15\linewidth,valign=m]{fig/supplementary/table_recon/good_cases/18871/point_cloud0001.png} & 
\includegraphics[width=.15\linewidth,valign=m]{fig/supplementary/table_recon/good_cases/18942/point_cloud0001.png} &
\includegraphics[width=.15\linewidth,valign=m]{fig/supplementary/table_recon/good_cases/18969/point_cloud0001.png} &
\includegraphics[width=.15\linewidth,valign=m]{fig/supplementary/table_recon/good_cases/19095/point_cloud0001.png} &
\includegraphics[width=.15\linewidth,valign=m]{fig/supplementary/table_recon/good_cases/19171/point_cloud0001.png} &
\includegraphics[width=.15\linewidth,valign=m]{fig/supplementary/table_recon/good_cases/19254/point_cloud0001.png} &
\includegraphics[width=.15\linewidth,valign=m]{fig/supplementary/table_recon/good_cases/19347/point_cloud0001.png} &
\includegraphics[width=.15\linewidth,valign=m]{fig/supplementary/table_recon/good_cases/19449/point_cloud0001.png} &
\includegraphics[width=.15\linewidth,valign=m]{fig/supplementary/table_recon/good_cases/19491/point_cloud0001.png} &
 \\
\rotatebox[origin=c]{90}{Recon} \vspace{-15pt}& \includegraphics[width=.15\linewidth,valign=m]{fig/supplementary/table_recon/good_cases/18871/recon0001.png} & 
\includegraphics[width=.15\linewidth,valign=m]{fig/supplementary/table_recon/good_cases/18942/recon0001.png} &
\includegraphics[width=.15\linewidth,valign=m]{fig/supplementary/table_recon/good_cases/18969/recon0001.png} &
\includegraphics[width=.15\linewidth,valign=m]{fig/supplementary/table_recon/good_cases/19095/recon0001.png} &
\includegraphics[width=.15\linewidth,valign=m]{fig/supplementary/table_recon/good_cases/19171/recon0001.png} &
\includegraphics[width=.15\linewidth,valign=m]{fig/supplementary/table_recon/good_cases/19254/recon0001.png} &
\includegraphics[width=.15\linewidth,valign=m]{fig/supplementary/table_recon/good_cases/19347/recon0001.png} &
\includegraphics[width=.15\linewidth,valign=m]{fig/supplementary/table_recon/good_cases/19449/recon0001.png} &
\includegraphics[width=.15\linewidth,valign=m]{fig/supplementary/table_recon/good_cases/19491/recon0001.png} &
 \\
\rotatebox[origin=c]{90}{GT}\vspace{-8pt} & \includegraphics[width=.15\linewidth,valign=m]{fig/supplementary/table_recon/good_cases/18871/orig0001.png} & 
\includegraphics[width=.15\linewidth,valign=m]{fig/supplementary/table_recon/good_cases/18942/orig0001.png} &
\includegraphics[width=.15\linewidth,valign=m]{fig/supplementary/table_recon/good_cases/18969/orig0001.png} &
\includegraphics[width=.15\linewidth,valign=m]{fig/supplementary/table_recon/good_cases/19095/orig0001.png} &
\includegraphics[width=.15\linewidth,valign=m]{fig/supplementary/table_recon/good_cases/19171/orig0001.png} &
\includegraphics[width=.15\linewidth,valign=m]{fig/supplementary/table_recon/good_cases/19254/orig0001.png} &
\includegraphics[width=.15\linewidth,valign=m]{fig/supplementary/table_recon/good_cases/19347/orig0001.png} &
\includegraphics[width=.15\linewidth,valign=m]{fig/supplementary/table_recon/good_cases/19449/orig0001.png} &
\includegraphics[width=.15\linewidth,valign=m]{fig/supplementary/table_recon/good_cases/19491/orig0001.png} &
 \\
\end{tabular}
\noindent\rule[0.5ex]{\linewidth}{0.5pt}
%+==============================================================
%================================================================
\begin{tabular}{c@{\hskip 5pt}llllllllll}
\rotatebox[origin=c]{90}{Input}\vspace{-15pt} & \includegraphics[width=.15\linewidth,valign=m]{fig/supplementary/table_recon/average_cases/18983/point_cloud0001.png} & 
\includegraphics[width=.15\linewidth,valign=m]{fig/supplementary/table_recon/average_cases/19167/point_cloud0001.png} &
\includegraphics[width=.15\linewidth,valign=m]{fig/supplementary/table_recon/average_cases/19183/point_cloud0001.png} &
\includegraphics[width=.15\linewidth,valign=m]{fig/supplementary/table_recon/average_cases/19211/point_cloud0001.png} &
\includegraphics[width=.15\linewidth,valign=m]{fig/supplementary/table_recon/average_cases/19294/point_cloud0001.png} &
\includegraphics[width=.15\linewidth,valign=m]{fig/supplementary/table_recon/average_cases/19536/point_cloud0001.png} &
\includegraphics[width=.15\linewidth,valign=m]{fig/supplementary/table_recon/average_cases/19537/point_cloud0001.png} &
\includegraphics[width=.15\linewidth,valign=m]{fig/supplementary/table_recon/average_cases/19621/point_cloud0001.png} &
\includegraphics[width=.15\linewidth,valign=m]{fig/supplementary/table_recon/average_cases/19626/point_cloud0001.png} &
 \\
\rotatebox[origin=c]{90}{Recon} \vspace{-15pt}& \includegraphics[width=.15\linewidth,valign=m]{fig/supplementary/table_recon/average_cases/18983/recon0001.png} & 
\includegraphics[width=.15\linewidth,valign=m]{fig/supplementary/table_recon/average_cases/19167/recon0001.png} &
\includegraphics[width=.15\linewidth,valign=m]{fig/supplementary/table_recon/average_cases/19183/recon0001.png} &
\includegraphics[width=.15\linewidth,valign=m]{fig/supplementary/table_recon/average_cases/19211/recon0001.png} &
\includegraphics[width=.15\linewidth,valign=m]{fig/supplementary/table_recon/average_cases/19294/recon0001.png} &
\includegraphics[width=.15\linewidth,valign=m]{fig/supplementary/table_recon/average_cases/19536/recon0001.png} &
\includegraphics[width=.15\linewidth,valign=m]{fig/supplementary/table_recon/average_cases/19537/recon0001.png} &
\includegraphics[width=.15\linewidth,valign=m]{fig/supplementary/table_recon/average_cases/19621/recon0001.png} &
\includegraphics[width=.15\linewidth,valign=m]{fig/supplementary/table_recon/average_cases/19626/recon0001.png} &
 \\
\rotatebox[origin=c]{90}{GT}\vspace{-8pt} & \includegraphics[width=.15\linewidth,valign=m]{fig/supplementary/table_recon/average_cases/18983/orig0001.png} & 
\includegraphics[width=.15\linewidth,valign=m]{fig/supplementary/table_recon/average_cases/19167/orig0001.png} &
\includegraphics[width=.15\linewidth,valign=m]{fig/supplementary/table_recon/average_cases/19183/orig0001.png} &
\includegraphics[width=.15\linewidth,valign=m]{fig/supplementary/table_recon/average_cases/19211/orig0001.png} &
\includegraphics[width=.15\linewidth,valign=m]{fig/supplementary/table_recon/average_cases/19294/orig0001.png} &
\includegraphics[width=.15\linewidth,valign=m]{fig/supplementary/table_recon/average_cases/19536/orig0001.png} &
\includegraphics[width=.15\linewidth,valign=m]{fig/supplementary/table_recon/average_cases/19537/orig0001.png} &
\includegraphics[width=.15\linewidth,valign=m]{fig/supplementary/table_recon/average_cases/19621/orig0001.png} &
\includegraphics[width=.15\linewidth,valign=m]{fig/supplementary/table_recon/average_cases/19626/orig0001.png} &
 \\
\end{tabular}

\noindent\rule[0.5ex]{\linewidth}{0.5pt}
%+==============================================================
%================================================================
\begin{tabular}{c@{\hskip 5pt}llllllllll}
\rotatebox[origin=c]{90}{Input}\vspace{-15pt} & \includegraphics[width=.15\linewidth,valign=m]{fig/supplementary/table_recon/poor_cases/19115/point_cloud0001.png} & 
\includegraphics[width=.15\linewidth,valign=m]{fig/supplementary/table_recon/poor_cases/19486/point_cloud0001.png} &
\includegraphics[width=.15\linewidth,valign=m]{fig/supplementary/table_recon/poor_cases/19634/point_cloud0001.png} &
\includegraphics[width=.15\linewidth,valign=m]{fig/supplementary/table_recon/poor_cases/19641/point_cloud0001.png} &
\includegraphics[width=.15\linewidth,valign=m]{fig/supplementary/table_recon/poor_cases/19658/point_cloud0001.png} &
\includegraphics[width=.15\linewidth,valign=m]{fig/supplementary/table_recon/poor_cases/19688/point_cloud0001.png} &
\includegraphics[width=.15\linewidth,valign=m]{fig/supplementary/table_recon/poor_cases/19843/point_cloud0001.png} &
\includegraphics[width=.15\linewidth,valign=m]{fig/supplementary/table_recon/poor_cases/20344/point_cloud0001.png} &
\includegraphics[width=.15\linewidth,valign=m]{fig/supplementary/table_recon/poor_cases/20714/point_cloud0001.png} &
 \\
\rotatebox[origin=c]{90}{Recon} \vspace{-15pt}& \includegraphics[width=.15\linewidth,valign=m]{fig/supplementary/table_recon/poor_cases/19115/recon0001.png} & 
\includegraphics[width=.15\linewidth,valign=m]{fig/supplementary/table_recon/poor_cases/19486/recon0001.png} &
\includegraphics[width=.15\linewidth,valign=m]{fig/supplementary/table_recon/poor_cases/19634/recon0001.png} &
\includegraphics[width=.15\linewidth,valign=m]{fig/supplementary/table_recon/poor_cases/19641/recon0001.png} &
\includegraphics[width=.15\linewidth,valign=m]{fig/supplementary/table_recon/poor_cases/19658/recon0001.png} &
\includegraphics[width=.15\linewidth,valign=m]{fig/supplementary/table_recon/poor_cases/19688/recon0001.png} &
\includegraphics[width=.15\linewidth,valign=m]{fig/supplementary/table_recon/poor_cases/19843/recon0001.png} &
\includegraphics[width=.15\linewidth,valign=m]{fig/supplementary/table_recon/poor_cases/20344/recon0001.png} &
\includegraphics[width=.15\linewidth,valign=m]{fig/supplementary/table_recon/poor_cases/20714/recon0001.png} &
 \\
\rotatebox[origin=c]{90}{GT}\vspace{-8pt} & \includegraphics[width=.15\linewidth,valign=m]{fig/supplementary/table_recon/poor_cases/19115/orig0001.png} & 
\includegraphics[width=.15\linewidth,valign=m]{fig/supplementary/table_recon/poor_cases/19486/orig0001.png} &
\includegraphics[width=.15\linewidth,valign=m]{fig/supplementary/table_recon/poor_cases/19634/orig0001.png} &
\includegraphics[width=.15\linewidth,valign=m]{fig/supplementary/table_recon/poor_cases/19641/orig0001.png} &
\includegraphics[width=.15\linewidth,valign=m]{fig/supplementary/table_recon/poor_cases/19658/orig0001.png} &
\includegraphics[width=.15\linewidth,valign=m]{fig/supplementary/table_recon/poor_cases/19688/orig0001.png} &
\includegraphics[width=.15\linewidth,valign=m]{fig/supplementary/table_recon/poor_cases/19843/orig0001.png} &
\includegraphics[width=.15\linewidth,valign=m]{fig/supplementary/table_recon/poor_cases/20344/orig0001.png} &
\includegraphics[width=.15\linewidth,valign=m]{fig/supplementary/table_recon/poor_cases/20714/orig0001.png} &
 \\
\end{tabular}

\end{center}
\caption{
\textbf{Various reconstruction cases for the Table category.} \emph{Top group:} results with F-score greater than $0.9$. \emph{Middle group:} results with F-score between $0.7$ and $0.9$. \emph{Bottom group:} results with F-score between $0.45$ and $0.7$. Best viewed zoomed in on-screen.
}
\label{fig:misc_cases_table}
\end{figure*}